\documentclass[11pt]{article}

\usepackage[T1]{fontenc}
\usepackage{lmodern}
\usepackage[margin=1.15in]{geometry}
\usepackage{amsmath,amssymb,amsthm}
\usepackage{booktabs}
\usepackage{adjustbox}
\usepackage{array}
\usepackage{longtable}
\usepackage{graphicx}

\usepackage{microtype}
\usepackage{xcolor}
\usepackage{listings}
\usepackage[round]{natbib}
\usepackage[colorlinks=true,linkcolor=blue!50!black,citecolor=blue!50!black,urlcolor=blue!50!black]{hyperref}
\providecommand{\doi}[1]{\href{https://doi.org/#1}{\nolinkurl{doi:#1}}}

\newcommand{\norm}[1]{\left\lVert #1 \right\rVert}

\newcommand{\pcelean}[1]{\texttt{#1}}
\newcommand{\NLlmAstraApiMedian}{38.5}
\newcommand{\NLlmSolMedian}{38}
\newcommand{\SeedsGauss}{20}
\newcommand{\SeedsLlmAstraApi}{20}
\newcommand{\BudgetAllowance}{534}
\newcommand{\PFloor}{1.9\times 10^{-6}}
\newcommand{\DockFamily}{7}
\newcommand{\FilterHistAfterGenZero}{10{,}120}
\newcommand{\LlmSessions}{161}
\newcommand{\LlmSessionsWithCode}{58}
\newcommand{\LlmDiffEvoMentions}{115}
\newcommand{\LlmTextCalls}{160}
\newcommand{\LlmTextLive}{0}
\newcommand{\OutTokensGptossHi}{7{,}487}
\newcommand{\OutTokensFable}{1{,}089}
\newcommand{\OutTokensFableHi}{15{,}838}
\newcommand{\MaxTokensGptoss}{16{,}000}
\newcommand{\MaxTokensGptossHi}{28{,}000}
\newcommand{\MaxTokensFable}{16{,}000}
\newcommand{\MaxTokensFableHi}{32{,}000}
\newcommand{\AstraApiCoriolisSeeds}{20}
\newcommand{\LRatioAstraApi}{6.3}
\newcommand{\LRatioOpus}{3.6}
\newcommand{\LRatioGptoss}{3.5}
\newcommand{\LRatioGauss}{2.9}
\newcommand{\LRatioLmiEq}{4.5}
\newcommand{\LRatioSeed}{2.5}
\newcommand{\LRatioArms}{11}
\newcommand{\LRatioSpearman}{-0.83}
\newcommand{\LRatioSpearmanN}{220}
\newcommand{\TightBound}{39}
\newcommand{\TightCount}{70}
\newcommand{\TightRatioLo}{6.0}
\newcommand{\TightRatioHi}{6.9}
\newcommand{\FigChampKzeroThree}{-0.00226}
\newcommand{\TwoN}{0.00226}
\newcommand{\CoriolisResidual}{8.6\times 10^{-9}}
\newcommand{\FigChampN}{38}
\newcommand{\FigChampRho}{0.7720}
\newcommand{\StageARescale}{2.24}
\newcommand{\StageBRatioSoneLlm}{3.0}
\newcommand{\StageBFormRatioMin}{7}
\newcommand{\StageBFormRatioMax}{33}
\newcommand{\LqrRho}{0.4224}
\newcommand{\SatRho}{0.986409}
\newcommand{\SatC}{4.4128}
\newcommand{\SatCneeded}{1.03}
\newcommand{\SatTheta}{1/5}
\newcommand{\SatN}{641}
\newcommand{\SatDecls}{29}
\newcommand{\SatLines}{942}
\newcommand{\SatWitnessZ}{(0, 0, 0, -3/4)}
\newcommand{\SatWitnessV}{3.59}
\newcommand{\SatWitnessU}{0.23}
\newcommand{\SatUmax}{1/5}
\newcommand{\PursuitRcap}{0.03086}
\newcommand{\PCSeeds}{20}
\newcommand{\PCGens}{24}
\newcommand{\PCPop}{28}
\newcommand{\PCAllowance}{534}
\newcommand{\PCAstraMedian}{0.02857}
\newcommand{\PCOpusMedian}{0.02986}
\newcommand{\PCRuns}{240}
\newcommand{\PCExcluded}{1}
\newcommand{\PCArchivedTol}{5\times 10^{-6}}
\newcommand{\PCAstraArchivedSeeds}{20}
\newcommand{\PCSolFirstBatch}{0.04444}
\newcommand{\PCSolArchivedSeeds}{4}
\newcommand{\PCPdGridVsOpusWins}{16}
\newcommand{\PCProposalsXTwo}{1{,}062}
\newcommand{\PCProposalsXFour}{2{,}118}
\newcommand{\PCCmaesXFourMedian}{0.02943}
\newcommand{\PCMutScaleXFour}{2.2\times 10^{-6}}
\newcommand{\PCArchivedStageB}{0.02857}
\newcommand{\PCArchivedStandardL}{0.04522}
\newcommand{\PCKernelChecked}{12}
\newcommand{\PCKernelTheorems}{11}
\newcommand{\SoStart}{88.1}
\newcommand{\SoSeeds}{20}
\newcommand{\SoFamily}{15}
\newcommand{\SoMedCmaes}{105.2}
\newcommand{\SoMedOpusOnly}{115.7}
\newcommand{\SoMedSolOnly}{93.7}
\newcommand{\SoDAstraGauss}{+7.3}
\newcommand{\SoHAstraGauss}{0.0023}
\newcommand{\SoDOpusOnlyOpus}{+11.0}
\newcommand{\SoHOpusOnlyOpus}{1.7\times 10^{-5}}
\newcommand{\SoAheadOpusOnlyOpus}{19}
\newcommand{\SoNOpusOnlyOpus}{20}
\newcommand{\SoDSolOnlySol}{-6.5}
\newcommand{\SoHSolOnlySol}{0.0017}
\newcommand{\SoMoFamily}{6}
\newcommand{\SoRecordedAngle}{126.9}
\newcommand{\SoRecordedRho}{0.9937}
\newcommand{\SoGapToHand}{21.7}
\newcommand{\SoVerified}{158}
\newcommand{\SoChampions}{158}
\newcommand{\SoBestAngle}{128.2}
\newcommand{\SoBestRho}{0.99487}
\newcommand{\CmaesBipopMedian}{121}
\newcommand{\CmaesBipopN}{20}
\newcommand{\CmaesBipopVsGaussW}{18}
\newcommand{\CmaesBipopVsOpusL}{18}
\newcommand{\CmaesBipopVsOpusN}{20}
\newcommand{\CmaesBipopVsGptossL}{14}
\newcommand{\CmaesBipopVsGptossN}{20}
\newcommand{\ExclComparisons}{173}
\newcommand{\ExclStrictExtra}{14}
\newcommand{\ExclStrictChanged}{44}
\newcommand{\ExclStrictCrossings}{4}
\newcommand{\ExclStrictInstanceCrossings}{4}
\newcommand{\IfacePursuitApiBetter}{10}
\newcommand{\IfacePursuitCliBetter}{10}
\newcommand{\IfaceSoApiBetter}{10}
\newcommand{\IfaceSoCliBetter}{10}
\newcommand{\IfaceMinH}{0.94}
\newcommand{\PCPdGridVsOpusP}{1.0\times 10^{-4}}
\newcommand{\PCPdGridVsOpusCliP}{0.0266}
\newcommand{\GateOffCertified}{0}
\newcommand{\GateOnCertified}{515}
\newcommand{\GateProduced}{10{,}120}
\newcommand{\GateOnBound}{344}
\newcommand{\GateBoundObjBound}{152}
\newcommand{\ResynthControllers}{9{,}399}
\newcommand{\ResynthTextbook}{0}
\newcommand{\ResynthSdp}{2}
\newcommand{\ThrustLimit}{0.2}
\newcommand{\CoPursuitAstraD}{+5.1\times 10^{-5}}
\newcommand{\CoPursuitAstraCI}{[4.3\times 10^{-5}, 8.3\times 10^{-5}]}
\newcommand{\CoPursuitSolCI}{[-3.2\times 10^{-3}, 1.6\times 10^{-5}]}
\newcommand{\CoDockMedian}{40}
\newcommand{\CoDockStageOne}{41}
\newcommand{\CoPursuitMedian}{0.02862}
\newcommand{\CoDockProposals}{238}
\newcommand{\CoPursuitProposals}{340}
\newcommand{\CoDockSdpMin}{3}
\newcommand{\CoPursuitSdpMin}{8}
\newcommand{\CoSdpCap}{20{,}000}
\newcommand{\CoDockVsLmiEqD}{12.5}
\newcommand{\CoDockVsLmiEqN}{20}
\newcommand{\PCAstraMedianFive}{0.02857}
\newcommand{\PCSolMedianFive}{0.02960}
\newcommand{\JointLmiRcap}{0.02857}
\newcommand{\JointLmiGap}{0.5}
\newcommand{\CoPursuitAstraPct}{0.2}
\newcommand{\IfaceDockCiLo}{-4.5}
\newcommand{\IfaceDockCiHi}{7.0}
\newcommand{\IfacePursuitCiLo}{-6.2\times 10^{-4}}
\newcommand{\IfacePursuitCiHi}{2.9\times 10^{-4}}
\newcommand{\IfaceSoCiLo}{-2.9}
\newcommand{\IfaceSoCiHi}{2.9}
\newcommand{\GptossHiCiLo}{-25}
\newcommand{\GptossHiCiHi}{18}
\newcommand{\SoPositive}{158}
\newcommand{\SoMinAlpha}{0.34}
\newcommand{\PCFamilyRecorded}{21}
\newcommand{\PCFamilyReported}{16}
\newcommand{\PCFamilyUnreported}{5}
\newcommand{\FirstBatchSol}{41}
\newcommand{\FirstBatchSolFar}{8}
\newcommand{\FirstBatchAstraFar}{1}
\newcommand{\FirstBatchFar}{100}
\newcommand{\ResynthBoundA}{88}
\newcommand{\ResynthBoundB}{223}
\newcommand{\ResynthSeedA}{18}
\newcommand{\ResynthSeedB}{19}
\newcommand{\ResynthHours}{3.7}
\newcommand{\AttackSteps}{5{,}003}
\newcommand{\AttackTail}{0.01333}
\newcommand{\TierTwoAttemptsA}{4}
\newcommand{\TierTwoSecDD}{84}
\newcommand{\TierTwoSecUU}{94}
\newcommand{\TierTwoWitnessDigits}{46}
\newcommand{\TierTwoLongestDigits}{104}
\newcommand{\TierTwoAStatingWitness}{2}
\newcommand{\TROTotal}{11}
\newcommand{\TROTemplate}{11}
\newcommand{\TROClosedA}{0}
\newcommand{\TROClosedB}{0}
\newcommand{\TROCalls}{88}
\newcommand{\TRORounds}{4}
\newcommand{\TROBCells}{9}
\newcommand{\TROBIdentityFull}{8}
\newcommand{\TROCircular}{3}
\newcommand{\SolStaticDiff}{+0.0033}
\newcommand{\SolStaticCI}{[-0.0004, 0.0046]}
\newcommand{\SolStaticFirst}{0.0492}
\newcommand{\ProbeAstraLow}{2{,}470}
\newcommand{\ProbeAstraHigh}{4{,}283}
\newcommand{\ProbeSolLow}{6{,}160}
\newcommand{\ProbeSolHigh}{5{,}678}
\newcommand{\ProbeCalls}{2}
\newcommand{\ECCapHighCI}{[-29.5, 2]}
\newcommand{\ECStopsHiSixteen}{7}
\newcommand{\ECCallsHiSixteen}{160}
\newcommand{\ECOutCapLow}{1{,}123}
\newcommand{\SatChampTotal}{12}
\newcommand{\SatChampFound}{11}
\newcommand{\SatChampWitness}{8}
\newcommand{\SatChampRatioLo}{1.0}
\newcommand{\SatChampRatioHi}{2.2}
\newcommand{\NoiseTotal}{24}
\newcommand{\NoiseDockLo}{0.000498}
\newcommand{\NoiseDockHi}{0.000966}
\newcommand{\NoiseDockPctLo}{2.5}
\newcommand{\NoiseDockPctHi}{4.8}
\newcommand{\NoisePursLo}{0.00259}
\newcommand{\NoisePursHi}{0.00341}
\newcommand{\NoiseSingleShift}{0.4}
\newcommand{\NoiseDockNRatioLo}{1.05}
\newcommand{\NoiseDockNRatioHi}{2.1}
\newcommand{\NoisePursNExtra}{1}
\newcommand{\ZohTotal}{24}
\newcommand{\ZohTail}{$1.2\times 10^{-22}$}
\newcommand{\ZohDockRatioLo}{0.78}
\newcommand{\ZohDockRatioHi}{1.08}
\newcommand{\ZohDockSlower}{8}
\newcommand{\ZohDockFaster}{4}
\newcommand{\ZohDockExtraMax}{3}
\newcommand{\ZohTightRatioLo}{2.9}
\newcommand{\ZohTightRatioHi}{4.7}
\newcommand{\ZohSimTightLo}{1.5}
\newcommand{\ZohSimTightHi}{2.4}
\newcommand{\ZohRestRatioLo}{1.04}
\newcommand{\ZohRestRatioHi}{1.55}
\newcommand{\ZohSimRestLo}{0.83}
\newcommand{\ZohSimRestHi}{1.6}
\newcommand{\ZohTightRcapHi}{0.0318}
\newcommand{\ZohSimTightZohLo}{0.029}
\newcommand{\ZohSimTightZohHi}{0.043}
\newcommand{\ZohSimRestZohHi}{0.024}
\newcommand{\ArchiveTheorems}{70}
\newcommand{\ArchiveEntries}{6}
\newcommand{\ArchiveReconstructed}{3}
\newcommand{\CitedFiles}{102}
\newcommand{\CitedTheorems}{1535}
\newcommand{\CitedAudited}{102}
\newcommand{\RegenTheorems}{97}
\newcommand{\RegenArtifacts}{5}
\newcommand{\InstanceSeeds}{10}
\newcommand{\GridSize}{2500}
\newcommand{\TransferAsymTextCert}{1}
\newcommand{\SeedsAstraStatic}{12}
\newcommand{\FirstBatchAstraStatic}{41.0}
\newcommand{\FirstBatchAstra}{40.0}

\lstdefinestyle{lean}{
  basicstyle=\ttfamily\small,
  keywordstyle=\bfseries,
  morekeywords={theorem,def,by,fun},
  columns=fullflexible,
  keepspaces=true,
  frame=single,
  framerule=0.3pt,
  framesep=4pt,
  xleftmargin=4.3pt,
  xrightmargin=4.3pt,
  mathescape=true
}

\title{\bfseries Proof-Carrying Evolution:\\ Black-Box Search, White-Box
Flight}

\newcommand{\authorbreak}{\end{tabular}\par\begin{tabular}[t]{c}}
\author{%
  Amit Jain\thanks{Massachusetts Institute of Technology.}
  \and
  Sergio Sanchez-Hurtado\thanks{Universidad Polit\'ecnica de Madrid.}
  \authorbreak
  Victor Rodriguez-Fernandez\footnotemark[2]
  \and
  Richard Linares\footnotemark[1]%
}

\date{October 2026}

\usepackage{float}

\newcommand{\OptSpearman}{+0.37}
\newcommand{\OptSpearmanN}{22}

\newcommand{\OptOpusPrimaryProxy}{8{,}594}

\newcommand{\OptAuditedFiles}{44}
\newcommand{\OptAuditedTheorems}{606}
\newcommand{\OptSpearmanPrimary}{+0.07}

\newcommand{\NLmiEqMedian}{52}
\newcommand{\NLlmOpusMedian}{68.5}

\begin{document}
\maketitle

\begin{abstract}
We study whether a language model can help design controllers whose behavior
can be formally checked. In \emph{proof-carrying evolution} (PCE), a search
proposes controllers together with their certificates. Their mathematical forms
are fixed; a proposer chooses their numerical parameters together.
Certificate checks guide selection, and
only a certified candidate can be returned. The deployed artifact is a control
law and its certificate; the language model is used only during design.

We compare four language models with classical search and synthesis on docking
under a speed limit, pursuit of a maneuvering target, and attitude control on
SO(3), using paired runs and shared proposal allowances. Certificates are checked
in exact rational arithmetic during search on the linear problems. On SO(3),
the search uses floating-point checks, followed by exact re-verification of
every reported champion. The best run of each method also receives
Lean~4 proofs of its stated control properties.

On the linear problems, the strongest models beat the standard classical
baselines on every run, but a classical optimizer that chooses certificates
for the same objective closes most of that lead. Among the cited pursuit
champions, the ranking does not survive a change from Euler to zero-order-hold
dynamics. On SO(3), no model
detectably outperforms CMA-ES when mixed with classical mutation. When models
write every proposal, one tested model improves on CMA-ES and the other
performs worse than its mixed variant. All search medians remain below the
hand-tuned law. In a docking control with the objective and mutation operator
held fixed, retaining certified candidates yields new certified designs in
every run; removing this rule yields none.
\end{abstract}

\section{Introduction}

Spacecraft controllers need guarantees whose assumptions and scope can be
reviewed. For many critical maneuvers, opportunities for recovery are limited,
so a review board asks not only whether a design worked in simulation but why
it must work. Mission requirements are therefore often
requirements on the strength of a guarantee under a stated model: the time
within which a docking maneuver is guaranteed to reach its tolerance, the largest
measurement or initial-state uncertainty under which it still succeeds, or the
size of the set of states from which a vehicle is guaranteed to capture, or to
evade, another under stated assumptions about its actions. Meeting such a
requirement means finding both a control law and a certificate that
establishes the property over the relevant operating conditions. This paper
optimizes three such objectives: a certified docking-time bound, a certified
capture radius against a maneuvering target, and the size of a certified
attitude region on SO(3).

Classical methods such as LQR, $H_\infty$ design, gain scheduling and
Lyapunov-based nonlinear design produce explicit laws that engineers can
inspect and analyze, although explicit mathematical form alone does not make a
complex system easy to understand. Choosing an explicit controller family still leaves a
design problem: an initial law may admit no certificate in the chosen family,
or its certified bound may be too weak for the mission. The certificate itself
is part of this problem. Changing its parameters can improve the bound for a
fixed control law, while changing the law can improve performance but lose the
certificate. Searching over both gives the optimizer a way to improve what can
be guaranteed.

Reinforcement learning and neural policies
\citep{gaudet2020sixdof,jain2025multiphase} offer flexible ways to optimize
constrained objectives, but their behavior can be harder to inspect and verify.
A language model can instead act at design time, proposing parameters for an
explicit controller and its certificate; executing the deployed law requires
no model inference. We ask whether a language model can serve as the proposer
inside such a certifiable design space, rather than replace it with a black box.

\begin{figure}[tp]
\centering
\includegraphics[width=\textwidth]{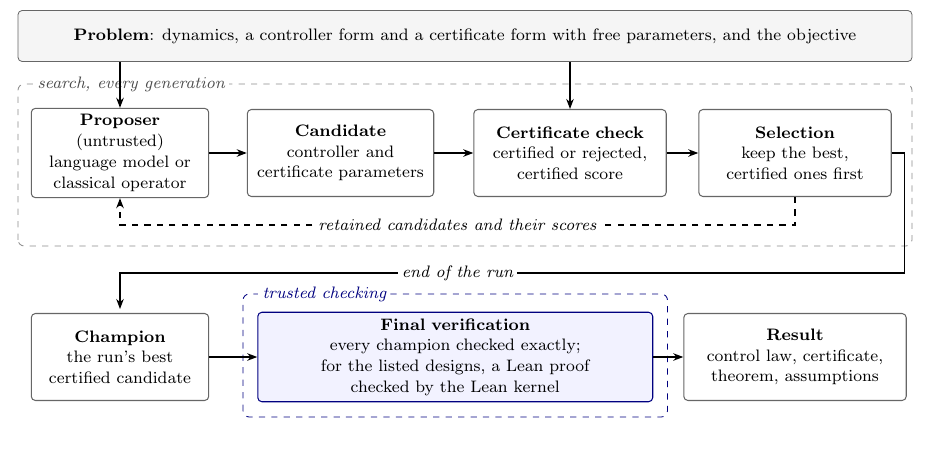}
\caption{Proof-carrying evolution. A proposer chooses the parameters of a
controller and its certificate, selection favors candidates whose certificate
passes the check, and each run's best certified design has an exact check; the
best run of every arm also receives a Lean proof
(Table~\ref{tab:problems} gives the checks on each problem).}
\label{fig:pipeline}
\end{figure}

We call this approach \textbf{proof-carrying evolution}, after proof-carrying
code \citep{necula1997pcc}. In proof-carrying code, an untrusted producer supplies
a program and a proof that it obeys a safety policy; a consumer checks the proof
before accepting the program. PCE applies this separation to control design
(Fig.~\ref{fig:pipeline}). A proposer modifies controller and certificate
parameters, a verifier checks the certificate conditions, and selection favors
candidates with better certified objective values. Retained designs provide
starting points for further proposals. Only a certified candidate can become a
run's result, so acceptance depends on checking rather than on trust in the
proposer. Machine checking establishes a stated property of a stated model; the
choice of property and the model assumptions still require engineering review.

The experiments here use fixed controller and certificate forms: linear state
feedback with a quadratic certificate on the linear problems, and a geometric
proportional--derivative (PD) law with a degree-4 certificate on SO(3). The
proposer chooses numerical parameters; it does not write or modify controller
code. Optimizing uncertainty tolerance or a certified evasion region would extend
this scope and would require new certificate conditions within this framework; here measurement noise is assessed after
search (\S\ref{sec:noise}). The questions are whether a language model proposes better
parameters than classical search and synthesis, and whether checking
certificates during search helps produce certified designs. On the linear
problems, the comparison depends on how the classical methods choose their
certificates (\S\ref{sec:certoptres}).

\paragraph{Contributions.}
\begin{enumerate}
\item \textbf{Method (\S\ref{sec:method}).} PCE, an evolutionary loop in
  which each candidate is a controller paired with its certificate. Inside the loop
  the search proposes and a checker decides: certificates are checked in exact
  rational arithmetic on the linear problems. A certificate compiler turns a checked
  certificate into a Lean proof. An independent audit rejects proofs that rely
  on unfinished obligations or unapproved axioms.
\item \textbf{Comparative evidence (\S\ref{sec:panel}).}
  Four models (gpt-6-astra, gpt-5.6-sol, Claude Opus~5.5 and the open-weight
  gpt-oss-120b; Astra, Sol, Opus and gpt-oss below) share paired seeds and proposal
  allowances with classical baselines (Table~\ref{tab:cross}).
  On the linear problems, Astra and Sol beat the standard classical baselines
  on every run. A classical, objective-aware certificate optimizer specified before its runs shows no
  detected difference from either on docking; on pursuit it beats Sol and trails Astra
  slightly (\S\ref{sec:certoptres}). Among the cited pursuit champions, this
  ranking does not survive zero-order-hold dynamics (\S\ref{sec:zoh}).
  These comparisons use one reasoning-effort setting per model; apart from a detected but unsized
  benefit for Sol on pursuit, they do not establish whether feedback improves
  model proposals.
  On SO(3), no model detectably beats CMA-ES when it writes about one proposal
  in five; Opus does when writing every proposal (\S\ref{sec:so3}).
\item \textbf{Verification during search (\S\ref{sec:filter}).} In a controlled
  docking experiment with the objective and mutation operator held fixed,
  retaining certified candidates produces new certified designs in every run;
  removing that rule produces none. Seeking new
  certificates afterwards recovers \ResynthSdp{} of the \ResynthControllers{}
  controllers produced by the unverified search.
\item \textbf{Guarantees beyond simulation (\S\S\ref{sec:games}, \ref{sec:so3method}, \ref{sec:sat} and~\ref{sec:noise}).} The pursuit designs with Lean
  proofs carry a capture theorem over every admissible thrust sequence of the
  target, and therefore every evader strategy. Under a thrust limit, a
  hand-designed law and the best docking run of every arm but one have decrease
  theorems for the clipped dynamics. The best attitude controller of every arm
  has a regional decrease certificate on SO(3). The cited docking and pursuit
  designs also have decrease and trajectory theorems under bounded measurement
  noise and rounding error, assuming a bound on floating-point error.
\item \textbf{Exploratory optimizer studies (\S\ref{sec:optimizers}).} The search itself need not be trusted, since the checker decides what we keep. On docking, we gave a language model control of the whole search, testing four optimizers with one model and, for one optimizer, four model configurations. Each combination used two seeds and the same exact evaluator. Every newer configuration lowered the cost on both seeds. The ordering of the optimizers changed with the seed, though, and their time and price differed by one to two orders of magnitude. Every final champion carries an audited Lean certificate. One optimizer also read an archived answer when its sandbox allowed it, so we need to isolate the proposer as part of the method.
\end{enumerate}

\section{Related work}\label{sec:related}

\paragraph{Language models in evolutionary search and in proof.} FunSearch
\citep{romeraparedes2024funsearch}, AlphaEvolve \citep{novikov2025alphaevolve}
and ShinkaEvolve \citep{lange2025shinkaevolve} use a language model as the
mutation operator of an evolutionary search over programs, and showed that such
loops can discover new algorithms under automated evaluation; ShinkaEvolve
targets the number of evaluations the search needs. Fitness in those systems is
empirical; it does not establish formal properties of the surviving code.
PCE uses the same propose, evaluate and select loop, with certificate checks
governing acceptance. Our search space consists of numerical parameters for
fixed controller and certificate forms. Searching over controller structure,
as those systems search over program code, remains outside our experiments.

Goedel-Prover~V2 \citep{lin2025goedel} uses an open 32B model with Lean compiler
feedback. It reports 90.4\% on MiniF2F (pass@32, with self-correction).
DeepSeek-Prover-V2 \citep{ren2025deepseekproverv2} and AlphaProof
\citep{hubert2025alphaproof} also prove competition mathematics.
To our knowledge, none gates engineering synthesis.
In PCE a prover would repair obligations beyond the templates
(\S\ref{sec:tier2}); all obligations here have templates (\S\ref{sec:decision}).

Verina \citep{ye2025verina} tests joint generation of code, specification and
proof in Lean. Its best frontier model proves 4.9\% of tasks in one trial
without compiler feedback. Our obligations are more structured: once a
certificate passes its exact check, a deterministic template can construct the
proof. We therefore use templates for the reported designs and test model-based
proof repair separately (\S\ref{sec:tier2}).

\paragraph{Language models as optimizers.}
Language models are used for more than evolutionary program search. They can act as general-purpose optimizers, proposing candidates from earlier results and evaluation feedback \citep{yang2024opro}. The problems range from prompts and heuristics to model pipelines, as in DSPy and GEPA \citep{khattab2024dspy,agrawal2025gepa}. More recent approaches let agents write, execute and refine code, and even change their own workflows \citep{jiang2025aide,lee2026metaharness,karpathy2026autoresearch}. How much of the search the model controls differs between these methods, but they all use external evaluation to guide the next improvements. We use a similar approach in PCE, with a separate decision about acceptance: the model can explore freely, while a deterministic certificate checker decides which candidates can be accepted.

\paragraph{Model or procedure.}
How much of an improvement comes from the model, and how much from the optimization procedure? Studies find that both can matter substantially, but a more elaborate search does not necessarily beat a simpler one \citep{zhang2024understanding,gideoni2026simple}. Agent benchmarks show a similar dependence on the model, the agent scaffold and their interaction \citep{chan2025mlebench,kapoor2026hal}. Sometimes a simple pipeline matches a more elaborate agent, and no optimization strategy does best on every problem \citep{xia2025agentless,tan2026omni}. We therefore need to account for the computational cost and what the search procedure contributes, as well as the final performance \citep{kapoor2025aiagents}. We use random search and genetic algorithms as model-free baselines for this reason. PCE also lets us examine two settings: campaigns with the model inside a fixed search loop, and optimizer experiments in which it runs the whole search (\S\ref{sec:optimizers}).

\paragraph{Certified control and certificate synthesis.} \citet{doll2026lean} provide a
systematic stability library in Lean~4 and mathlib, following machine-checked
control and ODE theory in Coq and Isabelle
\citep{cohen2017lasalle,rouhling2018pendulum,immler2012odes}. These are
libraries; they perform no synthesis. Closer to PCE, \citet{devadze2020sos}
formally certify sum-of-squares Lyapunov functions for polynomial systems in
Minlog, Pegasus \citep{sogokon2021pegasus} generates continuous invariants that
KeYmaera~X then proves, and VeriPhy \citep{bohrer2018veriphy} carries a verified
hybrid-system model through to a verified controller executable. Each certifies
or compiles a given controller or model, but none searches for the controller. Our generated obligations follow
the same spirit: docking certificates lift per-step checks to trajectories by an
induction the generator emits, and pursuit certificates by one shared
handwritten lemma, \pcelean{seq\_affine\_decay}. On the synthesis side, FOSSIL~2.0 and the CEGIS line
\citep{edwards2026fossil} synthesize Lyapunov and barrier certificates, and
controllers with them, as neural templates checked by satisfiability-modulo-theories
(SMT) solvers; trust rests on the solver and its floating-point interface. PCE
replaces the solver in the trust base with a proof kernel, through exact rational
sum-of-squares (SOS) witnesses.

\paragraph{Learning and assurance in spacecraft control.} Earlier work by the present
authors and colleagues placed language models in the control loop itself. Prompted and fine-tuned agents flew spacecraft in Kerbal Space Program
differential games, from text and from screenshots
\citep{rodriguezfernandez2024spacecraft,carrasco2025ksp,carrasco2025vlm}. Small fine-tuned models emitted control vectors for orbit transfer, cislunar control
and powered descent \citep{zucchelli2024finetuned}, and a model trained by
group relative policy optimization reasoned its way to control actions
\citep{jain2026grpo}. That line followed learned policies for landing and
multi-phase ascent \citep{gaudet2020sixdof,jain2025multiphase} and compact
recursive networks for optimal control \citep{jain2025trc}; language has also conditioned safe rendezvous trajectory generation
\citep{takubo2025language}. In each case the learned model acts at flight time
or the safety of its output rests on the optimizer that produced it. Closest in
spirit is AutoResearch \citep{jain2026autoresearch}, in which a language model
drives the research loop offline, never operates the vehicle, and has each
result certified against the problem's measured seed noise, a statistical
credibility layer. PCE likewise keeps the model at design time but makes the
credibility layer formal. Where learned policies do fly, run-time assurance wraps
reinforcement-learning policies for docking and inspection in
control-barrier-function shields
\citep{dunlap2023rta,mcquinn2024attitude,dunlap2024sixdof}, and neural
Lyapunov--barrier methods verify trained controllers after the fact
\citep{mandal2024nlb}. Those approaches retain a learned policy in the deployed
controller; PCE uses the model only during design. Separating untrusted search from trusted
checking is not new; to our knowledge, the new combination is a model jointly
proposing controller and certificate parameters in an evolutionary loop,
admission by exact certificate check, and kernel-checked theorems for reported designs.

\section{Method}\label{sec:method}

\subsection{The loop}\label{sec:loop}

PCE searches over a fixed family of controllers and certificates. A
\emph{candidate} is one choice of their parameters $\theta$, specifying a control
law $u = \pi_\theta(x)$ and a function $V_\theta$. The certificate conditions
express the desired closed-loop property, such as decrease of $V_\theta$ within
a region. On
the linear problems the controller is linear state feedback $u = Kz$ and the
certificate a quadratic Lyapunov function $V(z) = z^\top P z$, so a candidate is
the pair $(K, P)$; it is stored as $(K, L)$ with $P = LL^\top$, which makes $P$
positive definite by construction. On SO(3) a candidate is fifteen positive
diagonal gains that set both the controller and the certificate
(\S\ref{sec:so3method}). Following evolutionary-search usage, we call the stored
parameter vector the candidate's \emph{genome}. A \emph{generation} consists of
proposal, checking and selection. The retained candidates form the \emph{elite},
and the best certified candidate of a run is its \emph{champion}.
Table~\ref{tab:problems} lists what is searched and checked on each problem.

At each generation (Fig.~\ref{fig:pipeline}), a language model or classical
operator proposes candidates, typically by modifying retained designs. Each is
checked and scored using the problem's rules (Table~\ref{tab:problems}). On the
linear problems, a tiered verifier checks certification and a simulator measures
mission cost. On SO(3), a floating-point certificate program determines both
admissibility and score.

Selection retains a fixed-size elite. In the main campaigns, the objective is
a certified value, so certified candidates rank ahead of uncertified ones.
Uncertified candidates survive only when too few certified candidates are
available to fill the elite. In the control experiment ranked by simulated cost
(\S\ref{sec:filter}), half the elite places are reserved for certified
candidates. A run returns its best certified candidate.
Section~\ref{sec:problems} defines the problems, and Appendix~\ref{app:spec}
gives the search settings.

\begin{table}[t]
\centering\footnotesize\setlength{\tabcolsep}{4pt}
\begin{tabular}{@{}>{\raggedright\arraybackslash}p{0.09\textwidth}>{\raggedright\arraybackslash}p{0.25\textwidth}>{\raggedright\arraybackslash}p{0.28\textwidth}>{\raggedright\arraybackslash}p{0.30\textwidth}@{}}
\toprule
Problem & Parameters searched & Check during the search & Check afterwards\\
\midrule
Docking & gain $K$ and certificate factor $L$ & exact: stability, the speed-limit
envelope and the docking bound $N$ & Lean proof for the best run of each arm\\[2pt]
Pursuit & gain $K$ and certificate factor $L$ & exact: robust decrease and the
capture radius & exact re-certification of each champion; Lean
proof for the best run of each arm\\[2pt]
SO(3) & fifteen diagonal gains of the controller and the certificate &
floating point: regional decrease, by a linear program & exact re-check of each
champion and of the positivity of $V$; Lean proof for the best run of each arm
and for the hand-tuned law\\
\bottomrule
\end{tabular}
\caption{What is searched and checked on each problem.}
\label{tab:problems}
\end{table}

\subsection{Checking certificates instead of searching for
proofs}\label{sec:decision}

On the linear problems, the verifier checks each certificate condition
(\S\ref{sec:task}) by testing positive definiteness of a quadratic form. This is
sufficient for the required nonnegativity and provides an exact witness: a
weighted sum-of-squares decomposition
\citep{parrilo2003sdp,papachristodoulou2002lyapunov}, computed by $LDL^\top$
factorization in exact rational arithmetic. A non-positive pivot fails this
check and rejects the candidate. The check takes milliseconds without calling
Lean. A successful factorization supplies the proof data: the generated Lean file
states the decomposition as a polynomial identity with the exact coefficients as
data, so checking it needs no proof search, no language model and no external
solver (Appendix~\ref{app:lean}).

Verification proceeds in tiers. Tier~0 rejects candidates using numerical
spectral-radius checks and, on docking, rollout divergence. Tier~1 is the exact rational check, which
performs the same mathematics that Lean verifies, but at evolutionary
throughput. Tier~2, repair by a language model with compiler feedback, is held
in reserve for obligations that no template covers (\S\ref{sec:tier2}). On the
two linear problems every certification decision in this paper was made at
tier~1, by construction. On SO(3) the loop admits candidates with a
floating-point program, and every counted champion is re-verified at tier~1
afterwards (\S\ref{sec:so3}).

\paragraph{Which designs receive Lean proofs.}
Lean independently checks proofs for the best run of every arm in
Tables~\ref{tab:docking}, \ref{tab:pursuit} and~\ref{tab:so3}, including the
optimizer and model-only controls, and the tightest pursuit-sweep designs.
SO(3) proofs retain each run's gains, region level and contraction target.

Other proved designs are the champions of Astra's three docking control experiments, gpt-oss's best
static-prompt pursuit run, the preliminary single-run designs discussed in
\S\ref{sec:results}, the hand-designed saturated law, and the proof-repair file.
The hand-tuned SO(3) law and exact manifold-preserving step also have proofs.
We call these designs \emph{cited}.

For each, the Lean~4 kernel \citep{demoura2021lean4} checks the proof and an
independent audit permits only Lean's three standard axioms
(Appendix~\ref{app:lean}).
The kernel cannot check intent; theorem statements and model assumptions require review
(\S\ref{sec:discussion}).

These rules make verifiability part of selection: the search must find a
controller that performs well and admits a certificate in the chosen family.

\subsection{Proposers}\label{sec:mutation}

Every third generation the model proposes eight candidates; Gaussian
mutation supplies the rest (Table~\ref{tab:settings}).
Prompts show the elite's genomes, scores, certification status and objective,
and state the certification requirements.
Replies are parsed and, on the linear problems, clamped to fixed gain bounds.
Failed calls are replaced by Gaussian children.

We use bare requests with no system prompt or declared tools for all four models.
They cannot execute anything in the panel (but the CLI controls of \S\ref{sec:controls} and the optimizers of \S\ref{sec:optstudies} can execute code).
Astra and Sol share a gateway; Opus uses its provider's API; gpt-oss runs
locally from pinned weights.
We log all prompts, replies and token counts.
The models all use low reasoning effort. Sampling uses each provider's defaults; we did not set or equalize sampling across providers.
The closed models returned the aliases we requested (\texttt{gpt-6-astra},
\texttt{gpt-5.6-sol}, \texttt{claude-opus-5-5}) rather than dated snapshots;
we identify them by name and call dates, 13--27 September 2026.
We use a pinned revision of gpt-oss.

The non-LLM operator is \emph{Gaussian mutation with Lyapunov repair}. Half its
children perturb both controller and certificate parameters. The other half
mutate $K$ and re-solve the discrete Lyapunov equation for the modified loop.
This baseline therefore also uses control-specific structure. On SO(3), where
there is no corresponding Lyapunov equation to solve, mutation multiplies each
gain by a log-normal factor. Model proposals take seconds to minutes, whereas
exact checks take milliseconds, so the loop checks every child and queries the
model only every third generation.

Search cost is metered in \emph{proposals}: every distinct genome offered for
evaluation is charged to a ledger shared by all arms, so an equal budget means
an equal number of proposals offered, not an equal number of generations.

In the optimizer studies (\S\ref{sec:optstudies}), we go a step further and give the language model more control over the search, rather than just proposing candidates within a fixed evolutionary loop. The evaluator, budget counter and audit trail remain independent of the model, so all candidates are still checked under the same conditions. We compare different approaches, from predefined optimization strategies to agents that can organize and adapt their own search process.

Three use engines from GEPA's \texttt{optimize\_anything} framework, which we refer to as GOA \citep{agrawal2026optimizeanything}. \emph{GOA GEPA} uses reflective evolution, where the model revises candidates based on previous results and evaluator feedback \citep{agrawal2025gepa}. \emph{GOA Meta-Harness} gives a coding agent access to the history of previous candidates, their scores and feedback, allowing it to run code and propose up to ten new candidates per iteration \citep{lee2026metaharness}. \emph{GOA Omni} combines several optimization strategies \citep{tan2026omni}. In our configuration, GEPA, Meta-Harness and AutoResearch each receive 62 evaluations, followed by 187 evaluations for GEPA and another 187 for Meta-Harness, with each stage initialized from the best candidate found so far.

We also study a more autonomous approach using \emph{prime-agent}, a self-evolving agent harness built around a persistent Python environment and recursive subagents \citep{karten2026primeagent}. Rather than acting as an optimizer with a predefined search strategy, it follows an AutoResearch-like process in which the agent writes and executes code to explore candidate solutions, while reflecting on previous results and adapting its own search procedure during execution. This allows us to test whether giving the model control not only over the candidates, but also over how the search is conducted, leads to better results.

Despite these differences, all approaches submit genomes as text to the same exact checker, with every evaluation counted against the same proposal budget. We also include random search and the Gaussian-only evolutionary loop of \S\ref{sec:loop} as model-free baselines. Appendix Table~\ref{tab:optconfig} provides the configurations, limits and feedback used in each approach.

\section{The three problems}\label{sec:problems}

The method is applied to three problems on two dynamics models: docking and
pursuit on the linear Clohessy--Wiltshire plant, and attitude control on SO(3).

\subsection{Docking}\label{sec:task}

Our first testbed is in-plane rendezvous under the Clohessy--Wiltshire equations
\citep{clohessy1960terminal}, with mean motion $n = 1.13\times 10^{-3}$~rad/s and
an Euler discretization at $\Delta t = 1$~s. The state $z = (x, y, v_x, v_y)$
holds positions in metres and velocities in metres per second. The mission
initial conditions lie at $\norm{r_0} \approx 1.1$--$1.3$~m from the target, so
the task is the final part of an approach. The equations can be expressed in
other length units if all corresponding constants are rescaled. Here, the
speed limit allows 0.2~m/s at the target, the saturated actuator
considered below delivers at most 0.2~m/s$^2$, and the pursuit evader of
\S\ref{sec:games} thrusts up to 0.01~m/s$^2$. The docking tolerance and pursuit
capture radius use the unweighted state norm, combining the numerical position
and velocity components in the stated units. Thus a radius $r$ bounds both miss
distance and residual speed ($r = 0.02$ means 2~cm and 2~cm/s). This weighting is
a comparison convention, not a mission requirement.

The thrust is $u = Kz$, unbounded in the model except in the saturated case at the end of this subsection.
Safety is expressed as the distance-dependent speed limit
$\norm{v} \le \nu_0 + \nu_1 \norm{r}$ ($\nu_0 = 0.2$, $\nu_1 = 0.5$), which we
inner-approximate by the polynomial safe set
$h(z) = \nu_0^2 + \nu_1^2 (x^2 + y^2) - v_x^2 - v_y^2 \ge 0$. The certificate is
the quadratic function $V(z) = z^\top P z$ with $P = LL^\top$
(\S\ref{sec:loop}). Writing
$H_0=\operatorname{diag}(\nu_1^2,\nu_1^2,-1,-1)$ gives
$h(z)=\nu_0^2+z^\top H_0z$. Let $\mathcal Z_0$ be the mission initial set
listed in Appendix~\ref{app:spec}, and define
$c_{\mathrm{needed}}=\max_{z_0\in\mathcal Z_0}V(z_0)$, the smallest
Lyapunov level containing those starts. Coverage requires
$c_{\mathrm{needed}}\le c$.

For each candidate, fast numerics
estimate the decrease margin, the S-procedure multiplier and the envelope level
$c$. These estimates are rounded conservatively and then checked exactly. For
the cited designs (\S\ref{sec:decision}), the certificate compiler generates
Lean proofs of the following properties:
\begin{itemize}
\item \textbf{Stability}: $\alpha\norm{z}^2 \le V(z) \le \beta\norm{z}^2$ and
  $V(z^+) \le \rho\, V(z)$, hence $V(z_n) \le \rho^n V(z_0)$. The contraction
  $\rho$ is certified directly (Appendix~\ref{app:more} compares this with a padded proxy): a bisection
  whose test is exact, bracketed by a floating-point generalized eigenvalue,
  finds the tightest contraction, which is rounded up to a rational, and the
  exact check confirms $\rho P - A_{\mathrm{cl}}^\top P A_{\mathrm{cl}} \succ 0$
  at that value. Here $\alpha$ and $\beta$ bound the size of $V$, and
  $\rho<1$ bounds its contraction per step.
\item \textbf{Safety}: the envelope $\Omega_c = \{V \le c\}$ lies inside
  $\{h \ge 0\}$ by the S-procedure \citep{boyd1994lmi}. For a multiplier
  $\lambda\ge0$, the checks $H_0+\lambda P\succeq0$ and
  $\lambda c\le\nu_0^2$ imply
  $h(z)\ge\nu_0^2-\lambda V(z)\ge\nu_0^2-\lambda c\ge0$ on $\Omega_c$.
  The envelope is forward-invariant, so every trajectory that
  starts in $\Omega_c$ satisfies the speed limit at every future step
  (\pcelean{safe\_forever}).
\item \textbf{Coverage and docking time}: each mission initial condition satisfies
  $V(z_0)\le c_{\mathrm{needed}}\le c$. From each such start, the state is within the docking
  tolerance after $N$ steps and stays there (\pcelean{docking\_within\_N},
  \pcelean{docking\_stays\_within}). For a cited design the headline number is
  thus itself a kernel-checked theorem.
\end{itemize}

For each mission start, the stability inequalities give
$\alpha\norm{z_n}^2\le V(z_n)\le\rho^nV(z_0)\le\rho^n c_{\mathrm{needed}}$.
The objective is therefore the certified docking-time bound
$N = \min\{\, n : \rho^{\,n}\, c_{\mathrm{needed}} \le \alpha r^2 \,\}$ for the
state-norm tolerance $r = 0.02$. It is computed and \emph{re-verified in exact
arithmetic}; \S\ref{sec:audit} explains why this is not optional. Selection
therefore acts on what can be proved rather than on what is observed.

\paragraph{Saturating actuators.}\label{sec:satmethod} The docking model above leaves the thrust unbounded. When the thrust saturates,
$u = \mathrm{sat}(Kz)$ componentwise on $[-u_m, u_m]$, the closed loop becomes
piecewise and no single polynomial identity covers it. Writing the deadzone as
$\psi = Kz - \mathrm{sat}(Kz)$, the
loop becomes $z^+ = A_{\mathrm{cl}} z - B\psi$, and $\psi$ satisfies a regional
sector condition \citep{gomesdasilva2005antiwindup}: for diagonal $T \succeq 0$
and $G = \theta K$, if $|((K - G)z)_i| \le u_m$ then
$\psi^\top T(\psi - Gz) \le 0$. The certificate is then a joint SOS in
$(z, \psi)$,
\begin{equation}
M_{\mathrm{sat}} =
\begin{bmatrix}
\rho_s P - A_{\mathrm{cl}}^\top P A_{\mathrm{cl}} &
  A_{\mathrm{cl}}^\top P B - G^\top T\\[2pt]
(\cdot)^\top & 2T - B^\top P B
\end{bmatrix} \succeq 0,
\end{equation}
together with one row per thruster placing $\{V \le c\}$ inside the region where
the sector condition holds. The generated Lean file treats each thruster as below, inside or above its
unsaturated range. It states nine cell lemmas, one for each pair of thruster
states, with a separate SOS identity for each cell. A dispatch theorem combines
the cases. This proof needs case analysis as well as polynomial identities,
but a template still generates it. Section~\ref{sec:tier2} tests whether a
language model can reconstruct parts of this proof.

\subsection{Pursuit of a maneuvering target}\label{sec:games}

The second problem, on the same plant, adds adversarial robustness. A maneuvering target with
bounded thrust enters the relative dynamics as a disturbance,
$z^+ = A_{\mathrm{cl}} z + Bw$ with $\norm{w}^2 \le w_{\max}^2$. A joint
S-procedure certifies the robust decrease $V(z^+) \le \rho_r V(z) + \delta$ for
every admissible $w$ through
\begin{equation}
M_{\mathrm{rob}} \;=\;
\begin{bmatrix}
\rho_r P - A_{\mathrm{cl}}^\top P A_{\mathrm{cl}} & -A_{\mathrm{cl}}^\top P B\\[2pt]
-B^\top P A_{\mathrm{cl}} & \mu I - B^\top P B
\end{bmatrix} \succeq 0,
\qquad \delta = \mu\, w_{\max}^2,
\end{equation}
which is checked exactly at tier~1. A handwritten lemma
(\pcelean{seq\_affine\_decay}), proved in Lean like every other statement, lifts this per-step bound to trajectories under
arbitrary bounded disturbance sequences. A corollary,
\pcelean{pursuit\_bound\_feedback}, quantifies over every
state-feedback evader strategy with admissible thrust
(Listing~\ref{lst:leansample}). A simulator can only sample opponent strategies,
whereas the kernel-checked theorem covers all of them.
The objective is the certified ultimate capture radius
$r_{\mathrm{cap}} = \sqrt{C_{\mathrm{cap}}/\alpha}$ with
$C_{\mathrm{cap}} = \delta/(1 - \rho_r)$, the radius that this bound
approaches as $n \to \infty$; in finite time, the corollary certifies entry
into $1.5\,r_{\mathrm{cap}}$ from a stated step $N$ onward. No actuator limit is
modeled for the pursuer.

\begin{lstlisting}[style=lean, float=tp, label={lst:leansample},
  caption={The strategy-quantified capture guarantee for a certified pursuit
  law, abridged from the generated Lean file: $\mathbb{R}^4$ stands for the state
  type $\mathbb{R}\times\mathbb{R}\times\mathbb{R}\times\mathbb{R}$, $V$ is applied
  to the state's four components, $\rho_r$, $\delta$ and $w_{\max}^2$ stand for the
  exact rationals the file carries, and the proof is elided.}]
theorem pursuit_bound_feedback (g : $\mathbb{R}^4 \to \mathbb{R} \times \mathbb{R}$)
    (hg : $\forall$ q, (g q).1$^2$ + (g q).2$^2$ $\le$ $w_{\max}^2$)
    (p : $\mathbb{R}^4$) (n : $\mathbb{N}$) :
    V (trajS g p n) $\le$ $\rho_r^{\,n}$ * V p + $\delta$ / (1 - $\rho_r$) := $\ldots$
\end{lstlisting}

\subsection{Attitude control on SO(3)}\label{sec:so3method}

Docking and pursuit share the same Clohessy--Wiltshire dynamics; they are two
problems on one linear model, differing in the certificate structure (quadratic
envelope versus robust S-procedure) and the objective. On both, every certificate
condition is a linear matrix inequality in the certificate's parameters. Rigid-body
attitude is different. The state is a rotation matrix together with a body
rate, the decrease condition is a degree-4 polynomial inequality in twelve
variables on SO(3), and the guarantee is necessarily regional, because no
continuous feedback stabilizes SO(3) globally \citep{bhat2000topological}. We
step the dynamics with the Cayley transform at $h = 0.1$~s, which maps a
rational rotation to an exactly rational rotation, so every certificate
concerns a state that lies exactly on the manifold. With inertia $J$, diagonal
gains $\widehat G, G_c, M, C, K_\omega$,
$\Psi(R) = \operatorname{tr}(\widehat G(I - R))$ and
$e_R = (G_c R - R^\top G_c)^\vee$, the controller and certificate are
\begin{equation}
u = \omega \times J\omega - e_R - K_\omega\,\omega, \qquad
V(R, \omega) = \Psi(R) + \tfrac12\,\omega^\top M \omega + \omega^\top C\, e_R,
\end{equation}
and the theorem states $V(x^+) \le \rho\, V(x)$ on the region $\{V \le c\}$. The
certificate is a sum-of-squares identity with multipliers for the manifold
constraints and an S-procedure term for the region. We find it with a linear
program over Gram matrices that are diagonally dominant under a fixed diagonal
scaling, re-estimated between solves \citep{ahmadi2019dsos},
then rationalize it and re-derive it over $\mathbb{Q}$.

Positive definiteness of $V$ requires a separate exact check: a sum-of-squares
lower bound $V \ge \alpha\norm{y}^2$ in centered coordinates. The search screen
requires positive gains, but the cross term involving $C$ means that positive
gains alone do not ensure positive definiteness. We therefore check this
property for every counted champion after search (\S\ref{sec:so3}). Because the gains are
diagonal, the problem has a Klein four-group symmetry that splits the
90-monomial basis of the decrease certificate into blocks of sizes
$27 \oplus 21 \oplus 21 \oplus 21$.

We report the size of the region as the largest attitude error, about any axis,
that it is guaranteed to contain at rest,
$\theta_{\max} = \arccos\bigl(1 - c/(\operatorname{tr}\widehat G - \min_i \hat g_i)\bigr)$;
a state with nonzero rate must satisfy the full condition $V \le c$. The score is
this certified \emph{angle} rather than the level of $V$, which a rescaled gain
would inflate at no cost. In the geometric PD family \citep{lee2010geometric,chaturvedi2011attitude},
the certificate is tied to the controller ($\widehat G=G_c$, $M=J$).
A hand-tuned law has a certified region containing every attitude at rest
with error up to \SoRecordedAngle$^\circ$, with $\rho=\SoRecordedRho$.
The kernel checks the manifold-preserving step in \pcelean{SO3Step} and
regional decrease with a non-empty-region witness in \pcelean{SO3AttitudePD}.

\begin{figure}[tp]
\centering
\includegraphics[width=\textwidth]{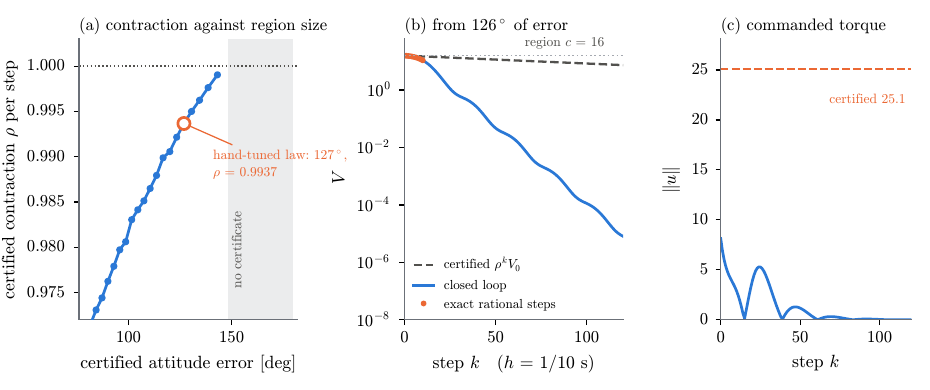}
\caption{The hand-tuned SO(3) law's regional certificate: (a)~the certified
contraction against the size of the region, (b)~the closed loop from the edge of
the region under the certified envelope, and (c)~the commanded torque against its
certified bound.}
\label{fig:so3cert}
\end{figure}

Figure~\ref{fig:so3cert} shows the certificate's cost and conservatism.
As the region approaches the antipodal set, its contraction approaches one;
beyond it no certificate is found.
From rest near the region's edge, the simulated trajectory decays much
faster than the envelope, with the first ten steps computed exactly.
Commanded torque remains well below its certified bound.

\section{Experimental setup}\label{sec:setup}

\paragraph{Problems and objectives.} On docking (\S\ref{sec:task}) the objective
is to minimize the certified docking-time bound $N$; on pursuit
(\S\ref{sec:games}) it is to minimize the certified capture radius
$r_{\mathrm{cap}}$; and on attitude control (\S\ref{sec:so3method}) it is to
maximize the attitude error that the certified region contains at rest. Each
problem uses one certificate formulation throughout. An \emph{arm} is one
proposer or baseline with fixed settings. Arms share the campaign's seeds and
proposal-accounting rules (Table~\ref{tab:settings}). Population-based arms
receive the same initial population on each seed; the pursuit PD grid instead
uses its own lattice. SO(3) populations are deliberately drawn from a weak
part of the controller family.

\begin{table}[t]
\centering\footnotesize\setlength{\tabcolsep}{4pt}
\begin{tabular}{@{}lccc@{}}
\toprule
 & Docking & Pursuit & SO(3) attitude\\
\midrule
Objective & bound $N$ (min.) & radius $r_{\mathrm{cap}}$ (min.) & angle $\theta_{\max}$ (max.)\\
Paired seeds & \SeedsGauss & \PCSeeds & \SoSeeds\\
Population $\times$ generations & $28 \times 24$ & $\PCPop \times \PCGens$ & $12 \times 10$\\
Proposal allowance per run & \BudgetAllowance & \PCAllowance & 120\\
Model proposals & \multicolumn{3}{c}{a batch of eight every third generation}\\
Share written by the model & about 1 in 8 & about 1 in 8 & about 1 in 5\\
Other & -- & evader thrust $\le 0.01$~m/s$^2$ & $\rho \le 0.995$, weak start\\
\bottomrule
\end{tabular}
\caption{Campaign settings for the three problems.}
\label{tab:settings}
\end{table}

\paragraph{Proposers and baselines.} The four models of \S\ref{sec:mutation} run
on all three problems. They are compared with Gaussian mutation with Lyapunov
repair, with CMA-ES \citep{hansen2001cmaes} over the same genome, and with a
structured classical method for each linear problem. On docking this method is
classical synthesis by an alternating pair of semidefinite programs over
$(K, P)$ on a grid of target contraction rates
\citep{boyd1994lmi,hassibi1999bmi}. It is a local BMI search rather than a
certificate of optimality, and we run it both to the end of its own grid and,
with seeded restarts, to the full allowance (``own grid'' and ``full allowance'' in the tables). On
pursuit the structured method is a PD grid matched to the allowance, with each
gain paired with a certificate from the textbook Lyapunov repair. We call these
the \emph{standard} classical baselines. A further classical optimizer chooses
certificates for each objective, as described below. We report a panel of four
models; other evaluated models are outside the reported panel. Astra and Opus
also run through their providers' coding-agent command-line tools (``CLI'') in
the interface controls (\S\ref{sec:controls}).

\paragraph{Classical certificate optimization.}\label{sec:certopt} The standard
baselines choose their certificates by textbook rules: the pursuit grid closes
each gain with a Lyapunov-equation repair at three shape weights, and docking
synthesis maximizes an absolute contraction margin under a trace cap, which is
not the docking bound. The language models propose gain and certificate
together, so their advantage could come from better certificate choices. To
test this explanation we added, after the main
campaigns, a classical optimizer that shapes the certificate for each problem's
objective, from classical starting gains only. Its comparison family (against
Astra and Sol on both problems, Holm over four) and the reading of each
outcome were fixed in code before its runs.

Both objectives are invariant to certificate scale, which each subproblem
fixes. At a fixed gain and rate, the certificate SDP minimizes $\mu$ on pursuit
subject to $P\succeq I$ and $M_{\mathrm{rob}}\succeq0$, minimizing
$r_{\mathrm{cap}}^2=\mu w_{\max}^2/((1-\rho_r)\alpha)$.
On docking it maximizes $\alpha$ subject to $P\succeq\alpha I$,
mission-start levels at most one, contraction and the safety envelope;
this minimizes $N$ at that rate.
With the certificate fixed, a second SDP improves the gain because the Schur
conditions are affine in $K$.

Pursuit also admits jointly convex invariant-ellipsoid synthesis at each rate
in $Q=P^{-1}$ and $Y=KQ$ \citep{boyd1994lmi}; docking's envelope does not.
Each run evaluates the shared initial population and applies the certificate
step to every gain in that population and in a PD lattice. Pursuit also runs
joint synthesis on 33 rates.
Gain and certificate steps then alternate from the best candidates until the
proposal allowance or \CoSdpCap{}-call solver cap is reached.
All candidates use the same exact evaluator and ledger.

\paragraph{Control experiments.}
Each control changes one setting, except the joint effort/ceiling experiments
(Table~\ref{tab:controls}). Interface controls use Astra's CLI with and
without execution, and Opus's CLI.
Reasoning controls use gpt-oss and Claude Fable~5.1 (Fable), a model used only in these controls, at low and high effort.
High effort also raises the output ceiling: from \MaxTokensFable{} to
\MaxTokensFableHi{} for Fable, and from \MaxTokensGptoss{} to
\MaxTokensGptossHi{} for gpt-oss.
Further gpt-oss arms vary these settings separately, using a three-comparison
family fixed before calls.

Static-prompt controls show Astra, gpt-oss and, on pursuit, Sol only the initial
elite. Gate controls remove verification from Gaussian search or retain it
with the objective fixed (\S\ref{sec:filter}).

\begin{figure}[tp]
\centering
\includegraphics[width=\textwidth]{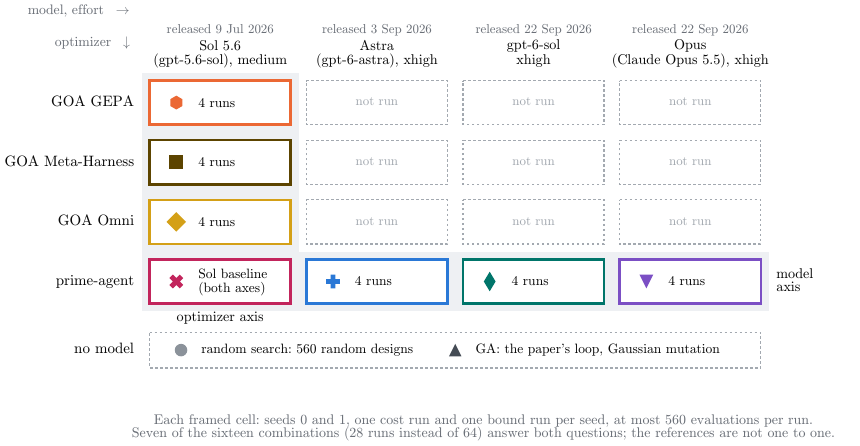}
\caption{Design of the optimizer studies: optimizers by models, with models in release order. We ran the framed cells using two seeds, with a cost run and a bound run per seed. The optimizer axis (a column) and the model axis (a row) meet at the Sol baseline; the references use no model.}
\label{fig:optdesign}
\end{figure}

\paragraph{Optimizer studies.}\label{sec:optstudies}
The docking studies vary the optimizer and model along two axes (Figure~\ref{fig:optdesign}). On the \emph{optimizer axis}, we compare GOA GEPA, GOA Meta-Harness, GOA Omni and prime-agent (\S\ref{sec:mutation}) with gpt-5.6-sol (Sol~5.6) at medium effort. On the \emph{model axis}, we fix prime-agent and compare Sol~5.6 with Astra, gpt-6-sol and Opus at xhigh effort. prime-agent was among the best-performing approaches in the first phase. Its flexible harness lets each model adapt its own search strategy. We chose it for the model comparison for these reasons. The axes share prime-agent with Sol~5.6 as a baseline, giving seven combinations rather than sixteen. We cannot separate model and reasoning-effort effects on the second axis, since both change together.

We run each combination on two seeds (0 and 1), with two docking objectives, giving 28 runs. One objective is the simulated cost $J$ (Appendix~\ref{app:spec}); the other is the padded proxy $N_{\mathrm{proxy}}$ (Appendix~\ref{app:more}), whose final designs we rescore on the actual bound $N$ (\S\ref{sec:task}). The same seed gives each approach the same initial design. A run can submit at most 560 genomes to the checker, and every submission counts, valid or not. We compare cost at a shared checkpoint of 491 evaluations.

Random search and the Gaussian-only GA provide references without a model. Random search evaluates the shared starting design followed by 559 independent samples, selecting directly on $J$ or $N$. The GA instead starts with its own random population; it uses 560 evaluations for cost and 672 for the bound. Every approach uses the same exact checker, and only certified candidates can become champions. We re-certify and audit the final designs. The optimizers run in isolated sandboxes with no access to the evaluator (\S\ref{sec:audit}).

Given the two seeds per combination, we treat the comparisons as descriptive, reporting both seeds and considering a difference consistent only when it has the same sign in both. We report tokens, time and price separately, since equal evaluation budgets do not imply equal computational costs. Comparisons with the campaigns are limited by different budget accounting, unpaired seeds and the use of the proxy for bound optimization. Full run configurations and results are provided in Appendix Table~\ref{tab:optruns} and the corresponding results tables.

\paragraph{Statistical analysis.} The unit of analysis is a run, and comparisons
pair arms by seed. Population-based arms share their initial population within
each pair. All twenty runs in a base campaign solve the same problem instance:
$n$ counts repeated searches, not independent control problems. When a
deterministic baseline is compared with an arm that returns the same design on
every run, as for Astra against the pursuit PD grid, the comparison measures
repeatability on that instance. We report medians because docking bounds are
integers with frequent ties, and a few weak runs produce long tails in both
linear objectives.

For each comparison we report the median paired difference, a percentile
bootstrap 95\% confidence interval (20{,}000 resamples of paired differences; pointwise,
without multiplicity adjustment), and W / L / T (wins, losses, ties) seed counts.
The two-sided Wilcoxon signed-rank test \citep{wilcoxon1945} drops zero
differences. It uses the exact distribution without tied ranks, exact
permutations for tied ranks among at most thirteen differences, and otherwise
the normal approximation without continuity correction.
At $n=\SeedsGauss$ without ties, its smallest attainable $p$ is $\PFloor$.

The test concerns symmetry of paired differences about zero.
We interpret small $p$ as evidence that an arm has a consistent within-seed
advantage, not as a test of the median; the interval describes the median effect.

Campaign tests use Holm adjustment \citep{holm1979} over stated families
(Appendix~\ref{app:more}).
Docking and pursuit families were fixed after results and are exploratory.
SO(3), control and optimizer families were fixed in code before their runs.
Other tests are exploratory and unadjusted; model orderings are descriptive.
No equivalence margin was set, so an undetected difference does not establish
that no difference exists.

A run that loses more than a quarter of its model calls is excluded as
degenerate; the rule excludes five runs, and Appendix~\ref{app:more} gives its
history and a sensitivity check under which no median difference changes sign.

\section{Results}\label{sec:results}

\subsection{Overview across the three problems}\label{sec:panel}

Table~\ref{tab:cross} compares the four models across the problems. Each model
is paired by seed with the strongest standard classical baseline: full-allowance
alternating-SDP synthesis on docking, the PD grid on pursuit, and CMA-ES on
SO(3). The linear problems also include the certificate optimizer. Comparisons
use twenty paired seeds, or nineteen when a run is excluded as degenerate.
The table gives medians and win/loss/tie counts;
Tables~\ref{tab:docking}, \ref{tab:pursuit} and~\ref{tab:so3} give paired
differences and Holm-adjusted $p$-values.

Astra and Sol beat the strongest standard baseline on every run of both linear
problems. gpt-oss loses to that baseline on every run; Opus loses on most docking
runs but wins on most pursuit runs. Certificate optimization closes most of
Astra and Sol's advantage (\S\ref{sec:certoptres}). The ordering changes on
SO(3): Astra matches CMA-ES in median, Sol ranks last, and Opus and gpt-oss lie
between them. The hand-tuned SO(3) law has a larger certified angle than any
search arm's median. In the optimizer studies reported in the last subsection (\S\ref{sec:optimizers}), a language model runs the whole search.

\begin{table}[t]
\centering\footnotesize\setlength{\tabcolsep}{3pt}
\begin{adjustbox}{max width=\textwidth}
\begin{tabular}{@{}lrccrccrc@{}}
\toprule
 & \multicolumn{3}{c}{docking, $N$ (lower is better)} & \multicolumn{3}{c}{pursuit, $r_{\mathrm{cap}}$ (lower)} & \multicolumn{2}{c}{SO(3), degrees (higher)} \\
\cmidrule(lr){2-4}\cmidrule(lr){5-7}\cmidrule(l){8-9}
Model & median & vs.\ synthesis & vs.\ optimizer & median & vs.\ PD grid & vs.\ optimizer & median & vs.\ CMA-ES \\
\midrule
Astra & 38.5 & 20/0/0 & 12/5/3 & 0.02857 & 20/0/0 & 20/0/0 & 105.2 & 9/10/0 \\
Sol & 38 & 20/0/0 & 15/3/2 & 0.02960 & 20/0/0 & 7/13/0 & 98.5 & 1/19/0 \\
Opus & 68.5 & 2/17/1 & 0/20/0 & 0.02986 & 16/4/0 & 0/20/0 & 102.0 & 6/14/0 \\
gpt-oss & 95 & 0/20/0 & 0/20/0 & 0.04009 & 0/20/0 & 0/20/0 & 99.9 & 3/16/0 \\
\midrule
\emph{strongest standard baseline} & 52 & & & 0.03388 & & & 105.2 & \\
\emph{certificate optimizer} & 40 & & & 0.02862 & & & -- & \\
\emph{hand-tuned law} & -- & & & -- & & & 126.9 & \\
\bottomrule
\end{tabular}
\end{adjustbox}
\caption{The four models on the three problems: each model's median, and the
seeds on which it beat, lost to or tied with (W/L/T) the strongest standard
baseline and the certificate optimizer.}
\label{tab:cross}
\end{table}

\subsection{Docking}\label{sec:docking}

\subsubsection{The campaign}\label{sec:campaign}

\begin{figure}[tp]
\centering
\includegraphics[width=\textwidth]{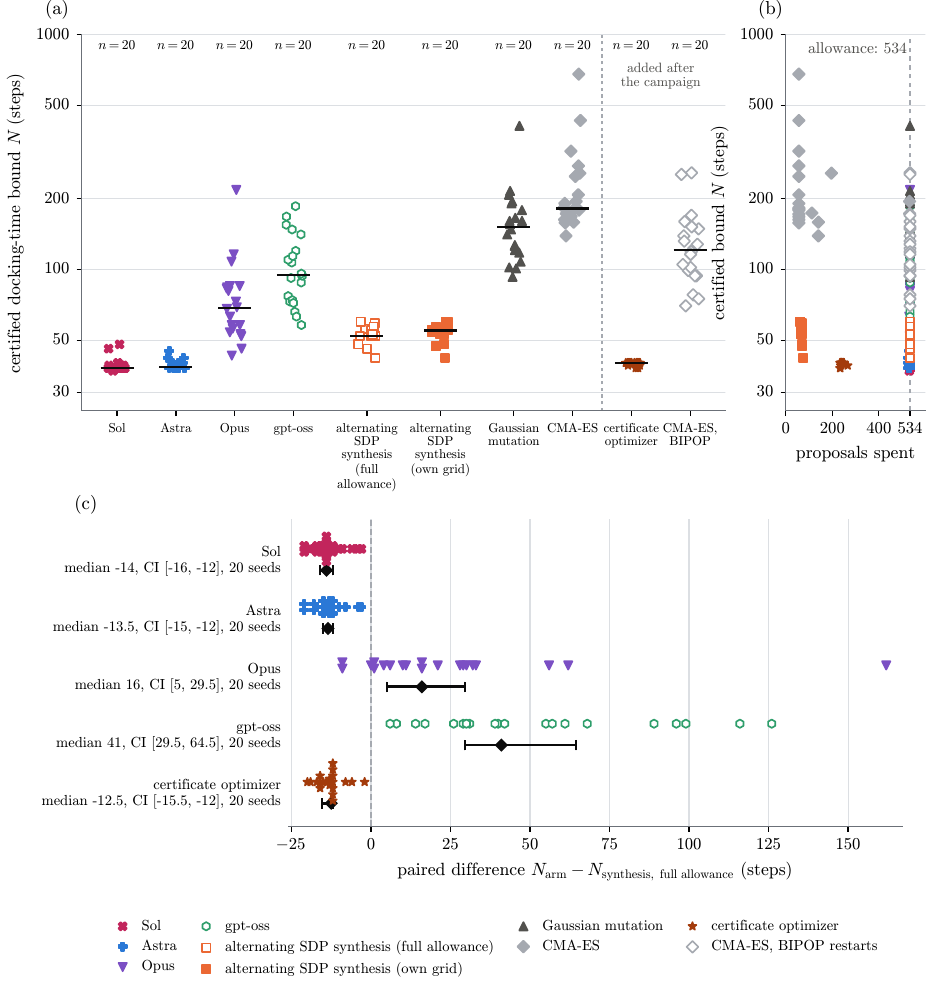}
\caption{The docking campaign, with BIPOP and the certificate optimizer at the right: (a)~the certified bound per seed and arm, (b)~the same bounds against
the proposals each run spent, and (c)~the paired difference from synthesis at the
full allowance, with the median and its bootstrap 95\% CI.}
\label{fig:campaign}
\end{figure}

Table~\ref{tab:docking} and Fig.~\ref{fig:campaign} report the docking campaign.
The paired design compares arms that start from the same population on each
seed. In Table~\ref{tab:docking}, ``Spent'' is the median number of proposals
offered before an arm stops or exhausts its allowance. ``Tokens'' is the median
output tokens per model call, including reasoning. The reference is synthesis
at the full allowance. Above the last horizontal rule, $p$-values are
Holm-adjusted over the \DockFamily{} campaign comparisons.

The final rows report classical methods added after the main campaign: CMA-ES
with BIPOP restarts and a smaller initial step (Appendix
Table~\ref{tab:baselines}), and the certificate optimizer
(\S\ref{sec:certopt}). Their comparisons with synthesis are outside the campaign
family; raw $p$-values are shown in parentheses. The optimizer stops at its
solver-call cap.

\begin{table}[t]
\centering\footnotesize\setlength{\tabcolsep}{1.8pt}
\begin{adjustbox}{max width=\textwidth}
\begin{tabular}{@{}lrrrrlcc@{}}
\toprule
 & & & & \multicolumn{4}{c}{vs.\ synthesis at the full allowance} \\
\cmidrule(l){5-8}
Arm & Median $N$ & Spent & Tokens & med.\ $\Delta N$ & 95\% CI & W / L / T & Holm \\
\midrule
Gaussian mutation & 152 & 534 & -- & $+100.5$ & $[72.0, 118.0]$ & 0 / 20 / 0 & $5.1\times 10^{-4}$ \\
CMA-ES & 182 & 56 & -- & $+133.0$ & $[121.5, 173.0]$ & 0 / 20 / 0 & $5.1\times 10^{-4}$ \\
alternating SDP synthesis (own grid) & 55 & 68 & -- & $+1.0$ & $[0.5, 2.5]$ & 0 / 14 / 6 & $0.00144$ \\
alternating SDP synthesis (full allowance) & 52 & 534 & -- & \multicolumn{4}{c}{reference} \\
\midrule
Astra & 38.5 & 531 & 3{,}349 & $-13.5$ & $[-15.0, -12.0]$ & 20 / 0 / 0 & $5.1\times 10^{-4}$ \\
Sol & 38 & 534 & 5{,}824 & $-14.0$ & $[-16.0, -12.0]$ & 20 / 0 / 0 & $5.1\times 10^{-4}$ \\
Opus & 68.5 & 534 & 1{,}086 & $+16.0$ & $[5.0, 29.5]$ & 2 / 17 / 1 & $0.00144$ \\
gpt-oss & 95 & 534 & 1{,}124 & $+41.0$ & $[29.5, 64.5]$ & 0 / 20 / 0 & $1.3\times 10^{-5}$ \\
\midrule
CMA-ES, BIPOP$^\dagger$ & 121 & 534 & -- & $+67.0$ & $[49.5, 87.5]$ & 0 / 20 / 0 & ($8.8\times 10^{-5}$) \\
certificate optimizer$^\dagger$ & 40 & 238 & -- & $-12.5$ & $[-15.5, -12.0]$ & 20 / 0 / 0 & ($8.1\times 10^{-5}$) \\
\bottomrule
\end{tabular}
\end{adjustbox}
\caption{Docking: the certified docking-time bound $N$ (lower is better) of each
arm, paired within seed against alternating SDP synthesis at the full allowance.
BIPOP CMA-ES and the certificate optimizer were added after the main campaigns.}
\label{tab:docking}
\end{table}

\begin{figure}[tp]
\centering
\includegraphics[width=\textwidth]{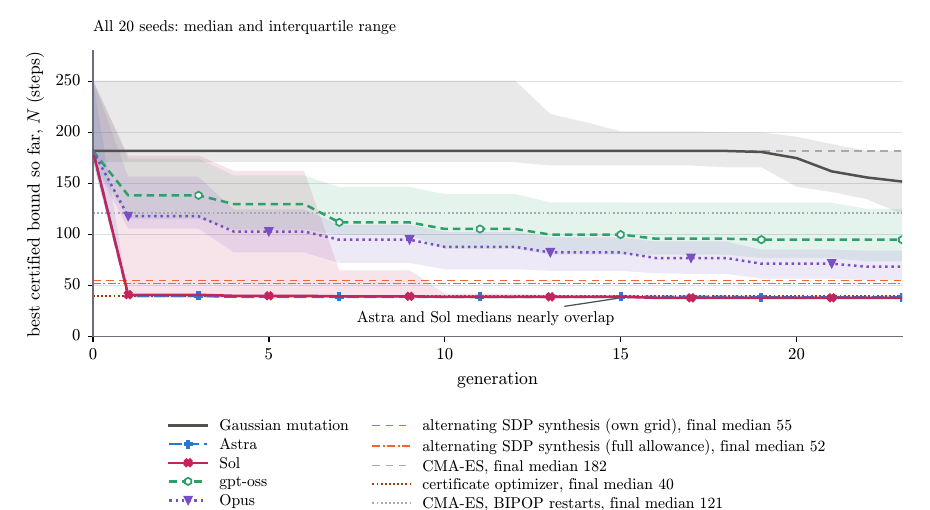}
\caption{Best certified docking bound so far against generation: median and
interquartile range over the twenty seeds, with the classical baselines' final
medians as horizontal lines.}
\label{fig:convergence}
\end{figure}

Synthesis is the strongest standard classical baseline by a wide margin
(Table~\ref{tab:docking}). Astra and Sol beat it at the full allowance on every seed, with median
bounds of \NLlmAstraApiMedian{} and \NLlmSolMedian{} steps.
Opus and gpt-oss rarely or never beat synthesis. Astra and Sol also spend the most reasoning
(Table~\ref{tab:docking}). Objective-aware certificate optimization closes most
of their lead over the standard classical baselines. No difference between
the certificate optimizer and either leading model is detected in the
declared comparison family (\S\ref{sec:certoptres}).

Figure~\ref{fig:convergence} shows how the searches progress. The medians of the
two leading models fall to near their final bounds with the first model batch
and change little afterwards; Opus and gpt-oss improve in steps over
the whole run, and Gaussian mutation stays near the seeded population's bound
until the last generations.

CMA-ES stops early because its initial step, set by the shared pool's spread,
is too wide for the certifiable set: nearly every child is uncertified and the fitness
landscape is flat (Fig.~\ref{fig:campaign}b). Same-step restarts
\citep{auger2005ipop} do not help. BIPOP restarts \citep{hansen2009bipop} with
the step scaled by $0.01$ spend the allowance and improve the median to
\CmaesBipopMedian{} steps (Table~\ref{tab:docking}).
The rerun beats Gaussian mutation on \CmaesBipopVsGaussW{} of
\CmaesBipopN{} seeds, but loses to synthesis and both leading models on every
seed. It also trails Opus and gpt-oss on \CmaesBipopVsOpusL{} of
\CmaesBipopVsOpusN{} and \CmaesBipopVsGptossL{} of
\CmaesBipopVsGptossN{} seeds.
The step scaling was the best tried and is therefore tuned
(Appendix Table~\ref{tab:baselines}).

\paragraph{Structure of the champions.}
Astra approximately cancels the Coriolis coupling in \AstraApiCoriolisSeeds{} of
\SeedsLlmAstraApi{} runs, but the measured benefit is at most about one step
(Appendix Table~\ref{tab:coriolis}).
The Lyapunov factor proposed with the gain tracks the bound
(Appendix~\ref{app:more}).

\begin{figure}[tp]
\centering
\begin{minipage}[b]{0.49\textwidth}\centering
\includegraphics[width=\textwidth]{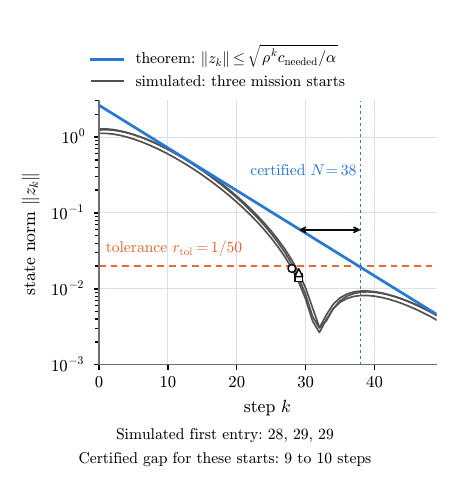}\\[-2pt]
{\small (a)}
\end{minipage}\hfill
\begin{minipage}[b]{0.49\textwidth}\centering
\includegraphics[width=\textwidth]{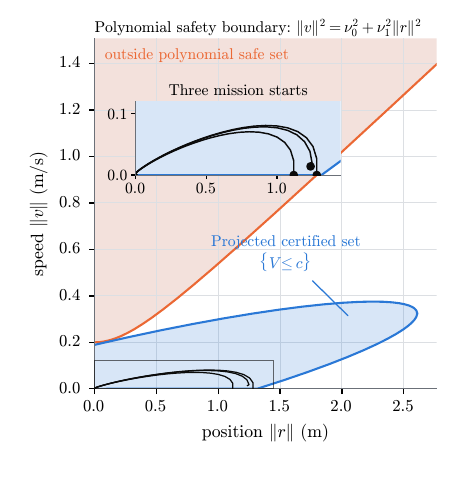}\\[-2pt]
{\small (b)}
\end{minipage}
\caption{The certificate of a typical Astra run ($N = \FigChampN$ steps,
$\rho = \FigChampRho$): (a)~the theorem's envelope bounding the simulated state
norm from the three mission starts, and (b)~the certified set, projected onto
position and speed, inside the safe set.}
\label{fig:champion}
\end{figure}

Figure~\ref{fig:champion} shows a typical Astra certificate, from seed~1 at
the arm's most common bound, kernel-checked with 19 theorems.
Its envelope $\sqrt{\rho^k c_{\mathrm{needed}}/\alpha}$ bounds the simulated
trajectories, which enter the tolerance before $N$.
These sampled trajectories enter early, but they do not determine the
worst-case entry time over all certified starts.
The exact projection of the certified set contains the three mission starts and their
trajectories and lies inside the polynomial safety boundary
$\|v\|^2=\nu_0^2+\nu_1^2\|r\|^2$.

\paragraph{Thrust.}
The comparison uses unbounded actuators and establishes no ranking under a
thrust limit.
Model champions have gentle simulated peaks, while synthesis commands more;
on every run its certified envelope admits states that would command more than
\ThrustLimit~m/s$^2$
(Supplementary Table~\ref{tab:suppthrust}).
No campaign imposes the limit on all arms.
The best run of each arm is checked under it afterwards (\S\ref{sec:sat}).

\subsubsection{Control experiments on the proposer}\label{sec:controls}

Each experiment compares settings of the same model, paired by seed
(Table~\ref{tab:controls}). Most change one setting; the joint reasoning-effort
controls also change the output ceiling. Experiments concern docking unless
stated otherwise. The Opus interface family covers all three problems; its
row in the table reports docking.

The table gives medians over paired seeds and unadjusted two-sided Wilcoxon
$p$-values, followed in parentheses by Holm-adjusted values where a comparison
family was declared. The docking static-prompt controls use the base problem
and seeds of the modified-instance campaign (\S\ref{sec:instances}).

\paragraph{Interface.}
Astra's execution-enabled CLI let the model run Python outside the proposal
ledger; the other CLI variant disabled execution. Session accounting and the
Opus CLI setup are given in Appendix~\ref{app:more}.
Neither CLI variant differs detectably from the API arm at twenty runs
(Table~\ref{tab:controls}).

For Opus, no interface difference is detected
on any problem in the family fixed before the API calls
(Holm $p \ge \IfaceMinH$). Pursuit wins split
\IfacePursuitCliBetter{} to \IfacePursuitApiBetter{}, and SO(3) wins
\IfaceSoCliBetter{} to \IfaceSoApiBetter{}.
The API-minus-CLI intervals are $[\IfaceDockCiLo,\IfaceDockCiHi]$ steps on
docking, $[\IfacePursuitCiLo,\IfacePursuitCiHi]$ in radius on pursuit, and
$[\IfaceSoCiLo,\IfaceSoCiHi]^\circ$ on SO(3).

\paragraph{Reasoning budget.} The shared effort setting bought different amounts
of reasoning (Table~\ref{tab:docking}); Fable used a median of
\OutTokensFable{} output tokens per call. Raising effort and the output ceiling
helped Fable, which used \OutTokensFableHi{} tokens per call and finished within
a step of the leading models (Table~\ref{tab:controls}).
For gpt-oss, using \OutTokensGptossHi{} tokens per call, no difference was
detected in the comparison fixed before its calls.
The low-minus-high interval, $[\GptossHiCiLo,+\GptossHiCiHi]$ steps,
still permits an effect that matters.

Changing effort and ceiling separately also detected no difference
(Table~\ref{tab:controls}). At low effort no call approached either ceiling
(median \ECOutCapLow{} output tokens), so the differing search medians reflect
run-to-run variation. At high effort, \ECStopsHiSixteen{} of
\ECCallsHiSixteen{} calls reached the lower ceiling; the interval for the
ceiling comparison was $\ECCapHighCI$ steps.

These controls do not isolate the role of reasoning in the cross-model
ordering: changing the effort setting also changes how the model generates
proposals. The observed token use is therefore needed to interpret the
configuration comparisons. Astra and Sol were not rerun at
other efforts.

\paragraph{Feedback.} Astra's static-prompt control sees only the initial elite,
so every call samples the same prompt. Its first batch certifies a median bound of
\FirstBatchAstraStatic{} steps, against \FirstBatchAstra{} with feedback;
selection among later draws closes the gap.
On the base problem, no feedback benefit is detected for Astra on the
\SeedsAstraStatic{} paired seeds
(Table~\ref{tab:controls}).
The static-prompt result limits what we can attribute to feedback. The
separate verification experiment (\S\ref{sec:filter}) tests the benefit of
retaining certified designs during selection.
gpt-oss is ahead with feedback on most docking runs and behind on most pursuit
runs, with neither difference detected (Table~\ref{tab:controls}). Its first
batch's certified bound is far above its final docking bound.

\begin{table}[t]
\centering\footnotesize\setlength{\tabcolsep}{2.5pt}
\begin{tabular}{@{}l>{\raggedright\arraybackslash}p{0.17\textwidth}>{\raggedright\arraybackslash}p{0.19\textwidth}rrrcl@{}}
\toprule
 & & & & \multicolumn{2}{c}{median} & & \\
\cmidrule(lr){5-6}
Control & A & B & $n$ & A & B & A / B / T & $p$ (Holm $p$) \\
\midrule
interface & Astra (coding agent) & Astra & 20 & 38 & 38.5 & 11 / 6 / 3 & $0.18$ \\
interface & Astra (text-only) & Astra & 20 & 38.5 & 38.5 & 6 / 5 / 9 & $0.72$ \\
interface & Opus (CLI) & Opus & 20 & 66.5 & 68.5 & 11 / 8 / 1 & $0.31$ ($0.94$) \\
reasoning effort & Fable (effort low) & Fable (effort high) & 20 & 49.5 & 39 & 2 / 17 / 1 & $2.5\times 10^{-4}$ \\
reasoning effort & gpt-oss & gpt-oss (effort high) & 20 & 95 & 95.5 & 10 / 9 / 1 & $0.70$ \\
feedback & Astra & Astra (no feedback) & 12 & 38.0 & 38.0 & 2 / 5 / 5 & $0.41$ ($0.59$) \\
feedback & gpt-oss & gpt-oss (no feedback) & 20 & 95.0 & 104.5 & 13 / 7 / 0 & $0.0592$ ($0.24$) \\
feedback (pursuit) & gpt-oss & gpt-oss (no feedback) & 19 & 0.04019 & 0.03996 & 7 / 12 / 0 & $0.28$ ($0.55$) \\
feedback (pursuit) & Sol & Sol (no feedback) & 20 & 0.02960 & 0.03259 & 12 / 6 / 2 & $0.0483$ ($0.0483$) \\
output ceiling & gpt-oss (low, 16k) & gpt-oss (low, 28k) & 20 & 95 & 97 & 7 / 11 / 2 & $0.98$ ($0.99$) \\
reasoning effort & gpt-oss (low, 28k) & gpt-oss (high, 28k) & 20 & 97 & 95.5 & 8 / 11 / 1 & $0.49$ ($0.99$) \\
output ceiling & gpt-oss (high, 16k) & gpt-oss (high, 28k) & 20 & 87 & 95.5 & 13 / 6 / 1 & $0.31$ ($0.94$) \\
\bottomrule
\end{tabular}

\caption{Control experiments: each row compares two runs of one model that differ in one setting (in the first two reasoning-effort rows, also the output-token ceiling), paired by seed, on docking unless marked (A / B / T: seeds on
which A is better, B is better, or they tie).}
\label{tab:controls}
\end{table}

\subsubsection{Transfer to modified problems}\label{sec:instances}

Astra's first batch is within a few steps of its final bound on all but
\FirstBatchAstraFar{}
runs. Sol's first-batch median is \FirstBatchSol{} steps, although
\FirstBatchSolFar{} of its 20 runs remain above \FirstBatchFar{} steps.
To probe search versus recall, we fixed variants with the cross-track thruster
at half strength, a tighter speed limit, an initial set a quarter farther out
and geostationary mean motion. A seeded-population feasibility probe preceded
instance selection; selection preceded the campaign.

Unchanged base designs transfer to geostationary orbit with at most a step's
change in any arm's median (Appendix Table~\ref{tab:transfer}).
Under unequal thrust, classical champions still certify on every seed, while
Astra's text-only champions certify on only \TransferAsymTextCert{}:
their nominal gains assume symmetric actuation.
Under the tighter limit and wider initial set, none of the leading arms'
champions certifies, although the seeded population still has certified members.

The campaign uses \InstanceSeeds{} seeds per arm on the modified problems
on which the leading base designs transfer least successfully (Table~\ref{tab:instances}).
Gaussian mutation and CMA-ES need two to six times their base bounds.
Synthesis remains close to its base-problem bound on the seeds it certifies
(Table~\ref{tab:instances}).
On unequal thrust, preserving certified starting designs makes synthesis
certify every seed (Appendix Table~\ref{tab:baselines}).

When told about unequal thrust, Astra and Fable at high effort recover the
base problem's bound, although most of their unchanged base-problem champions
do not certify on that plant (1 and 4 of 20, Appendix Table~\ref{tab:transfer}).
Provider spending caps allow only partial samples and stop every run on the
other modified problems before completion. The experiment is incomplete for
these models; even complete, it would test adaptation to a stated change
rather than distinguish recall from reasoning.
gpt-oss beats Gaussian mutation and loses to synthesis on every instance.

\begin{table}[t]
\centering\footnotesize\setlength{\tabcolsep}{3.5pt}
\begin{tabular}{@{}lrrrrrr@{}}
\toprule
 & \multicolumn{2}{c}{unequal thrusters} & \multicolumn{2}{c}{tighter speed limit} & \multicolumn{2}{c}{wider initial set} \\
\cmidrule(lr){2-3}\cmidrule(lr){4-5}\cmidrule(lr){6-7}
Arm & cert. & median $N$ & cert. & median $N$ & cert. & median $N$ \\
\midrule
Gaussian mutation & 10/10 & 370.5 & 8/10 & 430.0 & 10/10 & 454.0 \\
CMA-ES & 10/10 & 675.5 & 8/10 & 1005.5 & 10/10 & 610.0 \\
alternating SDP synthesis (full allowance) & 8/10$^{\dagger}$ & 51.0 & 10/10 & 64.0 & 10/10 & 62.0 \\
Astra & 4/4 & 38.0 & -- & -- & -- & -- \\
Fable (effort high) & 8/8 & 38.0 & -- & -- & -- & -- \\
gpt-oss & 10/10 & 192.0 & 9/10 & 154.0 & 10/10 & 104.0 \\
gpt-oss (no feedback) & 10/10 & 283.5 & 8/10 & 97.5 & 10/10 & 142.5 \\
\bottomrule
\end{tabular}
\par\smallskip\noindent{\footnotesize $^{\dagger}$On unequal thrusters synthesis certifies nothing of its own on seeds 0 and 4, where the shared seeded population holds certified members (best $N$ = 20{,}187 and 6{,}879); the implementation keeps no certified starting design as its best so far. Rerun keeping those designs (Appendix Table~\ref{tab:baselines}), it certifies 10/10, median 55.0, and reproduces the recorded bound on the other 8 seeds.}

\caption{Docking on the three modified problems: runs that certified out of runs completed, and the median certified bound (``--'': no seed completed before the provider's spending
cap stopped the runs).}
\label{tab:instances}
\end{table}

\subsubsection{Verification during search versus afterwards}\label{sec:filter}

Searching on simulated performance and certifying afterwards can fail even
when simulation looks safe. LQR with $Q=R=I$ has spectral radius \LqrRho{}
and violates no sampled speed limit, yet our completion finds no certificate.
Aggressive LQR-class gains also violate no sampled constraint and certify
stability, yet no safety certificate is found: no admissible level both covers
the mission starts and lies inside the speed-limit set.

These failures do not establish that the controllers are unsafe. They mean
that the controllers do not meet our certificate requirements. A safety
certificate must cover every state in its envelope at every future sample
instant; a rollout checks only the trajectories simulated.

We reran Gaussian search without verification, keeping its operator and seeded
populations and ranking by simulated mission cost. Exact checking afterwards
certifies no member of any final population under either certificate
formulation (Table~\ref{tab:filter}).

Checking the whole search history finds that every certified
genome came from the initial population. None of the
\FilterHistAfterGenZero{} genomes produced by search certifies.
After the first generation, no population member certifies. By comparison,
search with verification retains certified candidates in every run.

The next control holds the simulated-cost objective fixed and changes only
selection. With verification, half the elite places are reserved for certified
candidates; without it, candidates are retained solely by simulated cost.
The verification rule produces \GateOnCertified{} certified genomes out of
\GateProduced{} new proposals, compared with \GateOffCertified{} without the
rule (Table~\ref{tab:filter}).

The objective still affects the quality of the certified designs. With the
verification rule in place, ranking by simulated cost ends at a median bound of
\GateOnBound{} steps; ranking by the certified bound gives
\GateBoundObjBound{} steps.

We also sought fresh certificates for the \ResynthControllers{} controllers
produced after initialization that pass the spectral screen, out of
\GateProduced{} produced in total.
Lyapunov-equation repairs at eight weightings certify \ResynthTextbook{};
the synthesis SDP at twenty rates certifies \ResynthSdp{}
(Table~\ref{tab:filter}).
The recovered controllers are on seeds \ResynthSeedA{} and \ResynthSeedB{},
with bounds \ResynthBoundA{} and \ResynthBoundB{} steps.
The SDP search took \ResynthHours{} worker-hours.

These procedures do certify controllers from the initial population. Their
limited success on later controllers suggests that replacing the certificate
alone does not recover certification under the tested procedures. Other
certificates may still exist. In this experiment, verification acted as a selection pressure during the
search, not as a filter that could be applied afterwards.

\begin{table}[t]
\centering\footnotesize\setlength{\tabcolsep}{3pt}
\begin{adjustbox}{max width=\textwidth}
\begin{tabular}{@{}lrrr>{\raggedright\arraybackslash}p{0.3\textwidth}@{}}
\toprule
Search & Certified & Pass screen & Designs & What was checked \\
\midrule
without verification, every design evaluated (20 seeds) & 144 & 9{,}943 & 10{,}680 & the 144 certified are all in the initial population; 0 of the 10{,}120 designs the search produced certify \\
without verification, final populations (20 seeds) & 0 & 538 & 560 & every design of each final population \\
without verification, padded proxy (2 preliminary seeds) & 0 & 51 & 56 & final populations of 25 and 26; 5 designs failing the spectral screen were never scored \\
\midrule
with verification, same simulated cost (20 seeds) & 515 & 7{,}002 & 10{,}120 & the same search with certified candidates kept in the elite; designs the search produced \\
new certificates sought afterwards (20 seeds) & 2 & 9{,}399 & 10{,}120 & each controller of the unverified search that passes the spectral screen, given a new Lyapunov-equation or SDP certificate \\
\bottomrule
\end{tabular}
\end{adjustbox}
\caption{Searching without verification and checking afterwards: the Gaussian
search ranked by simulated cost, with its two control experiments below the
line. ``Pass screen'': designs that pass the numerical tier-0 screen.}
\label{tab:filter}
\end{table}

\subsection{Pursuit}\label{sec:pursuit}

\subsubsection{The campaign}\label{sec:pursuitcampaign}

\begin{figure}[tp]
\centering
\includegraphics[width=\textwidth]{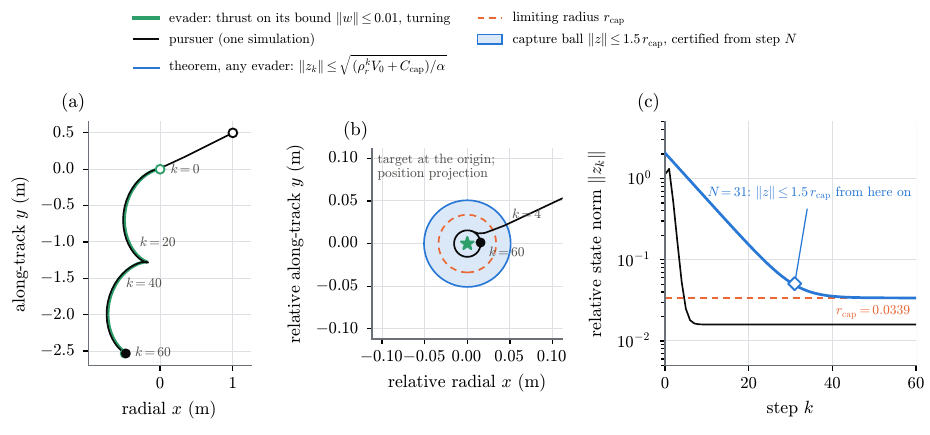}
\caption{The pursuit task on one certified controller against an evader that
thrusts on its bound and keeps turning: (a)~the local orbital frame, (b)~the
target's frame with the certified capture ball, and (c)~the relative state norm
under the theorem's envelope.}
\label{fig:pursuittask}
\end{figure}

Pursuit (\S\ref{sec:games}) tests the same search methods with an adversarial
disturbance. The objective is the certified capture radius
$r_{\mathrm{cap}}$ against an evader whose thrust is bounded by
0.01~m/s$^2$. Figure~\ref{fig:pursuittask} shows the task for one certified
controller, the PD grid's, kernel-checked as \pcelean{CWHPursuitPdGridS0}: the
target keeps thrusting and turning, and the theorem bounds the relative state
for every admissible evader, not only for the one simulated. In panel~(b) the
filled disk is the projection onto the position plane of the capture ball
$\|z\| \le 1.5\,r_{\mathrm{cap}}$ that the kernel-checked corollary certifies
from step $N$ on, and the dashed circle is $r_{\mathrm{cap}}$; panel~(c) shows
the envelope $\sqrt{(\rho_r^k V_0 + C_{\mathrm{cap}})/\alpha}$, with $N$ marked.

A preliminary single-run law certifies $r_{\mathrm{cap}}=\PCArchivedStageB$
under the campaign formulation and \PursuitRcap{} under the formulation used to find it. Its significant parameters are
$k_p\approx0.75$, $k_d\approx1.75$
and two Coriolis-cancelling gains.
An attack over \AttackSteps{} adversarial steps finds no violation of the
certified bound; its strongest tail radius is \AttackTail{}, which does not
establish the worst-case radius.
Replacing the evolved certificate with the textbook shape worsens the bound
to \PCArchivedStandardL{} at the same gains: the certificate is part of the design.

The PD grid searches log-spaced gains
$k_p \in [0.02,2]$ and $k_d \in [0.1,4]$ at three Lyapunov shape weights.
It is sized to the allowance, and failed repairs count against it. It is deterministic at each
allowance, so paired comparisons measure the other arm's variation.

\begin{figure}[tp]
\centering
\includegraphics[width=\textwidth]{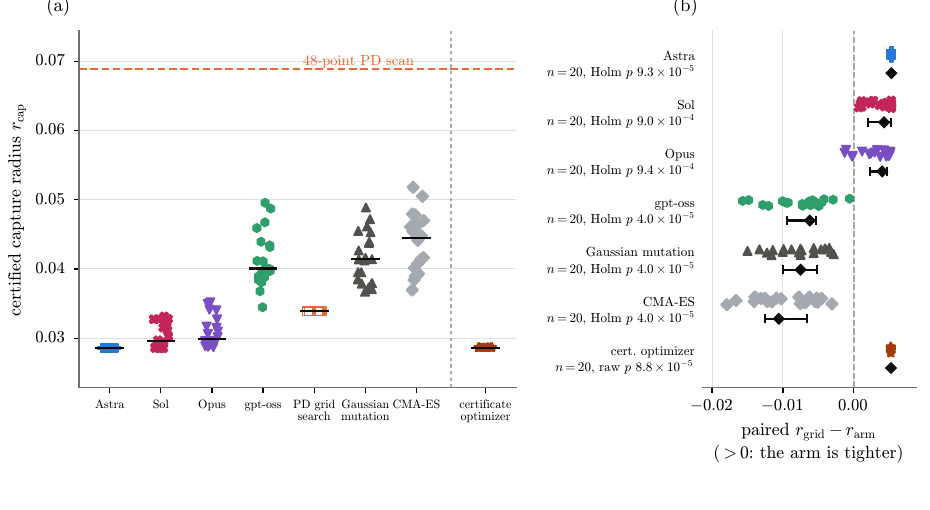}
\caption{The pursuit campaign, with the certificate optimizer at the right:
(a)~the certified capture radius per seed and arm, and (b)~each arm's paired
difference from the PD grid, with the median's bootstrap 95\% CI.}
\label{fig:pursuitcamp}
\end{figure}

Astra and Sol lead the panel, beating every standard classical baseline on
every seed (Table~\ref{tab:pursuit}; Fig.~\ref{fig:pursuitcamp}).
Their median radii are \PCAstraMedianFive{} and \PCSolMedianFive.
The certificate optimizer closes most of this advantage: the signed-rank test
detects a benefit over Sol, while Astra retains a small lead in the declared comparison family
(\S\ref{sec:certoptres}).
Astra returns the same design every time, so its wins against the deterministic
grid measure repeatability on this instance; Sol varies across runs.

The table also reports first-batch radii. In the figure, the dashed line is a
48-point PD scan re-certified under the campaign formulation, and positive
paired differences favor the arm. Campaign tests are Holm-adjusted; the
optimizer's test is raw and outside that family.

Astra proposes its final design in the first batch
(Table~\ref{tab:pursuit}). On \PCAstraArchivedSeeds{} of \PCSeeds{} seeds it
matches the preliminary single-run reference within $\PCArchivedTol$; this is
also the best design found by a finer sweep of robust synthesis (\S\ref{sec:certoptres}).

Sol reaches the same radius within that tolerance on \PCSolArchivedSeeds{}
seeds, through later proposals and selection. Its first-batch radius is
\PCSolFirstBatch{} (Table~\ref{tab:pursuit}).
The static-prompt control also improves, starting at \SolStaticFirst{}, but
finishes worse on most seeds (Table~\ref{tab:controls}).
The signed-rank test detects a feedback benefit in the single-pair family declared
before the calls. The median paired difference, $\SolStaticDiff$, has interval
$\SolStaticCI$, which includes zero.

Opus (\PCOpusMedian) is ahead of the PD grid on \PCPdGridVsOpusWins{} of
\PCSeeds{} runs and beats Gaussian mutation on every seed
(Table~\ref{tab:pursuit}; Supplementary Table~\ref{tab:supppursuit}).
Its advantage over the grid is fragile: the CLI wins just as often, but its
losses are larger (raw $p = \PCPdGridVsOpusCliP$ against
$\PCPdGridVsOpusP$ for the API arm).
No difference between gpt-oss and Gaussian mutation is detected
(Supplementary Table~\ref{tab:supppursuit}).

Every retained run certifies: \PCExcluded{} of \PCRuns{} reported-arm runs was
excluded as degenerate, and every champion re-certifies at its recorded radius.
The best run of each arm, including the optimizer, and the tightest budget-sweep
champions have Lean proofs: \PCKernelChecked{} certificates with
\PCKernelTheorems{} theorems each.

These radii compare unbounded-thrust designs. Simulated peak thrusts far
exceed the \ThrustLimit~m/s$^2$ saturated-actuator limit
(Supplementary Table~\ref{tab:suppthrust}).

\begin{table}[t]
\centering\footnotesize\setlength{\tabcolsep}{3.5pt}
\begin{tabular}{@{}lrrrrrrr@{}}
\toprule
 & & \multicolumn{3}{c}{$r_{\mathrm{cap}}$} & \multicolumn{3}{c}{vs.\ PD grid} \\
\cmidrule(lr){3-5}\cmidrule(l){6-8}
Arm & Cert. & Median & Best & First batch & med.\ $\Delta r$ & W / L / T & Holm \\
\midrule
Gaussian mutation & 20/20 & 0.04140 & 0.03671 & -- & $+0.0075$ & 0 / 20 / 0 & $4.0\times 10^{-5}$ \\
CMA-ES & 20/20 & 0.04446 & 0.03696 & -- & $+0.0106$ & 0 / 20 / 0 & $4.0\times 10^{-5}$ \\
PD grid search$^{\dagger}$ & 20/20 & 0.03388 & 0.03388 & -- & \multicolumn{3}{c}{reference} \\
\midrule
Astra & 20/20 & 0.02857 & 0.02857 & 0.02857 & $-0.0053$ & 20 / 0 / 0 & $9.3\times 10^{-5}$ \\
Sol & 20/20 & 0.02960 & 0.02857 & 0.04444 & $-0.0043$ & 20 / 0 / 0 & $9.0\times 10^{-4}$ \\
Opus & 20/20 & 0.02986 & 0.02870 & 0.04935 & $-0.0040$ & 16 / 4 / 0 & $9.4\times 10^{-4}$ \\
gpt-oss & 20/20 & 0.04009 & 0.03446 & 0.06100 & $+0.0062$ & 0 / 20 / 0 & $4.0\times 10^{-5}$ \\
\midrule
certificate optimizer$^{\ddagger}$ & 20/20 & 0.02862 & 0.02859 & -- & $-0.0053$ & 20 / 0 / 0 & ($8.8\times 10^{-5}$) \\
\bottomrule
\end{tabular}
\par\smallskip\noindent{\footnotesize $^{\dagger}$Deterministic given the allowance. Holm over the campaign's 21 recorded comparisons; adjusting over the 16 between reported arms instead moves none across $p = 0.05$. $^{\ddagger}$Added after the main campaigns and paired outside that family: raw $p$ in parentheses.}

\caption{Pursuit: the certified capture radius (smaller is better) of each arm,
paired within seed against the PD grid.}
\label{tab:pursuit}
\end{table}

\subsubsection{Effect of the search budget}\label{sec:budget}

\begin{figure}[tp]
\centering
\includegraphics[width=\textwidth]{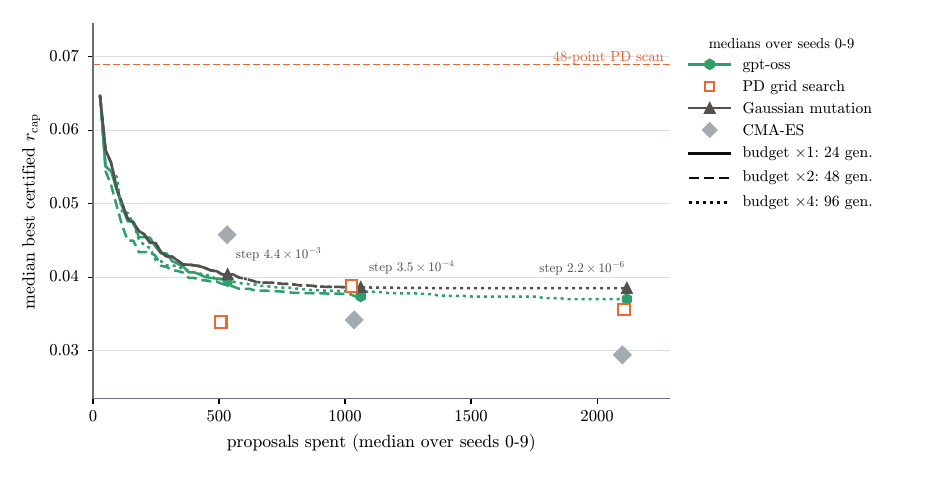}
\caption{Median certified capture radius against proposals spent, seeds 0--9, at
one, two and four times the generations; the grey labels give Gaussian
mutation's step size in its last generation.}
\label{fig:pursuitbudget}
\end{figure}

We reran Gaussian mutation, CMA-ES, the PD grid and gpt-oss on seeds 0--9 at
twice and four times the generations, allowing \PCProposalsXTwo{} and
\PCProposalsXFour{} proposals. CMA-ES and the grid were matched to Gaussian
mutation's spend (Table~\ref{tab:sweep}; Fig.~\ref{fig:pursuitbudget}).

CMA-ES improves with budget and beats Gaussian mutation on every seed at both
larger allowances; its champions converge on the preliminary single-run
reference's gains.
gpt-oss and Gaussian mutation barely improve. The latter's schedule shrinks its
step to $\PCMutScaleXFour$ in the longest run; the schedule was not varied, so
the plateau is conditional on it.

The allowance-specific PD grids are not nested, which explains their
nonmonotone scores. A single lattice refined with the allowance has the same
score at every allowance (Appendix Table~\ref{tab:baselines}). It shows no
detected difference from CMA-ES at twice the allowance and loses on every seed
at four times.

On the ten sweep seeds at the largest allowance, CMA-ES approaches but does not reach Astra's first-batch radius
(\PCCmaesXFourMedian{} against \PCAstraMedian) and is already below Sol's
original-allowance median. Astra and Sol were run only at the original
allowance, so the sweep does not establish how they scale.

\begin{table}[t]
\centering\footnotesize\setlength{\tabcolsep}{5pt}
\begin{tabular}{@{}lrrr@{}}
\toprule
Arm & $\times1$ (24 gen.) & $\times2$ (48 gen.) & $\times4$ (96 gen.) \\
\midrule
Gaussian mutation & 0.04039 & 0.03861 & 0.03850 \\
CMA-ES & 0.04577 & 0.03418 & 0.02943 \\
PD grid search & 0.03388 & 0.03879 & 0.03563 \\
gpt-oss & 0.03946 & 0.03739 & 0.03703 \\
\bottomrule
\end{tabular}
\par\smallskip\noindent{\footnotesize Median certified $r_{\mathrm{cap}}$ on seeds 0--9. Seeds on which the arm certified the smaller radius, of those paired: $\times1$: gpt-oss vs Gaussian 7/10, Holm $0.34$; CMA-ES vs Gaussian 4/10, Holm $0.34$; $\times2$: gpt-oss vs Gaussian 7/10, Holm $0.50$; CMA-ES vs Gaussian 10/10, Holm $0.0156$; $\times4$: gpt-oss vs Gaussian 8/10, Holm $0.25$; CMA-ES vs Gaussian 10/10, Holm $0.0156$. Holm within each level's family of 8.}

\caption{The budget sweep: median certified capture radius on seeds 0--9 at one,
two and four times the generations.}
\label{tab:sweep}
\end{table}

Random mutation improves the certified pursuit radius, whereas it makes
little progress on docking. This pattern is consistent with a more restrictive
certifiable set on docking; the experiments do not isolate feasibility geometry
from the other differences between the problems.
The textbook controller-only grid plateaus; every arm beating it chooses the
certificate with the controller (Tables~\ref{tab:pursuit} and~\ref{tab:sweep};
\S\ref{sec:certoptres}).

The guarantee remains stronger than simulation: \pcelean{pursuit\_bound}
covers every admissible thrust sequence, hence every evader strategy regardless
of its information. Listing~\ref{lst:leansample}'s corollary states the
state-feedback case.

\subsection{Classical certificate optimization on the linear problems}\label{sec:certoptres}

The standard baselines use textbook certificate choices, while the models
choose controller and certificate parameters together. The optimizer of
\S\ref{sec:certopt} tests how much of the models' advantage remains when a
classical method also chooses certificates for the objective. Its results are
the final rows of Tables~\ref{tab:docking} and~\ref{tab:pursuit}.

On docking the optimizer reaches a median of \CoDockMedian{} steps, ahead of
synthesis at the full allowance on all \CoDockVsLmiEqN{} seeds, with a median
paired advantage of \CoDockVsLmiEqD{} steps. Shaping the certificate at fixed classical gains
alone reaches \CoDockStageOne. No difference from either leading model is
detected (Supplementary Table~\ref{tab:suppcertopt}).
The bootstrap intervals describe median paired differences and do not
establish equivalence. Sol's interval excludes zero although the
Holm-adjusted signed-rank test does not detect a difference. The interval is
pointwise and concerns the median; the test uses the signed ranks and adjusts
for the declared comparison family.

On pursuit the optimizer reaches \CoPursuitMedian. The signed-rank test detects
a benefit over Sol, although the median-difference interval
$\CoPursuitSolCI$ includes zero. Astra remains ahead on every seed by a median
of $\CoPursuitAstraD$, with interval $\CoPursuitAstraCI$
(Supplementary Table~\ref{tab:suppcertopt}). The gap is
\CoPursuitAstraPct\% of Astra's radius. The optimizer beats Opus and gpt-oss on
every seed on both problems.

An exploratory finer rate sweep, run after the campaign, reaches Astra's radius,
\JointLmiRcap{}, at $1-\rho_r = \JointLmiGap$. This seed-independent joint
program shows that refining the rate grid can close the remaining gap. It is
convex only at a fixed rate: the sweep finds a best value, not a proven optimum.

Objective-aware certificates explain most of the leading models' advantage.
Astra's pursuit lead survives the declared comparison; an exploratory finer
rate sweep of textbook synthesis reaches the same radius.
This compares equal proposal allowances, not equal computation:
the optimizer uses medians of \CoDockProposals{} proposals on docking and
\CoPursuitProposals{} on pursuit before its solver cap.
Median solver time is \CoDockSdpMin{} and \CoPursuitSdpMin{} minutes per run.

The optimizer is local and provides no global optimality guarantee. Taken
together, these results do not establish a model advantage over classical
design on the linear problems. Astra reaches its result in the first batch;
Sol improves through later proposals and selection. Classical certificate
optimization reaches comparable scores.

\subsection{Attitude control on SO(3)}\label{sec:so3}

\paragraph{The campaign.}
SO(3) tests whether the linear-problem ranking extends to a nonlinear manifold,
using the known law of \S\ref{sec:so3method} as reference.
The panel searches fifteen diagonal gains from weak seeded populations
(median best \SoStart$^\circ$), with Table~\ref{tab:settings}'s allowance.
Every proposal counts against the allowance, certified or not, and is scored
by the floating-point linear program.

Exact re-verification accepts \SoVerified{} of \SoChampions{} counted
champions with their angles unchanged.
This checks decrease, not positivity of $V$.
A separate exact check finds all \SoPositive{} positive definite, with
smallest $\alpha=\SoMinAlpha$.
The best run of each arm also has a Lean proof of regional decrease, torque
bound and a non-empty region at its own gains, level and contraction target
(Supplementary Table~\ref{tab:coverage}).

\begin{figure}[tp]
\centering
\includegraphics[width=\textwidth]{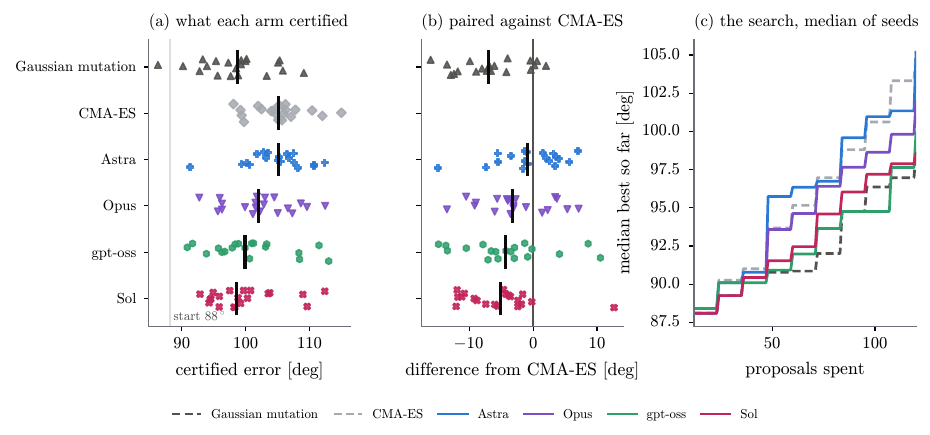}
\caption{The SO(3) campaign: (a)~the certified angle per seed and arm, (b)~each
arm's paired difference from CMA-ES, and (c)~the median best angle so far
against proposals spent.}
\label{fig:so3}
\end{figure}

Table~\ref{tab:so3} and Fig.~\ref{fig:so3} report the campaign. The table's
\emph{Gain} measures improvement over the shared initial best; its tests are
Holm-adjusted over the \SoFamily{} pairs of the campaign arms. The rows below
the line report the model-only control.

CMA-ES is the strongest classical arm, and no model in this mixed design is
detectably better.
Astra matches its median of \SoMedCmaes$^\circ$ and beats Gaussian mutation
by $\SoDAstraGauss^\circ$ ($p = \SoHAstraGauss$).
Opus lies between the baselines in median, and CMA-ES beats Sol
(Table~\ref{tab:so3}).
No arm reaches its final value in the first model batch; the medians rise
in steps over the allowance (Fig.~\ref{fig:so3}c).
The best medians remain \SoGapToHand$^\circ$ below the hand-tuned law,
which no run of this mixed design reaches. The deliberately weak starts make
this a test of recovery within the allowance, not of the best design in the family.

\begin{table}[t]
\centering\footnotesize\setlength{\tabcolsep}{2.5pt}
\begin{adjustbox}{max width=\textwidth}
\begin{tabular}{@{}lrrrrrlrr@{}}
\toprule
Arm & $n$ & Median & Best & Gain over start & vs CMA-ES & 95\% CI & W / L / T & Holm \\
\midrule
Gaussian mutation & 20 & 98.7 & 109.1 & $+12.6$ & $-7.0$ & $[-9.4, -4.1]$ & 3 / 17 / 0 & $0.00123$ \\
CMA-ES & 20 & 105.2 & 115.0 & $+17.1$ & \multicolumn{4}{c}{--} \\
Astra & 19 & 105.2 & 112.4 & $+16.5$ & $-0.9$ & $[-1.6, +3.2]$ & 9 / 10 / 0 & $1.00$ \\
Opus & 20 & 102.0 & 112.4 & $+15.2$ & $-3.3$ & $[-4.9, +0.1]$ & 6 / 14 / 0 & $0.35$ \\
gpt-oss & 19 & 99.9 & 113.0 & $+15.9$ & $-4.3$ & $[-7.1, -1.3]$ & 3 / 16 / 0 & $0.18$ \\
Sol & 20 & 98.5 & 112.4 & $+11.8$ & $-5.1$ & $[-9.0, -3.0]$ & 1 / 19 / 0 & $0.0092$ \\
\midrule
Opus (CLI, model only) & 20 & 115.7 & 128.2 & $+26.7$ & $+10.3$ & $[+5.2, +13.8]$ & 19 / 1 / 0 & $1.5\times 10^{-5}$ \\
Sol (model only) & 20 & 93.7 & 102.8 & $+4.9$ & $-11.2$ & $[-14.4, -7.6]$ & 0 / 20 / 0 & $1.1\times 10^{-5}$ \\
\bottomrule
\end{tabular}
\end{adjustbox}
\caption{SO(3): the attitude error in degrees that the certified region contains
at rest (higher is better), each arm paired by seed against CMA-ES; below the
line, the control in which the model writes every proposal.}
\label{tab:so3}
\end{table}

\paragraph{Control experiment: models writing every proposal.}
Opus through its CLI and Sol also wrote every proposal, replacing the mixed
design's model share of about one in five. The control has its own family of
\SoMoFamily{} comparisons (Table~\ref{tab:so3}).

Opus improves to \SoMedOpusOnly$^\circ$ and beats its CLI mixed arm by
$\SoDOpusOnlyOpus^\circ$. It wins that comparison on
\SoAheadOpusOnlyOpus{} of \SoNOpusOnlyOpus{} seeds
($p = \SoHOpusOnlyOpus$).
It is the only arm detectably better than CMA-ES (Table~\ref{tab:so3}).
Its median remains below the hand-tuned law, but one run reaches
\SoBestAngle$^\circ$ at rest with $\rho = \SoBestRho$.
This contracts more slowly than the hand-tuned law's $\rho = \SoRecordedRho$,
nearer the cap. The controller remains geometric PD; its certificate uses independent weights outside
the tied family ($\widehat G \ne G_c$, $M \ne J$).

Sol instead falls to \SoMedSolOnly$^\circ$, with paired change
$\SoDSolOnlySol^\circ$ against its mixed arm ($p = \SoHSolOnlySol$),
and trails CMA-ES on every seed (Table~\ref{tab:so3}).
Each call requests twelve controllers. Sol supplies near-copies, while Opus
varies its proposals more widely. Increasing the model's share of proposals
therefore has opposite effects for these configurations. This comparison also
uses nine or ten model calls where CMA-ES uses none, so it compares the stated
search configurations without matching inference cost.

\subsection{Optimizer studies: models that run the whole search}\label{sec:optimizers}

\begin{figure}[t]
\centering
\includegraphics[width=\textwidth]{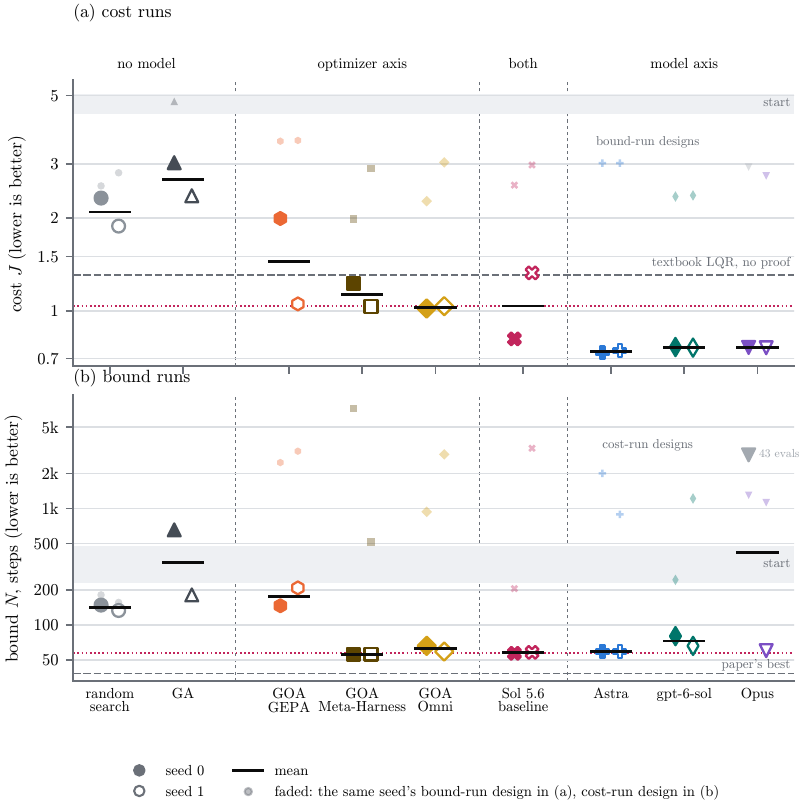}
\caption{Certified cost $J$ (a) and bound $N$ (b). Markers distinguish seeds, bars show geometric means, dotted lines the Sol baseline, and faded markers cross-objective scores; grey marks the time-limited Opus run.}
\label{fig:optoverview}
\end{figure}

The optimizer studies (\S\ref{sec:optstudies}) ask how much the final performance depends on the search strategy and how much on the model. Figure~\ref{fig:optoverview} shows the final certified cost $J$ and bound $N$ for the two docking objectives. There are only two seeds for each combination, so we describe the results and call an improvement consistent only if it occurs on both. The bound runs optimize the padded proxy $N_{\mathrm{proxy}}$, but the figure gives the actual bound $N$ of their final designs. Appendix Tables~\ref{tab:optall} and~\ref{tab:optruns} give the complete results for both axes.

\paragraph{Final performance across approaches.}
Final costs for the three newer models lie between 0.73 and 0.76 on both seeds (Fig.~\ref{fig:optoverview}a). These values are 82 to 85\% below the starting design and 42 to 44\% below LQR with $Q = R = I$ ($J = 1.31$), although our completion procedure cannot certify the reference controller (\S\ref{sec:filter}). The range is wider when we keep Sol~5.6 fixed and change the optimizer: 0.81 to 1.99 across the four approaches. The model-free references, random search and the Gaussian-only GA, have costs between 1.88 and 3.02.

Most approaches also improve the bound substantially from the starting designs' 231 and 477 steps (Fig.~\ref{fig:optoverview}b). Of the 14 bound runs, 11 finish between 56 and 80 steps. GOA GEPA is an exception, at 146 and 208 steps, as is the Opus run stopped by its time limit, at 2{,}896 steps. That Opus design has the same proxy score as Astra's seed-1 design ($N_{\mathrm{proxy}} = \OptOpusPrimaryProxy$), yet Astra's certified bound is 59 steps. Similar proxy scores can therefore correspond to quite different certified bounds.

The final designs differ between objectives. Optimizing the bound gives costs from 1.99 to 3.57, whereas optimizing cost gives bounds from 205 to 7{,}224 steps. These results suggest that an improvement on one objective need not carry over to the other, so the choice of optimization target matters.

\begin{figure}[t]
\centering
\includegraphics[width=\textwidth]{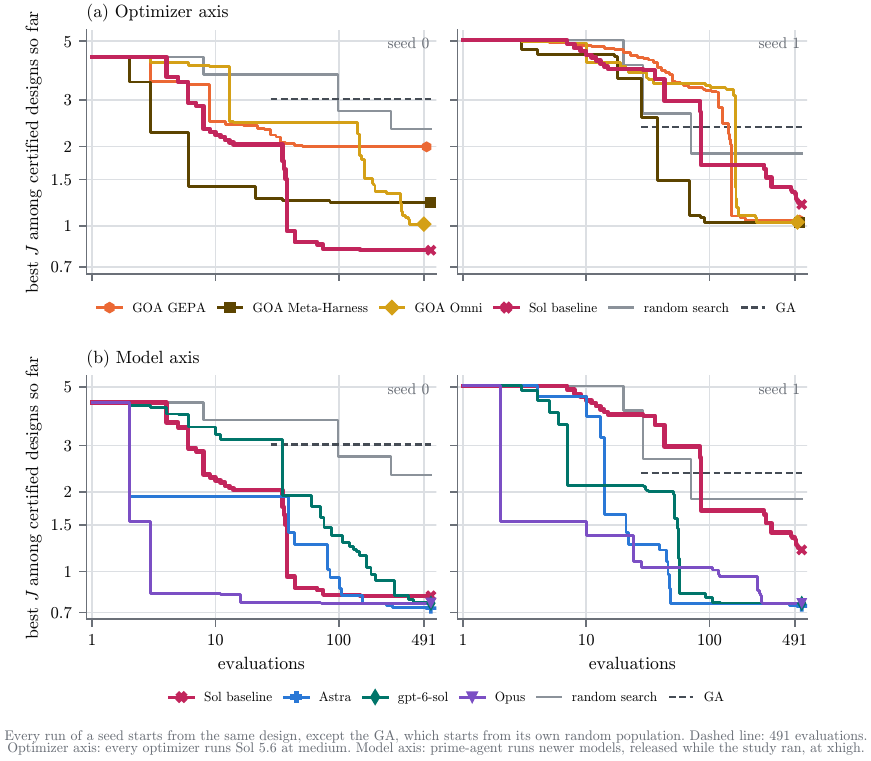}
\caption{Best certified cost $J$ over evaluations for the optimizer (a) and model (b) axes; the dashed line marks 491 evaluations.}
\label{fig:optprogress}
\end{figure}

\begin{figure}[t]
\centering
\includegraphics[width=\textwidth]{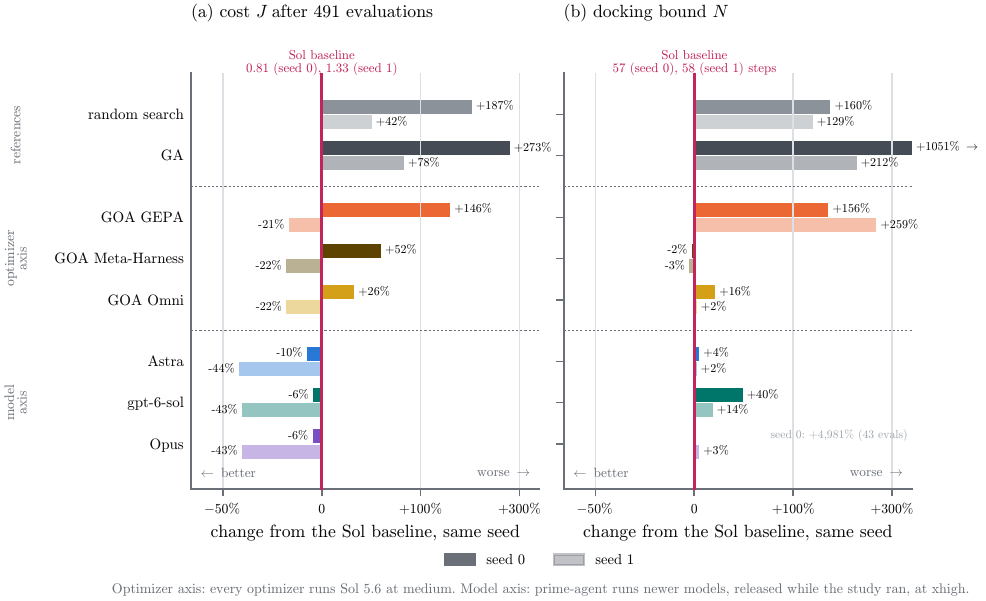}
\caption{Cost (a) and bound (b) relative to the Sol baseline for each seed; leftward bars indicate better results.}
\label{fig:optchange}
\end{figure}

\paragraph{The optimizer or the model.}
The best optimizer changes with the seed when we keep Sol~5.6 fixed (Fig.~\ref{fig:optprogress}a). prime-agent has the lowest cost on seed~0 and the highest on seed~1. Relative to it, the GOA engines have 26 to 146\% higher costs on the first seed, but 21 to 22\% lower costs on the second (Fig.~\ref{fig:optchange}a). With prime-agent fixed instead, all three newer models lower cost relative to Sol~5.6 on both seeds: by 6 to 10\% on seed~0 and 43 to 44\% on seed~1. The checkpoint affects the size of this gap. The Sol baseline on seed~1 continued from 1.33 at 491 evaluations to 1.21 at completion, but the improvement still holds on both seeds.

Although neither axis shows a larger effect on both seeds, the newer models produce much more consistent results, with final costs differing by less than 2\% between seeds and 5\% between models. In comparison, the Sol~5.6 optimizers show differences of up to a factor of 1.9 between seeds. For the bound objective (Fig.~\ref{fig:optchange}b), the differences are smaller, with GOA Meta-Harness being the only optimizer to improve over the Sol baseline on both seeds, by just one and two steps, while none of the newer models improves the bound on either seed.

\begin{figure}[t]
\centering
\includegraphics[width=0.8\textwidth]{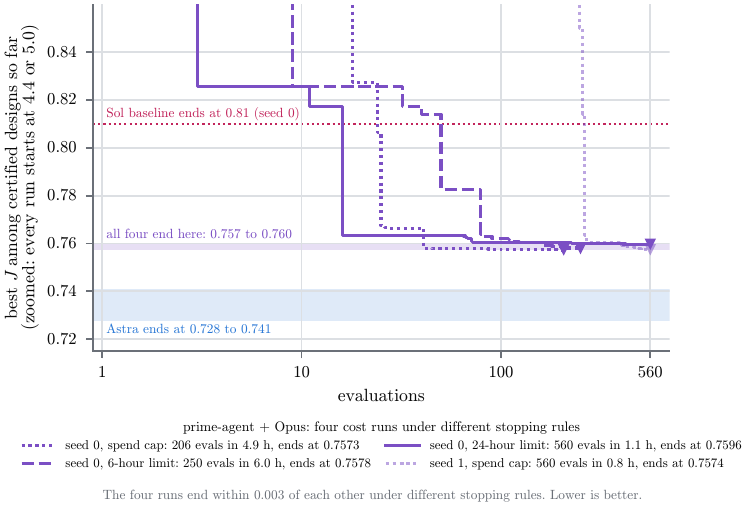}
\caption{Best certified cost $J$ across four prime-agent runs with Opus; bands show final Opus and Astra costs.}
\label{fig:optrepeats}
\end{figure}

We also repeated prime-agent with Opus on the cost objective, running it three times on seed~0 with different stopping limits and once on seed~1 (Fig.~\ref{fig:optrepeats}). The runs followed different search paths and completed between 206 and 560 evaluations, but all finished with very similar costs, between 0.757 and 0.760. In the shortest run, the operator interrupted a computation after two hours and asked the agent to keep subsequent computations short, although its best design had already been found. These runs do not establish repeatability under identical conditions, but the small variation in their final costs suggests that the different search paths can still lead to similar solutions.

\begin{figure}[t]
\centering
\includegraphics[width=0.72\textwidth]{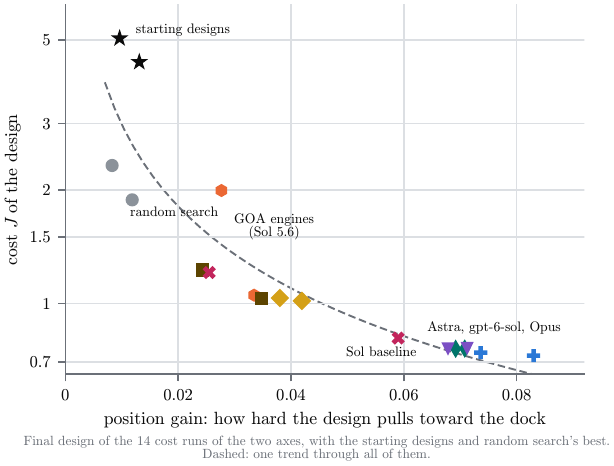}
\caption{Final cost-run designs against position gain $k_p=-(K_{11}+K_{22})/2$; dashed line shows the log-log fit.}
\label{fig:optdesigns}
\end{figure}

The final designs also show a clear relationship between cost and position gain (Fig.~\ref{fig:optdesigns}). The starting designs have gains $k_p$ between 0.010 and 0.013, while the GOA engines reach 0.024 to 0.042 and the newer models 0.068 to 0.083. In general, larger gains correspond to lower costs, although the other parameters of $K$ and the certificate also change, so this relationship cannot be attributed to position gain alone.

With Sol~5.6 fixed, the optimizers differ widely in what they spend (Appendix Fig.~\ref{fig:optresources} and Table~\ref{tab:optall}). A GOA run took 2 to 5 hours and a median of 8 to 48 million tokens, \$12 to \$46 at list price; a prime-agent run took 8 to 11 minutes and about one million tokens, \$0.25 to \$1.49. Per logged call, the prime-agent runs averaged 3 to 33 designs sent to the checker, often from scripts that submit many genomes, compared with 2 to 9 for the GOA engines. A check takes milliseconds and model inference seconds to minutes, so batching plausibly accounts for part of the gap in time; context length and latency also play a role. The newer models reached a cost below 0.8 within 3 to 22 minutes of the start; no GOA run went below 1. The optimizers log different operations as model calls (for example, a reflection, an agent session or an agent turn), so calls are not comparable units of inference.

Rejection rates change during the search (Appendix Fig.~\ref{fig:optverdicts}). For prime-agent, 24 to 45\% of the first hundred candidates are rejected, compared with 7 to 24\% afterwards. Astra's failures due to insufficient margins for the exact checker fall from 12.5\% to 0.1\%. Most remaining failures for Sol~5.6, Astra and gpt-6-sol have a safe region that does not contain the mission start. Opus still fails mainly on thin margins: 69 of its 93 rejections. GOA GEPA and GOA Omni instead show a slight rise in rejections, from 4 and 7\% to 13 and 10\%, mostly from a failed decrease condition or insufficient margins. Random search has much higher rejection rates, around 76 to 77\%, mostly for invalid safe regions. Every optimizer receives the failing condition and numerical hints as feedback, but these observations describe the search over time. They do not establish that feedback caused the improvement.

We can also compare the final designs with the earlier campaigns, although the different objectives and budgets prevent a direct comparison. Under the campaigns' objective, most bound-run designs certify between 56 and 80 steps, apart from the three exceptions discussed above. These bounds are higher than those obtained by synthesis at the full allowance (\NLmiEqMedian), Astra, Sol and the certificate optimizer (\NLlmAstraApiMedian, \NLlmSolMedian{} and \CoDockMedian), but ten of the eleven are lower than the Opus campaign result (\NLlmOpusMedian). The campaigns optimized $N$ directly over twenty seeds, while these studies used the padded proxy. In fact, the relationship between the two objectives is relatively weak, with Spearman correlations of $\OptSpearmanPrimary$ across the 14 main bound runs and $\OptSpearman$ across all \OptSpearmanN{} docking-bound designs, including pilots and repeats (Appendix Fig.~\ref{fig:proxystageb}).

These studies suggest that model choice can matter substantially when the search procedure lets the model adapt its strategy. With prime-agent fixed, all three newer models lower the cost on both seeds and reach similar final designs. Changing the optimizer with Sol~5.6 fixed is less consistent: the ranking changes with the seed, and computational costs differ by one to two orders of magnitude. We cannot fully separate model and effort effects, or measure their interaction with the optimizer, because model and reasoning effort change together and we tested only one row and one column of the full grid. All 28 final champions passed the exact checker and carry audited, kernel-checked Lean certificates. The studies cover docking only, not pursuit or SO(3).

\section{Verification beyond the campaigns}\label{sec:pipeline}

\subsection{A certified controller with saturating thrusters}\label{sec:sat}

For a hand-designed law at thrust limit $u_m=\SatUmax$, the completion of
\S\ref{sec:satmethod} gives $\theta=\SatTheta$ and $\rho_s=\SatRho$.
Its envelope, $c=\SatC$, covers the mission starts
($c_{\mathrm{needed}}=\SatCneeded$).

The audited file has \SatLines{} lines and \SatDecls{} declarations covering
the cell lemmas, dispatch and trajectory chain. Because decrease holds only
inside $\{V\le c\}$, the global iterate lemma does not apply.
Instead, \pcelean{sat\_trajectory} proves invariance and geometric decay
together: invariance supplies the region hypothesis needed for decay.
This yields bounded thrust and an enter-and-remain bound of \SatN{} steps.

The proof covers the true clipped dynamics.
The theorem \pcelean{saturation\_active\_witness} gives
$z=\SatWitnessZ$ with $V(z)=\SatWitnessV\le c$ and commanded thrust
$\SatWitnessU>u_m$.
The mission-start simulations never saturate in 1{,}200 steps.
Only the theorem covers off-nominal states in the certified region where
saturation is active.

At the same limit, using each design's own gain and certificate, completion
certifies \SatChampFound{} of \SatChampTotal{} cited docking champions,
each with an audited Lean file.
The optimizer is the exception: no searched sector parameter both covers the
mission starts and admits a multiplier. This is a failure to find a certificate.

The saturated settling bounds are \SatChampRatioLo--\SatChampRatioHi{} times
the linear bounds. Exact witnesses activate saturation in \SatChampWitness{}
designs. For the others, searched states do not saturate, so their certificates
do not establish that the limit is reached.

\subsection{Measurement noise and rounding}\label{sec:noise}

A flight computer measures $z + n$ and evaluates $K(z+n)$ in floating point, so
the applied thrust is $K(z+n) + e$ with $e$ the rounding error. Both errors reach
the plant through $B$, as the evader's thrust does on pursuit, and each becomes a
bounded channel of the robust decrease of \S\ref{sec:games}, with one S-procedure
multiplier per channel and every cross term kept in the exact LDL$^\top$ check.
The measurement error is bounded by $\norm{n} \le \bar n$ in the state's own
units, and $\norm{Kn} \le \sigma \bar n$ with $\sigma^2 I - K^\top K \succ 0$
checked exactly. The rounding error follows the standard model of binary
arithmetic \citep{higham2002accuracy}: on the certified invariant set,
$|e_i| \le \gamma_5 \sum_j |K_{ij}|\,|z_j + n_j|$ plus an underflow term, with
$\gamma_5 = 5u/(1-5u)$ for unit roundoff $u$.

All \NoiseTotal{} cited docking and pursuit designs remain certified with
their own gains and certificate matrices. The targets were fixed before
computation: on docking the ultimate bound must remain inside the tolerance
and the safety envelope invariant; on pursuit the radius must be at most
$1.5\,r_{\mathrm{cap}}$.

The admissible measurement-error bound is
$\bar n=\NoiseDockLo$--$\NoiseDockHi$ on docking,
or \NoiseDockPctLo--\NoiseDockPctHi\% of its tolerance.
On pursuit it is $\bar n=\NoisePursLo$--$\NoisePursHi$.
Double precision changes no bound to three figures; single precision reduces
none by more than \NoiseSingleShift\%.

At half the admissible noise bound, docking takes
\NoiseDockNRatioLo--\NoiseDockNRatioHi{} times the noise-free certified steps.
The certificate allocates contraction to disturbance rejection, using a split
chosen for the smallest ultimate bound rather than the fastest docking.
On pursuit the steps to $2\,r_{\mathrm{cap}}$ increase by at most
\NoisePursNExtra{}.

Each design's audited Lean file states the update from $A$, $B$ and $K$,
proves it equals the disturbed step, and proves decrease and trajectory bounds
for every bounded noise, rounding-error and evader-thrust sequence.
The floating-point bound is a hypothesis, not a theorem.

\subsection{The zero-order-hold plant}\label{sec:zoh}

Every certificate on the linear problems is for the Euler step. A thruster that
holds its command over the step gives instead the zero-order-hold plant
$z^+ = A_d z + B_d u$, with $A_d = e^{A_c \Delta t}$ and
$B_d = \int_0^{\Delta t} e^{A_c s}\,ds\,B_c$. It is exact at the sample instants
for the stated continuous-time model. Unlike the Euler step, it includes
the effect of thrust on position within the interval, whose leading term is
$\Delta t^2/2$ per unit acceleration. We enclose $A_d$ and $B_d$ by exact rational Taylor sums of order 22,
with every neglected entry below \ZohTail, and re-check each design's own gain and
certificate matrix against every plant in the enclosure by exact LDL$^\top$, the
perturbation entering through Young's inequality. The settings and outcomes were
fixed before the full computation.

All \ZohTotal{} cited designs retain a certificate.
Docking step counts change by factors of \ZohDockRatioLo--\ZohDockRatioHi.
Of these designs, \ZohDockSlower{} need at most \ZohDockExtraMax{} more steps;
\ZohDockFaster{} need fewer.

On pursuit, we also held the evader's thrust over each step and separated the six
tightest Euler designs (radius at most \ZohTightRcapHi) from the other six.
The tight group includes Astra, Sol, Opus and the optimizer.
Its certified radii grow by factors of \ZohTightRatioLo--\ZohTightRatioHi;
the others grow by \ZohRestRatioLo--\ZohRestRatioHi.
An Euler-shaped certificate explains some, but not all, of this growth.

An exploratory simulation, not fixed in advance, uses the search's greedy
evader and the largest distance over the last 100 of 600 steps.
In the tight group this distance grows by \ZohSimTightLo--\ZohSimTightHi{}
times, to \ZohSimTightZohLo--\ZohSimTightZohHi.
For the rest it changes by \ZohSimRestLo--\ZohSimRestHi{} times
and stays below \ZohSimRestZohHi.
Among the cited pursuit champions, the Euler leaders lose their lead on the
zero-order-hold plant. The trajectory comparison above is exploratory; it does
not establish a new paired-campaign ranking.
The certificate checks use exact rational Python arithmetic, without Lean;
the exploratory trajectories are simulations.

\subsection{Proof repair by a language model}\label{sec:tier2}

The campaign certificates all have proof templates. In the recorded Codex
CLI repair experiment, we removed the proofs of two saturated cells,
\pcelean{cell\_DD} and \pcelean{cell\_UU}, from a reduced file that retains a
complete proof of \pcelean{cell\_LL}. A language model then attempted to fill the
missing proofs, initially marked with \texttt{sorry}.

The model receives the reduced file and an author-written proof plan.
Appendix~\ref{app:lean} gives the repair protocol and recorded model settings.

The recorded comparison uses attempts with and without a supplied identity.
Without the witness, all \TierTwoAttemptsA{} attempts on \pcelean{cell\_DD}
fail. With it, both obligations close on their first attempts, taking
\TierTwoSecDD~s and \TierTwoSecUU~s.
The shared witness is an exact identity bounding the first thruster's sector
on the certified region; its longest integer has \TierTwoWitnessDigits{} digits.

This does not show that the model lacked the identity:
\TierTwoAStatingWitness{} failed attempts already state it exactly.
Attempts in both conditions also write decrease coefficients with integers
up to \TierTwoLongestDigits{} digits.
The prompts and compiler messages were not retained, so the failed step is
unknown.

These are successful reconstructions in the condition with a supplied
identity. They do not establish why the outcomes differ, estimate a general
repair success rate, or demonstrate repair beyond the template's reach.

Without a supplied plan, gpt-oss attempts \TROTotal{} obligations:
the saturated certificate's nine cells and the SO(3) decrease and torque bound.
This is a separate experiment with a different model, not a controlled test
of removing the first experiment's plan. Its obligations, conditions and
settings were fixed before any call; the protocol is in
Appendix~\ref{app:lean}.

The same acceptance gate admits all \TROTemplate{} generator proofs.
gpt-oss closes \TROClosedA{} obligations without the witness and
\TROClosedB{} with it, across \TROCalls{} calls.
An exploratory audit of the last attempts finds the complete long-integer
identity in \TROBIdentityFull{} of \TROBCells{} witnessed cell attempts;
they fail later. Without witnesses, \TROCircular{} attempts invoke the
dispatch theorem that depends on the cells themselves.
At this effort and without a plan, the model does not assemble these proofs.
A stronger prover was not tried.

\subsection{Auditing the pipeline}\label{sec:audit}

The Lean kernel checks control theorems; exact arithmetic checks linear
admissibility, step counts and SO(3) champions; floating-point statistics,
simulation and forensics produce every comparison between arms and the search scores.
Appendix Table~\ref{tab:provenance} assigns each statement to its checking layer.

Four adversarial reviews of the lower two layers, run as separate
author-orchestrated LLM-agent sessions with distinct briefs and archived
attacks, confirmed the certificate mathematics and exposed two defects.

Floating-point logarithms can undercount a bound by one step:
$\rho=99996/100000$ gives $N=10{,}000$ where exact arithmetic requires
$10{,}001$. Exact re-verification left every shipped bound unchanged
and is now required for every reported $N$.

Build-log axiom reports can be forged with \texttt{\#eval IO.println};
the original build-log audit accepted a proof of \texttt{False}. An independent
\texttt{Lean.collectAxioms} driver now checks dependencies directly.
A negative regression test injects \texttt{axiom fake : False} and requires
rejection.

We also had to consider risks from the proposer, independently of the checker. A model with file access may find earlier solutions, and coding agents have been observed trying to bypass their sandbox restrictions \citep{zelikman2024stop}. This happened in an initial GEPA pilot on the docking bound (\S\ref{sec:optimizers}). The coding-agent CLI used a read-only sandbox but inherited the repository as its working directory. The model found an archived champion from an earlier LLM-GA run and returned it as its own proposal. The proposal matched all 24 genome values and the proxy bound of 9{,}484 steps. After we hid the repository, the same seed reached 24{,}629 steps, a 2.6 times worse result. We excluded the contaminated run from the studies.

The model could read previous results without write access. An empty working directory did not take away read access either: it could still use absolute paths. For all the later runs we used an allowlist sandbox to hide the repository and the previous results. We also had to exclude another source of previous answers, namely the prompts and replies that the CLI saves from earlier sessions. We therefore isolated each run with a separate home directory. The evaluator is a separate external service, so the agent cannot manipulate the budget counter and audit trail. We also compared the submitted genomes and model replies with previous champions. This way we can detect an exact copy, but data can leak without an exact copy in the model reply. That is why we combine the comparison with sandbox isolation, rather than relying on the comparison alone.

Continuous integration builds and audits every generated certificate,
regenerates \RegenArtifacts{} of them from the archived genomes, compares their
bytes and theorem lists with a pinned manifest (\RegenTheorems{} theorems), and
re-audits every certificate the paper cites. Model-written and template-written
proofs face the same kernel check and axiom audit. Using a model to propose a
proof therefore does not require trusting the model.

\section{Discussion and limitations}\label{sec:discussion}

\paragraph{Scope of the guarantees.}
Guarantees concern rationalized controllers and certificates over exact reals
at sample instants: Euler steps at $\Delta t=1$~s on the linear problems and
Cayley steps at $h=0.1$~s on SO(3)
(Supplementary Table~\ref{tab:coverage}).

Linear measurement noise and rounding are covered only as bounded errors;
the floating-point error bound is assumed, not proved.
SO(3) noise, sensor bias and noise with thrust saturation are uncovered.
Docking champions are checked under a thrust limit only after search;
their speed-limit envelopes and the pursuit guarantee are not covered under
that limit. Pursuit bounds the evader's thrust but leaves the pursuer's unbounded.

Zero-order hold is checked at sample instants outside Lean. Among the cited
pursuit champions, the Euler ranking does not survive this change of plant
(\S\ref{sec:zoh}).
Inter-sample behavior and the validity envelope of the Clohessy--Wiltshire
linearization remain uncovered.
On the linear problems, Lean companions relate the certified updates to the
stated Euler model and gain (Appendix~\ref{app:lean}); they do not relate the discrete step to
continuous dynamics.

\paragraph{Limitations of the evidence.}
The campaigns cover three problems, only one nonlinear, with quadratic
certificates on the linear pair. Each arm has twenty runs per base instance;
the pursuit sweep and harder docking instances have ten. Spending caps leave
Astra and Fable with fewer runs on one modified instance and none on the others.

We compare the four panel configurations at one effort setting, which gives different amounts of reasoning across models. Raising effort helps one tested
model. We detect no benefit for the other. These controls do not isolate
how reasoning effort contributes to the ordering between models.
Astra and Sol were not swept over budget.
In a prompt-only probe of \ProbeCalls{} calls per setting, Astra uses a median of
\ProbeAstraHigh{} output tokens at high effort against \ProbeAstraLow{} at low;
Sol shows no detectable change (\ProbeSolHigh{} against \ProbeSolLow{}).
Closed models are identified by alias and date rather than snapshot. The optimizer studies (\S\ref{sec:optimizers}) were limited to docking. They use two seeds and only seven of the sixteen optimizer and model combinations. Since model and effort change together, we cannot isolate their effects or fully explore how they interact with the optimizer. (Also, for the bound runs we used a proxy instead of $N$, so they cannot be as easily compared to the campaigns.) We would need more seeds, matched reasoning efforts, more combinations, a repeated-sampling baseline and direct optimization of $N$ for a fuller evaluation.

Correcting the standard baselines' implementation weaknesses changes no
conclusion (Appendix Table~\ref{tab:baselines}).
The objective-aware optimizer closes most of the leading models' advantage
(\S\ref{sec:certoptres}). It remains local, and solver work is not equated with
model inference; a stronger classical search is not excluded.

Astra proposes most of its linear-problem result immediately; Sol improves
through later proposals and selection. Pursuit feedback helps Sol, but these
runs do not pin down the effect's size (\S\ref{sec:pursuit}).
The experiments neither distinguish recall of the nominal design from
reasoning nor explain the proposed Lyapunov shape.
Saturation and noise certificates were completed after search.
The recorded proof-repair attempts close two template-provable obligations
when both a plan and an exact identity are supplied. All recorded attempts
without the supplied identity fail, and the surviving records do not identify
the failed step. In a separate experiment without a supplied plan, gpt-oss
closes none of \TROTotal{} obligations (\S\ref{sec:tier2}).

On the linear problems, classical design with objective-aware certificates
reaches scores close to the best models'. Establishing an advantage beyond
these classical synthesis methods requires further evidence on nonlinear
problems.
SO(3) gives mixed evidence: no model detectably beats CMA-ES in the mixed design; Opus does
when it writes every proposal, and one run exceeds the hand-tuned law while its
median remains below it (\S\ref{sec:so3}).
Full trajectory induction, torque saturation and switching logic remain to be tested.

\paragraph{The model or the optimizer.}
The exploratory docking studies (\S\ref{sec:optimizers}) suggest that changing the model can improve results more consistently than changing the optimizer, but the setup is too limited for a general conclusion. Other studies also find that model and search procedure both matter, and that comparisons should account for computational cost \citep{zhang2024understanding,kapoor2026hal,kapoor2025aiagents}. In PCE the reliability of a result does not depend on either: the same certifying evaluator checks every candidate, however it was generated. Our isolation failure shows why this is not enough to protect the search environment. The checker can remain trustworthy while the proposer accesses earlier solutions (\S\ref{sec:audit}).

\paragraph{Implications for certification.} PCE produces explicit control laws
with checked Lyapunov certificates. The kernel establishes the stated
properties of the stated model. Comparisons between proposers still depend on
statistical analysis, and applying a controller to a vehicle requires assessing
the model's fidelity. The language model remains a design tool; its execution
is not part of the deployed controller.

\section{Conclusion}

PCE lets a language model search over controller and certificate parameters
while returning an explicit law with checked guarantees. In the docking control,
with the simulated-cost objective and mutation operator held fixed,
\GateOnCertified{} of \GateProduced{} new proposals certify when selection
retains certified candidates, compared with \GateOffCertified{} without that
rule. Subsequent certificate synthesis recovers \ResynthSdp{} of \ResynthControllers{}
controllers (Table~\ref{tab:filter}).

The comparative results depend on the problem and search configuration. On the
linear problems, the strongest models beat the standard baselines but do not
clearly outperform classical design that also optimizes the certificate. The
ranking among the cited pursuit champions does not survive zero-order-hold
dynamics. On SO(3), no model detectably beats CMA-ES in the mixed search;
Opus does when it writes every
proposal, while its median remains below the hand-tuned law.

The recorded proof-repair attempts succeed in closing two template-provable obligations
when both a plan and an exact identity are supplied; all recorded attempts
without the supplied identity fail. Some failed attempts already contain that identity. The surviving records neither identify the failed step nor explain why the outcomes differ. In another experiment without a supplied plan,
gpt-oss closes none of \TROTotal{} obligations, with or without the identity.
The exploratory docking studies show that the model can run the search as well, with the same exact checker deciding which designs are accepted. They also show why we need to isolate the search environment: a trustworthy evaluator does not prevent the model from accessing earlier solutions. We do not know yet whether model proposals improve synthesis in controller families with fewer classical methods and proof templates.

\section*{Acknowledgments}
This work is supported by the Aerospace Corporation's University Partnership
Program, and partially by the Spanish Agencia Estatal de Investigación (AEI)
under grant \mbox{PID2024-161963OB-C22} (ACTIVATION project) and by the Madrid
Government (Comunidad de Madrid-Spain) under the Multiannual Agreement
2023-2026 with Universidad Politécnica de Madrid in the Line A, Emerging PhD
researchers.

\bibliographystyle{plainnat}
\bibliography{refs}

\clearpage
\appendix
\makeatletter
\let\pce@seccntformat\@seccntformat
\newcommand{\pce@prefix@section}{Appendix~}
\renewcommand{\@seccntformat}[1]{%
  \ifcsname pce@prefix@#1\endcsname
    \csname pce@prefix@#1\endcsname
  \fi
  \pce@seccntformat{#1}}
\makeatother
\section{Problem and search specification}\label{app:spec}

This appendix specifies the dynamics, parameter bounds, initialization,
proposal rules and certificate checks used in the campaigns. These settings
are taken from the implementation; Table~\ref{tab:settings} gives the campaign
budgets. The certificate optimizer is described separately at the end.

\subsection{Dynamics and controller parameters}

\paragraph{Linear dynamics.} The state $z = (x, y, v_x, v_y)$ is in metres and
metres per second, and one Euler step of $\Delta t = 1$~s of the
Clohessy--Wiltshire equations with $n = 1.13\times10^{-3}$~rad/s gives
$z^+ = Az + Bu$ with $A = I + \Delta t\,A_c$,
\[
A_c = \begin{pmatrix} 0&0&1&0\\ 0&0&0&1\\ 3n^2&0&0&2n\\ 0&0&-2n&0 \end{pmatrix},
\qquad B = \Delta t\begin{pmatrix} 0&0\\ 0&0\\ 1&0\\ 0&1 \end{pmatrix},
\]
and $u = Kz$, unbounded except in \S\ref{sec:sat}. The mission starts are
$(1, 0.5, 0, 0)$, $(-0.8, 1, 0, 0)$ and $(0.3, -1.2, 0.01, -0.01)$; the speed
limit has $\nu_0 = 0.2$ and $\nu_1 = 0.5$, the docking tolerance is
$\norm{z} \le 0.02$, and $K$ and $P$ are rationalized to fractions with denominators at most $10^4$. On
pursuit the evader's thrust is at most $0.01$~m/s$^2$ and the capture statement is at
$1.5\,r_{\mathrm{cap}}$.

\paragraph{Attitude dynamics.} The inertia is $J = \operatorname{diag}(4.5, 5,
5.5)$ and $h = 0.1$~s. Both updates of a step use the current state:
$R^+ = R\,(aI + h\hat\omega + b\,\omega\omega^\top)/d$ with
$a = 1 - h^2\norm{\omega}^2/4$, $b = h^2/2$ and $d = 1 + h^2\norm{\omega}^2/4$,
which is $R$ times the Cayley map of $h\omega$, and
$\omega^+ = \omega + hJ^{-1}\bigl((J\omega)\times\omega + u\bigr)$, where the
feed-forward term of $u$ cancels the gyroscopic term. The torque bound of
Fig.~\ref{fig:so3cert} is a certified output, the maximum of $\norm{u}^2$ on
$\{V \le c\}$ from a second linear program, not an imposed limit.

\paragraph{Genomes.} On the linear problems $K \in \mathbb{R}^{2\times4}$ has all
entries free and $L$ is lower triangular; every proposed genome is clamped to
$|K_{ij}| \le 5$, $|L_{ij}| \le 50$ and $L_{ii} \ge 10^{-3}$, so $P = LL^\top$ is
positive definite. On SO(3) the genome is the fifteen diagonal gains of
$\widehat G, G_c, M, C, K_\omega$ as positive rationals, rounded to denominators of
at most 1000; a proposal with a non-positive gain is rejected, and there are no
other bounds.

\subsection{Initialization, selection and proposals}

\paragraph{Seeded populations.} Each docking draw chooses one of the following:
\begin{itemize}
\item With probability 0.4, a gentle PD law with
  $k_p\sim U(0.001,0.02)$ and $k_d\sim U(0.03,0.25)$. Its certificate comes from
  the Lyapunov equation with $Q=\operatorname{diag}(q,q,1,1)$ and
  $q\sim U(0.001,0.1)$.
\item With probability 0.3, a perturbation at scale 0.05 of LQR with $Q=R=I$.
\item With probability 0.3, a random gain $K\sim\mathcal N(0,0.3^2)$ that
  passes tier~0, with a Lyapunov certificate for $Q=I$.
\end{itemize}
Failed draws are discarded and replaced, so the proportions in the accepted
population can differ from the draw probabilities.

Pursuit starts from PD laws with $k_p\sim U(0.05,0.35)$,
$k_d\sim U(0.3,1.3)$ and Lyapunov certificates for $q\sim U(0.05,1.5)$.
Half the designs are then perturbed at scale 0.02.

SO(3) starts from isotropic geometric PD laws:
$\widehat G=G_c=k_pI$, $M=J$, $C=\varepsilon I$ and $K_\omega=k_dJ$, with
$k_p\sim U(0.6,1.4)$, $k_d\sim U(0.5,2)$ and
$\varepsilon\sim U(0.12,0.35)$. These are the deliberately weak starts
of \S\ref{sec:so3}.

\paragraph{Selection and mutation.} On the linear problems, the elite contains
six genomes. Runs with verification first retain the three best certified
candidates, then fill the remaining places by score. Under the certified
objective, every certified genome ranks above every uncertified one. Parents
are drawn uniformly from this elite. The next population contains the elite,
the model's proposals, and enough Gaussian children to reach 28 members.

Each Gaussian child uses one of two operations, with equal probability:
\begin{itemize}
\item Perturb $K$ at scale $s_g$ and perturb $L$ at a relative scale.
\item Perturb $K$ at the same scale and solve the discrete Lyapunov equation
  with $Q = \operatorname{diag}(q,q,1,1)$ and
  $\log_{10}q \sim U(-3.5,-0.5)$. If the modified loop is unstable, retain the
  parent's $L$.
\end{itemize}
Both operations also zero one randomly chosen gain with probability 0.15.
The mutation scale at generation $g$ is $s_g = 0.06\cdot0.88^g$ on docking and
$0.05\cdot0.9^g$ on pursuit.

On SO(3), selection retains the twelve best certified angles from parents and
children together; uncertified genomes score zero. The model sees the best
four candidates. Gaussian mutation multiplies each gain by
$\exp(\mathcal N(0,\sigma^2))$, with $\sigma=0.25\cdot0.92^g$.

\paragraph{Model calls and the ledger.} The model is called at generations
$0, 3, \dots, 21$ on the linear problems and $3, 6, 9$ on SO(3), for eight
candidates each time, and is shown the elite; a failed call is logged and
replaced by Gaussian children. A proposal is a distinct genome; repeats are free
and tier-0 rejects are charged. On the linear problems the loop arms offer at
most 534 distinct genomes: the initial population contributes 28, and each
of the 23 later generations adds at most 22, giving $23\cdot22=506$ new
genomes and $28+506=534$ in total. Each classical arm receives the Gaussian
arm's spend on the same seed as its allowance. On SO(3), the allowance of 120
is enforced.

\subsection{Certificate checks and objectives}

\paragraph{Docking.}
On docking, exact $LDL^\top$ bisection finds $\alpha$, $\beta$ and
$\varepsilon$. The same test finds $\rho$, bracketed by the largest
floating-point generalized eigenvalue; $1-\rho$ is rounded down to nine
significant figures and confirmed exactly.
The safety multiplier starts at 1.05 times its floating-point estimate,
rounded up on a $10^{-6}$ grid. Up to four exact checks are tried,
multiplying by $6/5$ after each failure.
The docking bound is the smallest $n$ with
$\rho^n c_{\mathrm{needed}}\le\alpha(0.02)^2$, verified exactly.

\paragraph{Pursuit.} A 33-point scan chooses $\rho_r$ to minimize $C_{\mathrm{cap}}$.
Exact bisection on the S-procedure matrix finds $\mu$, and
$r_{\mathrm{cap}}=\sqrt{C_{\mathrm{cap}}/\alpha}$.

\paragraph{SO(3).} The region level $c$ starts at $1/4$ and doubles up to eight times while the linear
program certifies $\rho\le0.995$, followed by seven bisection steps.
CVXPY with Clarabel uses a Gram dominance margin of $10^{-3}$ and two scaling
rounds. Champions are re-verified at denominators $10^7$ and $10^9$, requiring
the exact identity, exact diagonal dominance and $\rho\le0.995$.
For Lean, the decrease program is re-solved with a sparsity penalty to shorten
the file. Sol's model-only champion uses the unpenalized solution because
its penalized exact rate exceeded $0.995$.

\paragraph{Simulated cost.} The cost of \S\ref{sec:filter} is, over 1500 steps
from each mission start,
\[
J = \sum_{z_0} \Bigl[\,\sum_{t=0}^{1499} \bigl(u_t^\top u_t + 200\max(0, -h(z_{t+1}))\bigr)
+ 0.02\,t_s + 100\,\norm{z_{1500}}\Bigr],
\]
where $h(z) = \nu_0^2 + \nu_1^2\norm{r}^2 - \norm{v}^2$ and $t_s$ is the smallest
$t$ with $\norm{z_{t+1}} < 0.02$ (1500 if there is none).

\subsection{Classical certificate optimizer} The PD lattices are $9 \times 9$ on docking
($k_p \in [10^{-3}, 10^{-1}]$, $k_d \in [0.03, 1]$, each with and without the
Coriolis cancellation) and $13 \times 13$ on pursuit ($k_p \in [0.02, 2]$,
$k_d \in [0.1, 4]$), plus the gains of the seeded pool. Each certificate step
tries twelve rates per gain, $1 - \rho$ geometric from
$0.98\,(1 - \rho(A_{\mathrm{cl}})^2)$ down to $10^{-4}$, and the joint program
on pursuit uses 33 rates with $1 - \rho$ geometric from $0.98$ to $10^{-4}$. The
alternation runs ten rounds with restarts at scales 0.05 and 0.2, up to
\CoSdpCap{} solver calls (Clarabel, with SCS as fallback). Each fixed-rate
program imposes the contraction at $\rho' = \rho - 10^{-3}(1 - \rho)$, and on
docking it fixes the S-procedure multiplier at
$\lambda_0 = \nu_0^2 (1 - 10^{-3})/1.05$, so that the exact completion, which
pads the multiplier by 1.05, accepts the result.

\section{Formal checking in Lean}\label{app:lean}

For each cited design the certificate compiler writes a Lean~4 file
\citep{demoura2021lean4} against mathlib \citep{mathlib2020}. On the linear
problems each certificate condition becomes one polynomial identity, closed by
\pcelean{ring}, a statement that the squares of the $LDL^\top$ decomposition are
nonnegative, and a chain of inequalities closed by \pcelean{linarith}. The exact
rational coefficients appear in the file as literal data, so the only search left
to \pcelean{linarith} is for a linear combination of facts the file supplies. No
step requires a language model or an external solver. The templates supply
the identities and inequalities that the tactics discharge. Elaboration
turns these tactic scripts into proof terms, and the Lean kernel checks the
terms; the mathlib lemmas they use are kernel-checked in the same way. The
compiler and the tactics are therefore outside the trust base.

The docking and pursuit files state the closed-loop update using literal
rational coefficients. These coefficients already combine the plant, time step
and gain, so checking the certificate alone would not establish that the
compiler used the intended model.

Each cited file therefore has a companion proof that states the mean motion,
thrust gains, time step $\Delta t$ and gain $K$ separately. For every state
component, it proves that the certified update equals
$z_i+\Delta t\,f_i(z,Kz)$, adding $\Delta t\,b_iw_i$ on pursuit, where $f$ is the
continuous-time Clohessy--Wiltshire field. The tactic \pcelean{ring} checks
these identities. An inconsistency between the stated model and the
certificate's coefficients fails the build. The companions are generated from
the same run records and undergo the same audit as the certificate files.

An independent driver then calls \texttt{Lean.collectAxioms} on every declared
theorem and requires its axiom footprint to lie within \texttt{[propext,
Classical.choice, Quot.sound]}, the three axioms of Lean's standard foundations.
Any other axiom is rejected, including \texttt{sorryAx}, which marks an
unfinished proof, the axiom introduced by native evaluation, and any axiom the
file declares. Section~\ref{sec:audit} describes why the audit calls Lean
directly rather than reading the build log. What the kernel does not check is
whether the theorem states the intended property of the intended model: the
theorem statements and the model assumptions require review
(\S\ref{sec:discussion}).

\paragraph{Proof-repair protocols.}
The repair loop uses the Codex CLI coding agent at low reasoning effort; its
underlying model version was not recorded. The model proposes a tactic block,
\texttt{lake build} checks it, and compiler errors are returned for the next
attempt. Success requires the project to compile, the number of open
obligations to decrease, and the proposed block to contain no \texttt{sorry}.
These obligations are already known to be provable by template. The model sees
the reduced file and a proof plan written by the authors, so the experiment
tests assembly of a supplied plan.

The separate gpt-oss experiment uses the following protocol.
Each file retains definitions and statements but no proofs, preventing copying
from neighbouring proofs or earlier repairs.
Each obligation is attempted from the same file with and without its key
identity from the generator.
The model uses the panel's low effort and \TRORounds{} feedback rounds.
Obligations, conditions and settings were fixed before any call.

\section{Artifact provenance}\label{app:artifacts}

The paper uses separate checks for formal guarantees, numerical certificate
verification and empirical comparisons. Table~\ref{tab:provenance} identifies
the check behind each type of statement. A kernel-checked control theorem
does not verify a campaign statistic or a simulated trajectory.

The command \texttt{python -m tools.paper\_tables --check} checks the tables
and numerical macros in \path{paper/latex/tables/} against their recorded
sources. The source paths are listed in \path{tools/paper_tables.py}, and
\path{paper/latex/tables/MANIFEST.json} records source and output hashes.
Figures and their inputs are recorded in \path{paper/figures/MANIFEST.json}
and checked by \texttt{python -m tools.paper\_figures --check}.

\begin{table}[h]
\centering
\small
\begin{tabular}{@{}p{0.62\textwidth}p{0.32\textwidth}@{}}
\toprule
\textbf{Statement class} & \textbf{Checked by}\\
\midrule
Per-candidate control theorems: stability, safety envelope,
\pcelean{safe\_forever}, docking within $N$; the strategy-quantified pursuit
bound and trajectory-level capture; the nine-cell saturated decrease and its
trajectory chain; the SO(3) step and regional decrease; template- or
LLM-authored alike & \textbf{Lean 4 kernel}; footprint by an independent
\texttt{Lean.collectAxioms} driver, a subset of \texttt{[propext,
Classical.choice, Quot.sound]}\\[4pt]
In-loop admissibility on the linear problems; step counts $N$; certificate
completions; the stage
rescoring of Table~\ref{tab:stages}; the re-verification of every SO(3) champion
and the positive definiteness of its $V$ & \textbf{Exact rational arithmetic}
(Python \texttt{Fraction})\\[4pt]
Campaign medians, paired differences, bootstrap intervals, Wilcoxon and Holm
$p$ & \textbf{Float statistics}, recomputed from the run summaries
before every release\\[4pt]
Fitness, the SO(3) in-loop scoring, trajectories, attack results,
session-log audits & \textbf{Float
simulation and forensics} (unverified)\\
\bottomrule
\end{tabular}
\caption{Which part of the pipeline checks each kind of statement.}
\label{tab:provenance}
\end{table}

\paragraph{Certificates.}
We cite \CitedFiles{} Lean files carrying \CitedTheorems{} theorems.
The independent \texttt{collectAxioms} audit passed \CitedAudited{} files.
The optimizer studies are not included in these counts. They add \OptAuditedFiles{} champion files and \OptAuditedTheorems{} theorems from the 28 main runs, four supervised runs, six additional Opus runs and six GEPA pilots, two of them on pursuit. All these files passed the same kernel check and axiom audit.
The archive contains \ArchiveEntries{} earlier champions with audited
certificates, carrying \ArchiveTheorems{} theorems.
Its \ArchiveReconstructed{} genome reconstructions are marked as such.
All paper quantities are generated from run records; the release gate rejects
the record, table and PDF inconsistencies it checks.

\section{Additional results}\label{app:more}

\paragraph{Interface-control protocol.} Astra's CLI provides a read-only shell.
In the execution-enabled variant, the model ran Python in \LlmSessionsWithCode{}
of \LlmSessions{} sessions, including \LlmDiffEvoMentions{} commands invoking
\texttt{scipy}'s differential evolution. This computation was not charged to
the proposal ledger. With execution disabled, live tool use was
\LlmTextLive{} of \LlmTextCalls{} calls, verified from the event stream.

Opus's CLI ran in an empty directory with no tools, settings or stored
session and used the required
system line, \emph{Answer the user's request.}

\begin{figure}[H]
\centering
\includegraphics[width=\textwidth,height=0.70\textheight,keepaspectratio]{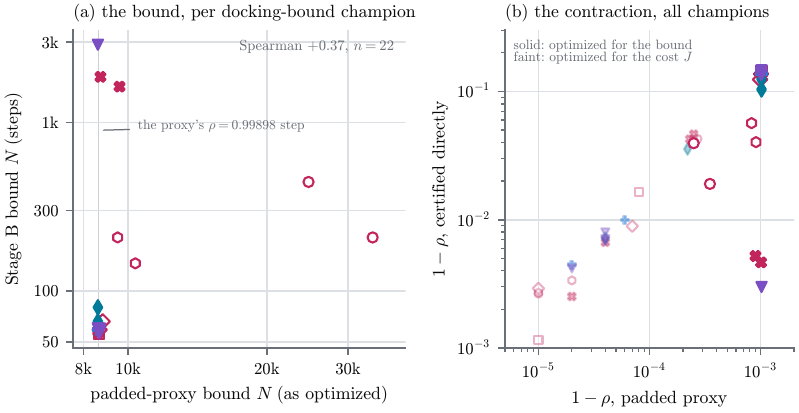}
\caption{Padded proxy versus certified bound (a) and contraction factors (b) for optimizer-study designs.}
\label{fig:proxystageb}
\end{figure}

\paragraph{Choice of the docking objective.}
Preliminary runs used the proxy $\rho=1-\varepsilon/\beta$ with fixed
floating-point padding, which can reverse the ranking obtained by directly certifying contraction
(Table~\ref{tab:stages}).
Exact bisection within the proxy formulation (Stage~A) rescales the bounds
almost uniformly by \StageARescale$\times$.
Direct certification (Stage~B) tightens them by a further
\StageBFormRatioMin--\StageBFormRatioMax$\times$ and reverses their order:
the Gaussian design ranked last by the proxy beats its leading LLM design by
\StageBRatioSoneLlm$\times$.
Every reported campaign optimizes Stage~B; proxy rankings are not used. We used the padded proxy for the bound runs in the optimizer studies (\S\ref{sec:optimizers}). Figure~\ref{fig:proxystageb} compares its results with Stage~B across \OptSpearmanN{} docking-bound designs, including pilots and repeats.

\begin{table}[h]
\centering\footnotesize
\begin{adjustbox}{max width=\textwidth}
\begin{tabular}{@{}lrrrrrrrr@{}}
\toprule
Design & $\rho(A_{cl})$ & padded $N$ & A $N$ & B $N$ & padded/A & A/B & padded $\rho$ & B $\rho$ \\
\midrule
Gaussian mutation, seed 0 & 0.959 & 10{,}335 & 4{,}595 & 656 & 2.25 & 7.0 & 0.99913 & 0.98652 \\
Gaussian mutation, seed 1 & 0.929 & 13{,}191 & 5{,}907 & 181 & 2.23 & 32.6 & 0.99932 & 0.95215 \\
Astra (preliminary run), seed 0 & 0.951 & 9{,}484 & 4{,}222 & 535 & 2.25 & 7.9 & 0.99905 & 0.98349 \\
\bottomrule
\end{tabular}
\end{adjustbox}
\caption{Three designs from preliminary runs, scored by the padded proxy,
by the same proxy with exact bisection (Stage~A), and with the contraction
certified directly (Stage~B), the objective used throughout.}
\label{tab:stages}
\end{table}

\paragraph{Comparison families.}
Docking compares each panel arm with full-allowance synthesis:
\DockFamily{} comparisons, fixed after the docking results.

Pursuit includes pairs involving a classical baseline, feedback controls
against their arms, and one model pair, with a separate family per budget.
Of \PCFamilyRecorded{} comparisons, \PCFamilyUnreported{} involve unreported
models. Restricting Holm adjustment to the \PCFamilyReported{} reported
pairs (Table~\ref{tab:supppursuit} and the gpt-oss feedback control)
changes no decision at $p=0.05$.
Including unreported pairs can only increase adjusted $p$.

SO(3) uses all pairs of the six campaign arms and a separate model-only family,
both fixed in code before model calls. The API Opus arm occupies the campaign
family under a rule fixed before its calls; model-only Opus is compared with
its CLI mixed arm. Harder docking instances each have their own family.

Docking and pursuit families are outcome-informed and exploratory.
Adjustment does not make them confirmatory.

\paragraph{Exclusion of degenerate runs.}
Runs losing more than a quarter of model calls are excluded and archived.
The rule preceded all five exclusions: one pursuit run, two SO(3) runs and
two harder-instance runs.
Every retained run lost at most a quarter; every excluded run lost at least
a third. Any threshold from 25\% up to, but excluding, one third gives the
same exclusions.

Sensitivity checks recompute every comparison and family.
At 20\%, \ExclStrictExtra{} additional runs losing exactly two of eight calls
are excluded, affecting \ExclStrictChanged{} of \ExclComparisons{} comparisons.
No median difference changes sign.
Of \ExclStrictCrossings{} comparisons moving above $p=0.05$,
\ExclStrictInstanceCrossings{} belong to the ten-seed instance families.
With no exclusions, no median paired difference changes sign and one
comparison moves below the
threshold: high-effort Fable against Gaussian mutation on unequal thrust.
None is interpreted as a detected difference.

\paragraph{Transfer of existing designs.}
Table~\ref{tab:transfer} re-certifies existing designs without further search.
It reports certified seeds and median best bounds for the initial population,
the best of \GridSize{} PD-grid designs with textbook repair, and certification
counts and median bounds for archived base champions.
The final rows give settings for which the initialization procedure found
a certified starting population on only one of ten seeds.

\begin{table}[h]
\centering\footnotesize\setlength{\tabcolsep}{3.5pt}
\begin{adjustbox}{max width=\textwidth}
\begin{tabular}{@{}lrrrrrrrrr@{}}
\toprule
 & \multicolumn{2}{c}{initial population} & PD grid & \multicolumn{2}{c}{synthesis, full allowance} & \multicolumn{2}{c}{Astra (text-only)} & \multicolumn{2}{c}{Fable (effort high)} \\
problem & certified & median $N$ & best $N$ & certified & median & certified & median & certified & median \\
\midrule
base problem & 10/10 & 194.5 & 133 & 20/20 & 52.0 & 20/20 & 38.5 & 20/20 & 39.0 \\
geostationary orbit & 10/10 & 194.5 & 133 & 20/20 & 52.0 & 20/20 & 39.0 & 20/20 & 38.0 \\
unequal thrusters & 10/10 & 675.5 & 287 & 20/20 & 54.0 & 1/20 & 43.0 & 4/20 & 51.5 \\
tighter speed limit & 8/10 & 1005.5 & 192 & 0/20 & -- & 0/20 & -- & 0/20 & -- \\
wider initial set & 10/10 & 610.0 & 178 & 0/20 & -- & 0/20 & -- & 0/20 & -- \\
speed limit halved & 1/10 & 5560.0 & 324 & 0/20 & -- & 0/20 & -- & 0/20 & -- \\
initial set doubled & 1/10 & 6437.0 & 377 & 0/20 & -- & 0/20 & -- & 0/20 & -- \\
\bottomrule
\end{tabular}
\end{adjustbox}
\caption{Existing designs re-checked on each modified docking problem, before any search on it.}
\label{tab:transfer}
\end{table}

\paragraph{Structure of the docking champions.}
Astra approximately cancels Coriolis coupling in \AstraApiCoriolisSeeds{} of
\SeedsLlmAstraApi{} runs. The Fig.~\ref{fig:champion} gain is
$K_{14}=-K_{23}=\FigChampKzeroThree$, close to $\mp2n=\mp\TwoN$;
the rationalized residual is $\CoriolisResidual$.
This decouples the loop into two nearly identical second-order systems.
No Opus or gpt-oss champion has the cancellation.

Table~\ref{tab:coriolis} tests its value under Stage~B by removing it from
Astra designs, varying it on a fixed PD grid, and fixing it in Gaussian
mutation. Removing it changes the median by about one step.
The grid gives the same best bound with or without cancellation, both with
Lyapunov repair and with a champion's fixed $L$.
Fixing cancellation in Gaussian search detects no difference.

The Lyapunov cross-term ratio tracks the docking bound in the rescored
champions (Table~\ref{tab:lyapunovratios}). This association is exploratory,
mixes arm and seed effects, and is not a causal measurement. The rescore
excludes Sol and grid-bounded synthesis.

\begin{table}[h]
\centering\small
\begin{tabular}{@{}lr@{}}
\toprule
Population or statistic & Value\\
\midrule
Seeded population: cross-term ratio & \LRatioSeed\\
Gaussian champions: cross-term ratio & \LRatioGauss\\
Synthesis champions: cross-term ratio & \LRatioLmiEq\\
gpt-oss champions: cross-term ratio & \LRatioGptoss\\
Opus champions: cross-term ratio & \LRatioOpus\\
Astra champions: cross-term ratio & \LRatioAstraApi\\
Rescored champions & \LRatioSpearmanN\\
Rescored arms, including controls & \LRatioArms\\
Spearman correlation with the bound & $\LRatioSpearman$\\
Champions with $N\le\TightBound$ & \TightCount\\
Cross-term ratio of those champions & $\TightRatioLo$--$\TightRatioHi$\\
\bottomrule
\end{tabular}
\caption{Ratio of the cross terms of $L$ to its position weights.}
\label{tab:lyapunovratios}
\end{table}

\begin{table}[h]
\centering\scriptsize\setlength{\tabcolsep}{3pt}
\begin{adjustbox}{max width=\textwidth}
\begin{tabular}{@{}lrrrrp{0.25\textwidth}@{}}
\toprule
Design & Designs & Certified & Best $N$ & Median $N$ & Note \\
\midrule
Astra designs, as found & 21 & 21 & 37 & 38 & Astra's 20 text-only designs and one coding-agent design, re-certified \\
\quad Coriolis gains set to zero & 21 & 21 & 37 & 39 &  \\
\quad Coriolis gains sign-flipped & 21 & 21 & 37 & 40 &  \\
\midrule
PD grid, no Coriolis term & 1{,}512 & 523 & 140 & 838 & certificate from the Lyapunov equation \\
PD grid, no Coriolis term & 378 & 71 & 39 & 147 & certificate factor $L$ of one Astra design \\
PD grid, $\mp 2n$ (cancellation) & 1{,}512 & 523 & 140 & 838 & certificate from the Lyapunov equation \\
PD grid, $\mp 2n$ (cancellation) & 378 & 71 & 39 & 147 & certificate factor $L$ of one Astra design \\
PD grid, $\pm 2n$ (flipped) & 1{,}508 & 523 & 140 & 839 & certificate from the Lyapunov equation \\
PD grid, $\pm 2n$ (flipped) & 377 & 71 & 39 & 147 & certificate factor $L$ of one Astra design \\
\midrule
Gaussian mutation, cancelling gains fixed (20 seeds) & 20 & 20 & 93 & 154.5 & the other six gains searched; vs.\ Gaussian mutation: median $\Delta N = +0.0$, ahead on 9, level on 4 of 20; vs.\ Astra (text-only): $\Delta N = +116.5$, ahead on 0/20 \\
\bottomrule
\end{tabular}
\end{adjustbox}
\caption{The Coriolis-cancelling gains of Astra's docking designs, measured three
ways: removed from the designs, added to a PD grid, and fixed in a Gaussian
search.}
\label{tab:coriolis}
\end{table}

\paragraph{Baselines rerun.} Table~\ref{tab:baselines} compares each revised
classical implementation with its original version. It reports the median
certified docking bound or pursuit radius and the number of seeds that certify.
The ``vs recorded'' columns count seeds where the revised version is better,
worse or tied with the original. The ``vs'' columns compare it with the named
reference arm. CMA-ES step scalings $0.1$ and $0.001$ are sensitivity checks on
seeds 0--9.

\begin{table}[h]
\centering\scriptsize\setlength{\tabcolsep}{3pt}
\begin{adjustbox}{max width=\textwidth}
\begin{tabular}{@{}llrrrcc@{}}
\toprule
Baseline & Change & $n$ & original & rerun & vs original & vs reference \\
\midrule
synthesis, unequal thrusters & keeps certified starting designs & 10 & 8/10, 51.0 & 10/10, 55.0 & 0 / 0 / 8 & Fable (high): 0 / 8 / 0 \\
\midrule
CMA-ES, docking & restarts, growing population (IPOP) & 20 & 20/20, 182.0 & 20/20, 182.0 & 0 / 0 / 20 & synthesis: 0 / 20 / 0 \\
 & BIPOP, step $\times 0.01$ & 20 & 20/20, 182.0 & 20/20, 121.0 & 20 / 0 / 0 & synthesis: 0 / 20 / 0 \\
 & BIPOP, step $\times 0.1$ & 10 & 10/10, 194.5 & 10/10, 165.5 & 6 / 0 / 4 & synthesis: 0 / 10 / 0 \\
 & BIPOP, step $\times 0.001$ & 10 & 10/10, 194.5 & 10/10, 135.0 & 10 / 0 / 0 & synthesis: 0 / 10 / 0 \\
\midrule
PD grid, pursuit & one refined lattice, $\times1$ & 10 & 10/10, 0.03388 & 10/10, 0.03388 & 0 / 0 / 10 & CMA-ES: 10 / 0 / 0 \\
 & one refined lattice, $\times2$ & 10 & 10/10, 0.03879 & 10/10, 0.03388 & 10 / 0 / 0 & CMA-ES: 6 / 4 / 0 \\
 & one refined lattice, $\times4$ & 10 & 10/10, 0.03563 & 10/10, 0.03388 & 10 / 0 / 0 & CMA-ES: 0 / 10 / 0 \\
\bottomrule
\end{tabular}
\end{adjustbox}
\caption{Each classical baseline rerun with its implementation weakness fixed, beside the original run.}
\label{tab:baselines}
\end{table}

\paragraph{Optimizer studies: further details.}
The optimizer configurations are given in Table~\ref{tab:optconfig}. All approaches submit genomes to the same external evaluator, but differ in how they organize the search and use model calls. The table records the engine versions, what each model call does and the stopping limits.

\begin{table}[!htbp]
\centering\footnotesize
\setlength{\tabcolsep}{4pt}\begin{tabular}{@{}l>{\raggedright\arraybackslash}p{0.24\textwidth}>{\raggedright\arraybackslash}p{0.31\textwidth}>{\raggedright\arraybackslash}p{0.21\textwidth}@{}}
\toprule
optimizer & implementation & per model call & stops at \\
\midrule
GOA GEPA & GEPA 0.1.4 (PyPI), \texttt{gepa} engine & one reflection: reads a candidate and its feedback, writes one revision & 250 reflections or 560 evaluations \\
GOA Meta-Harness & GEPA GitHub commit \texttt{48f1f65d}, \texttt{meta\_harness} engine & one agent session: reads all earlier candidates, may run code, writes up to 10 & 60 sessions or 560 evaluations \\
GOA Omni & same commit, \texttt{optimize\_best\_of} then \texttt{optimize\_sequential} & as the engine running (GEPA, Meta-Harness, AutoResearch) & $3 \times 62$, then $2 \times 187$ evaluations; unused allowances are not passed on \\
prime-agent & prime-agent 0.9.7, autonomous mode & one agent turn in a persistent Python kernel, often submitting tens of genomes by script & 560 evaluations, or a time or spending limit \\
\bottomrule
\end{tabular}
\caption{Optimizer versions, model interactions and stopping limits; all use the same evaluator and 560-evaluation cap.}
\label{tab:optconfig}
\end{table}

Table~\ref{tab:optall} gives results and resources by arm; Table~\ref{tab:optruns} gives each run's evaluation count, stopping reason and final result on the two axes. We include these details because not every optimizer reached the full evaluation budget, and using the same number of evaluations does not mean using the same inference cost.

\begin{table}[!htbp]
\centering
\begin{adjustbox}{max width=\textwidth}\footnotesize\setlength{\tabcolsep}{4pt}
\begin{tabular}{@{}lrrrrrrrrr@{}}
\toprule
 & \multicolumn{2}{c}{docking cost $J$} & \multicolumn{2}{c}{docking bound $N$} & \multicolumn{5}{c}{per run (median)} \\
\cmidrule(lr){2-3}\cmidrule(lr){4-5}\cmidrule(l){6-10}
 & seed 0 & seed 1 & seed 0 & seed 1 & tokens & price (\$) & run (min) & first better (min) & model calls \\
\midrule
\multicolumn{10}{@{}l}{\emph{optimizer axis}} \\
\quad GOA GEPA + Sol 5.6 & 1.993 & 1.052 & 146 & 208 & 7.7M & 12 & 149 & 1.2 & 250 \\
\quad GOA Meta-Harness + Sol 5.6 & 1.227 & 1.033 & \textbf{56} & \textbf{56} & 48.4M & 46 & 285 & 4.5 & 60 \\
\quad GOA Omni + Sol 5.6 & 1.016 & 1.034 & 66 & 59 & 23.4M & 21 & 226 & 0.7 & 118 \\
\quad prime-agent + Sol 5.6 (Sol baseline) & 0.810 & 1.326 & 57 & 58 & 1.1M & 1.14 & 9.6 & 0.7 & 44 \\
\addlinespace[2pt]
\multicolumn{10}{@{}l}{\emph{model axis}} \\
\quad prime-agent + Astra & \textbf{0.729} & \textbf{0.742} & 59 & 59 & 2.6M & 5.32 & 14 & 1.2 & 60 \\
\quad prime-agent + gpt-6-sol & 0.762 & 0.759 & 80 & 66 & 2.9M & 1.17 & 21 & 0.8 & 66 \\
\quad prime-agent + Opus & 0.760 & 0.758 & 2{,}896$^{\dagger}$ & 60 & 5.2M & 5.13 & 212 & 2.2 & 112 \\
\addlinespace[2pt]
\multicolumn{10}{@{}l}{\emph{other runs}} \\
\quad prime-agent supervised + Sol 5.6 & 1.111 & 0.810 & 1{,}633 & 1{,}866 & 622k & 0.72 & 5.0 & 0.7 & 28 \\
\addlinespace[2pt]
\multicolumn{10}{@{}l}{\emph{references}} \\
\quad starting design & 4.365 & 5.044 & 231 & 477 &  &  &  &  &  \\
\quad random search & 2.321 & 1.882 & 148 & 133 & 0 & 0 & 0.3 & $<$0.1 & 0 \\
\quad GA & 3.024 & 2.361 & 656 & 181 & 0 & 0 & 0.1 & -- & 0 \\
\bottomrule
\end{tabular}
\end{adjustbox}
\caption{Final docking cost $J$ at 491 evaluations, bound $N$, and median resources per arm; $^{\dagger}$Opus run stopped at 43 evaluations.}
\label{tab:optall}
\end{table}

\begin{table}[!htbp]
\centering
\begin{adjustbox}{max width=\textwidth}\footnotesize\setlength{\tabcolsep}{4pt}
\begin{tabular}{@{}llrlrr@{}}
\toprule
cell & objective & evaluations & stopped by & seed 0 & seed 1 \\
\midrule
GOA GEPA, Sol 5.6 & cost & 517 / 537 & 250 reflections & 1.993 / 1.993 & 1.052 / 1.052 \\
GOA GEPA, Sol 5.6 & bound & 510 / 513 & 250 reflections & 10{,}363 / 146 & 9{,}487 / 208 \\
\addlinespace[1pt]
GOA Meta-Harness, Sol 5.6 & cost & 552 / 532 & 60 sessions & 1.227 / 1.227 & 1.033 / 1.033 \\
GOA Meta-Harness, Sol 5.6 & bound & 460 / 433 & 60 sessions & 8{,}639 / 56 & 8{,}629 / 56 \\
\addlinespace[1pt]
GOA Omni, Sol 5.6 & cost & 491 / 517 & phase allowances & 1.016 / 1.016 & 1.034 / 1.034 \\
GOA Omni, Sol 5.6 & bound & 500 / 539 & phase allowances & 8{,}806 / 66 & 8{,}673 / 59 \\
\addlinespace[1pt]
prime-agent, Sol 5.6 & cost & 560 / 560 & 560 evaluations & 0.810 / 0.810 & 1.326 / 1.208 \\
prime-agent, Sol 5.6 & bound & 560 / 560 & 560 evaluations & 8{,}632 / 57 & 8{,}605 / 58 \\
\addlinespace[1pt]
prime-agent, Astra & cost & 560 / 560 & 560 evaluations & 0.729 / 0.728 & 0.742 / 0.741 \\
prime-agent, Astra & bound & 560 / 560 & 560 evaluations & 8{,}593 / 59 & 8{,}594 / 59 \\
\addlinespace[1pt]
prime-agent, gpt-6-sol & cost & 560 / 560 & 560 evaluations & 0.762 / 0.760 & 0.759 / 0.758 \\
prime-agent, gpt-6-sol & bound & 560 / 560 & 560 evaluations & 8{,}597 / 80 & 8{,}594 / 66 \\
\addlinespace[1pt]
prime-agent, Opus & cost & 560 / 560 & 560 evaluations & 0.760 / 0.760 & 0.758 / 0.757 \\
prime-agent, Opus & bound & 43 / 560 & 6 h limit / 560 evaluations & 8{,}594 / 2{,}896 & 8{,}593 / 60 \\
\bottomrule
\end{tabular}
\end{adjustbox}
\caption{Run-level outcomes: cost $J$ at 491 evaluations and at completion, or proxy $N_{\mathrm{proxy}}$ and certified bound $N$.}
\label{tab:optruns}
\end{table}

We also tested a supervised variant of prime-agent with Sol~5.6, outside the two study axes. A supervisor could send one note after 80 evaluations without improvement, or when at least 30\% of the previous 40 evaluations were rejected. The rejection condition was triggered in every run. Each note gave the rejection rate and remaining budget and reminded the agent that feedback identifies the failing check, without suggesting controller or certificate parameters. With supervision, cost was lower on one seed and higher on the other. The final bounds were 1{,}633 and 1{,}866, against 57 and 58 without supervision. With only two runs per objective, we cannot determine what caused these differences.

\begin{figure}[H]
\centering
\includegraphics[width=\textwidth,height=0.70\textheight,keepaspectratio]{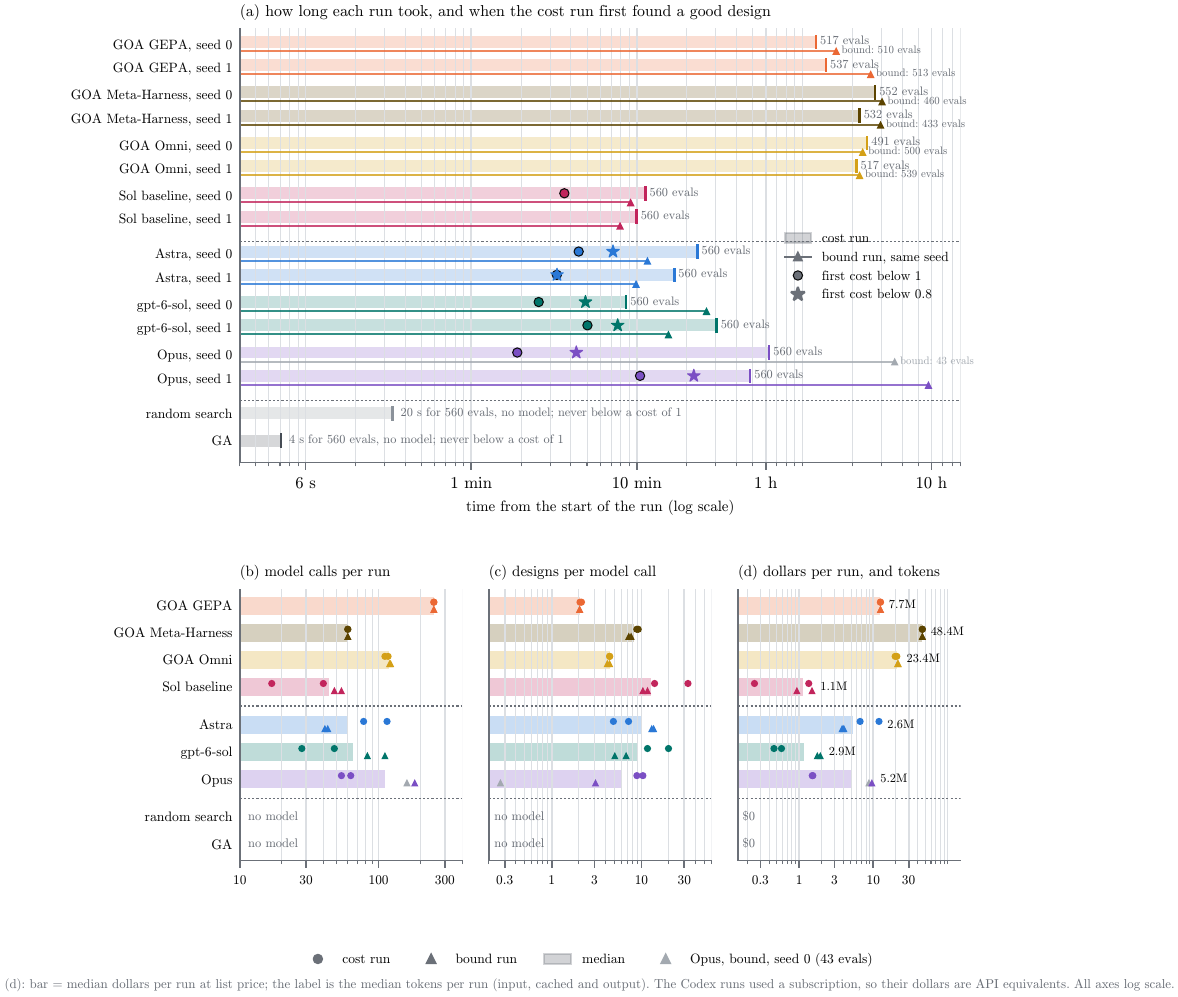}
\caption{Optimizer runtime (a), model calls (b), designs per call (c) and estimated price (d), on logarithmic axes.}
\label{fig:optresources}
\end{figure}

We also compare run time and resource use, and when each cost run first reached a certified design below 1 and 0.8 (Figure~\ref{fig:optresources}). This shows differences in how the approaches used their evaluation budgets beyond the final scores.

\begin{figure}[H]
\centering
\includegraphics[width=\textwidth,height=0.70\textheight,keepaspectratio]{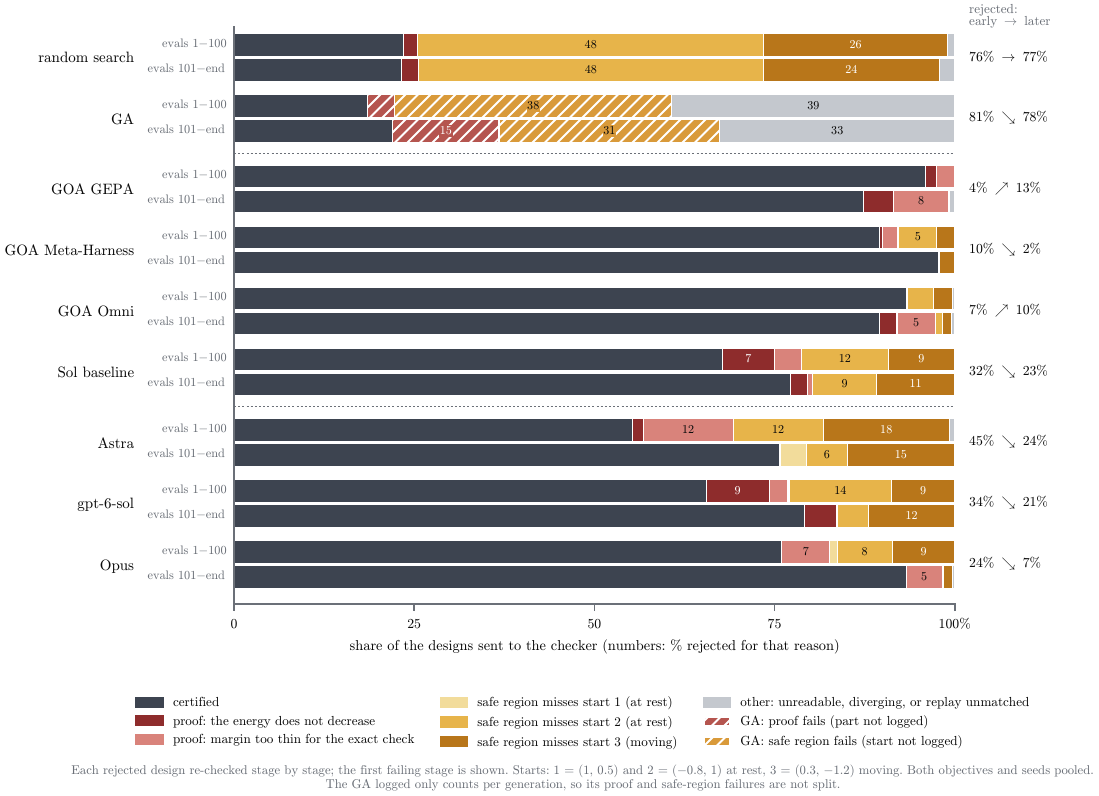}
\caption{First failing checker stage for rejected designs, comparing the first 100 evaluations with the remainder.}
\label{fig:optverdicts}
\end{figure}

We show rejections by the first failing stage in the checker's order in Figure~\ref{fig:optverdicts}. The original logs only give coarse verdicts, so we reconstructed the stages by re-checking designs from the 27 complete model runs and random search. For the model runs, the reconstructed stage agrees with the logged verdict in 14{,}492 of 14{,}501 evaluations; the remaining 9 are marked unmatched.

\newcommand{\mainref}[3]{#2~\ref{#1}}
\clearpage
\makeatletter
\let\@seccntformat\pce@seccntformat
\makeatother
\setcounter{section}{0}
\renewcommand{\thesection}{S\arabic{section}}
\renewcommand{\theHsection}{supplement.\arabic{section}}
\section*{Supplementary material}
\setcounter{table}{0}\setcounter{figure}{0}
\renewcommand{\thetable}{S\arabic{table}}\renewcommand{\theHtable}{S\arabic{table}}
\renewcommand{\thefigure}{S\arabic{figure}}\renewcommand{\theHfigure}{S\arabic{figure}}
\section{Paired pursuit comparisons}

Table~\ref{tab:supppursuit} reports the pursuit comparisons that include a
classical baseline. For each pair, the difference is computed within seed as
$r_A-r_B$, then summarized by its median and bootstrap 95\% confidence interval.
A negative difference favors A, because a smaller certified radius is better.
The A / B / T column counts seeds where A is better, B is better, or the methods
tie. The final columns give the unadjusted two-sided Wilcoxon $p$-value and its
Holm-adjusted value. The comparison family is described in
\mainref{app:more}{Appendix}{the paper's comparison-family appendix}.

\begin{table}[h]
\centering\scriptsize\setlength{\tabcolsep}{3pt}
\begin{adjustbox}{max width=\textwidth}
\begin{tabular}{@{}>{\raggedright\arraybackslash}p{0.2\textwidth}>{\raggedright\arraybackslash}p{0.2\textwidth}rrlccc@{}}
\toprule
A & B & $n$ & med.\ $A - B$ & 95\% CI & A / B / T & $p$ & Holm \\
\midrule
Gaussian mutation & CMA-ES & 20 & $-0.00115$ & $[-0.00389, +0.00075]$ & 11 / 9 / 0 & $0.0897$ & $0.41$ \\
PD grid search & Gaussian mutation & 20 & $-0.00752$ & $[-0.00997, -0.00514]$ & 20 / 0 / 0 & $1.9\times 10^{-6}$ & $4.0\times 10^{-5}$ \\
PD grid search & CMA-ES & 20 & $-0.01059$ & $[-0.01249, -0.00656]$ & 20 / 0 / 0 & $1.9\times 10^{-6}$ & $4.0\times 10^{-5}$ \\
Gaussian mutation & gpt-oss & 20 & $+0.00084$ & $[-0.00187, +0.00218]$ & 9 / 11 / 0 & $0.87$ & $0.87$ \\
Gaussian mutation & Astra & 20 & $+0.01283$ & $[+0.01045, +0.01528]$ & 0 / 20 / 0 & $1.9\times 10^{-6}$ & $4.0\times 10^{-5}$ \\
Gaussian mutation & Sol & 20 & $+0.01145$ & $[+0.00888, +0.01346]$ & 0 / 20 / 0 & $1.9\times 10^{-6}$ & $4.0\times 10^{-5}$ \\
Gaussian mutation & Opus & 20 & $+0.01027$ & $[+0.00918, +0.01197]$ & 0 / 20 / 0 & $1.9\times 10^{-6}$ & $4.0\times 10^{-5}$ \\
CMA-ES & gpt-oss & 20 & $+0.00188$ & $[-0.00073, +0.00615]$ & 6 / 14 / 0 & $0.0897$ & $0.41$ \\
CMA-ES & Astra & 20 & $+0.01589$ & $[+0.01187, +0.01780]$ & 0 / 20 / 0 & $1.9\times 10^{-6}$ & $4.0\times 10^{-5}$ \\
CMA-ES & Sol & 20 & $+0.01316$ & $[+0.01106, +0.01622]$ & 0 / 20 / 0 & $1.9\times 10^{-6}$ & $4.0\times 10^{-5}$ \\
CMA-ES & Opus & 20 & $+0.01151$ & $[+0.01033, +0.01585]$ & 0 / 20 / 0 & $1.9\times 10^{-6}$ & $4.0\times 10^{-5}$ \\
PD grid search & gpt-oss & 20 & $-0.00620$ & $[-0.00940, -0.00528]$ & 20 / 0 / 0 & $1.9\times 10^{-6}$ & $4.0\times 10^{-5}$ \\
PD grid search & Astra & 20 & $+0.00531$ & $[+0.00531, +0.00531]$ & 0 / 20 / 0 & $7.7\times 10^{-6}$ & $9.3\times 10^{-5}$ \\
PD grid search & Sol & 20 & $+0.00428$ & $[+0.00198, +0.00529]$ & 0 / 20 / 0 & $8.8\times 10^{-5}$ & $9.0\times 10^{-4}$ \\
PD grid search & Opus & 20 & $+0.00401$ & $[+0.00231, +0.00473]$ & 4 / 16 / 0 & $1.0\times 10^{-4}$ & $9.4\times 10^{-4}$ \\
\bottomrule
\end{tabular}
\end{adjustbox}
\caption{Paired pursuit comparisons. Negative differences favor A.}
\label{tab:supppursuit}
\end{table}

\clearpage
\section{Certificate optimization and thrust}

\paragraph{Certificate optimization.}
Table~\ref{tab:suppcertopt} compares the classical certificate optimizer with
Astra and Sol. Differences are the optimizer's value minus the model's value,
computed within seed. Both objectives are minimized, so a positive difference
means the optimizer is worse. Wins, losses and ties are counted from the
optimizer's side. Holm adjustment uses the declared family across both
problems.

\begin{table}[h]
\centering\scriptsize\setlength{\tabcolsep}{2.5pt}
\begin{adjustbox}{max width=\textwidth}
\begin{tabular}{@{}llrrlccc@{}}
\toprule
Problem & Against & $n$ & Median diff. & 95\% CI & W / L / T & $p$ & Holm \\
\midrule
docking & Astra & 20 & $+1.0$ & $[+0.0, +2.0]$ & 5 / 12 / 3 & $0.15$ & $0.15$ \\
docking & Sol & 20 & $+2.0$ & $[+1.0, +2.0]$ & 3 / 15 / 2 & $0.0697$ & $0.14$ \\
pursuit & Astra & 20 & $+0.00005$ & $[+0.00004, +0.00008]$ & 0 / 20 / 0 & $8.8\times 10^{-5}$ & $3.5\times 10^{-4}$ \\
pursuit & Sol & 20 & $-0.00098$ & $[-0.00324, +0.00002]$ & 13 / 7 / 0 & $0.00482$ & $0.0145$ \\
\bottomrule
\end{tabular}
\end{adjustbox}
\caption{The certificate optimizer against Astra and Sol. Positive differences
favor the model.}
\label{tab:suppcertopt}
\end{table}

\paragraph{Thrust demand.}
Table~\ref{tab:suppthrust} reports peak commanded acceleration per axis for the
final designs. Simulated peaks use the docking mission starts and the pursuit
fitness simulation's worst-case evader. For docking, the table also gives the
largest command anywhere in each certified set; this can exceed the simulated
peak. The ``$>u_m$'' columns count designs whose peak exceeds the actuator
limit $u_m=\ThrustLimit$~m/s$^2$. These measurements concern the original
unbounded-thrust designs; they do not establish performance with clipped thrust.

\begin{table}[h]
\centering\footnotesize\setlength{\tabcolsep}{3pt}
\begin{tabular}{@{}llrrrrrr@{}}
\toprule
 & & \multicolumn{3}{c}{simulated peak} & \multicolumn{3}{c}{certified-set peak} \\
\cmidrule(lr){3-5}\cmidrule(l){6-8}
Problem & Arm & median & max & $> u_m$ & median & max & $> u_m$ \\
\midrule
docking & Gaussian mutation & 0.021 & 0.028 & 0/20 & 0.030 & 0.040 & 0/20 \\
docking & CMA-ES & 0.014 & 0.023 & 0/20 & 0.025 & 0.035 & 0/20 \\
docking & alternating SDP synthesis (full allowance) & 0.137 & 0.158 & 0/20 & 0.213 & 0.219 & 20/20 \\
docking & Astra & 0.026 & 0.048 & 0/20 & 0.061 & 0.094 & 0/20 \\
docking & Sol & 0.027 & 0.170 & 0/20 & 0.063 & 0.223 & 2/20 \\
docking & Opus & 0.030 & 0.077 & 0/20 & 0.041 & 0.101 & 0/20 \\
docking & gpt-oss & 0.022 & 0.057 & 0/20 & 0.037 & 0.101 & 0/20 \\
pursuit & Gaussian mutation & 0.475 & 0.539 & 20/20 & -- & -- & -- \\
pursuit & CMA-ES & 0.513 & 0.634 & 20/20 & -- & -- & -- \\
pursuit & PD grid search & 0.775 & 0.775 & 20/20 & -- & -- & -- \\
pursuit & Astra & 0.918 & 0.918 & 20/20 & -- & -- & -- \\
pursuit & Sol & 0.881 & 1.079 & 20/20 & -- & -- & -- \\
pursuit & Opus & 0.864 & 1.097 & 20/20 & -- & -- & -- \\
pursuit & gpt-oss & 0.470 & 0.598 & 20/20 & -- & -- & -- \\
\bottomrule
\end{tabular}

\caption{Peak commanded acceleration per axis (m/s$^2$), in simulation and over
the docking certified sets.}
\label{tab:suppthrust}
\end{table}

\clearpage
\section{Coverage of the guarantees}

Table~\ref{tab:coverage} separates the guarantees established for each cited
design. Each group names the dynamics model. Statements concern rationalized
controllers and certificates over exact reals at sample instants. The columns
have the following meanings:

\begin{description}
\item[Exact check.] The certificate passes rational verification. ``In loop''
means this check is used during search; ``after search'' means the reported
design is checked afterwards.
\item[Lean theorems.] The number of kernel-checked theorems in the design's own
certificate file.
\item[Trajectory.] A theorem covers every trajectory from the mission initial
conditions, giving steps to the docking tolerance or to capture against every
admissible evader thrust. ``One step'' records regional decrease only, without
a trajectory theorem.
\item[Model step.] A proof identifies the certified update with the Euler step
of the stated model and gain.
\item[Noise, rounding.] Decrease and trajectory theorems cover bounded
measurement noise and rounding of $K(z+n)$, assuming the floating-point error
bound (\mainref{sec:noise}{Section}{the paper's noise section}).
\item[Thrust limit.] A decrease theorem covers clipped dynamics at
$u_m=\ThrustLimit$~m/s$^2$, checked after search
(\mainref{sec:sat}{Section}{the paper's saturation section}). This does not
extend the original speed-limit guarantee to clipped dynamics. The SO(3)
torque bound is a certified output, not an actuator limit.
\item[Zero-order hold.] The design's gain and certificate are checked on a
rational enclosure of the zero-order-hold plant. This exact check is outside
Lean (\mainref{sec:zoh}{Section}{the paper's zero-order-hold section}).
\end{description}

The marks distinguish the evidence: ``Lean'' is a kernel-checked theorem in an
audited file, ``exact'' is a rational check outside Lean, and ``--'' means the
property is not covered. The guarantees in different columns are separate;
for example, the table does not claim a combined noise-and-saturation theorem.

\begin{table}[h]
\centering\footnotesize\setlength{\tabcolsep}{3pt}
\begin{tabular}{@{}lccccccc@{}}
\toprule
Design & \begin{tabular}[b]{@{}c@{}}Exact\\check\end{tabular} & \begin{tabular}[b]{@{}c@{}}Lean\\theorems\end{tabular} & \begin{tabular}[b]{@{}c@{}}Trajectory\end{tabular} & \begin{tabular}[b]{@{}c@{}}Model\\step\end{tabular} & \begin{tabular}[b]{@{}c@{}}Noise,\\rounding\end{tabular} & \begin{tabular}[b]{@{}c@{}}Thrust\\limit\end{tabular} & \begin{tabular}[b]{@{}c@{}}Zero-order\\hold\end{tabular} \\
\midrule
\multicolumn{8}{@{}l}{\emph{docking: Clohessy--Wiltshire, Euler step at $\Delta t = 1$~s}} \\
\quad Gaussian mutation & in loop & 19 & Lean & Lean & Lean & Lean & exact \\
\quad CMA-ES & in loop & 19 & Lean & Lean & Lean & Lean & exact \\
\quad alternating SDP synthesis & in loop & 19 & Lean & Lean & Lean & Lean & exact \\
\quad Astra & in loop & 19 & Lean & Lean & Lean & Lean & exact \\
\quad Sol & in loop & 19 & Lean & Lean & Lean & Lean & exact \\
\quad Opus & in loop & 19 & Lean & Lean & Lean & Lean & exact \\
\quad gpt-oss & in loop & 19 & Lean & Lean & Lean & Lean & exact \\
\quad certificate optimizer & in loop & 19 & Lean & Lean & Lean & none$^{a}$ & exact \\
\quad hand-designed law & exact & 29 & Lean & -- & -- & Lean & -- \\
\multicolumn{8}{@{}l}{\emph{pursuit: Clohessy--Wiltshire, Euler step at $\Delta t = 1$~s; evader thrust bounded, pursuer not}} \\
\quad Gaussian mutation & in loop & 11 & Lean & Lean & Lean & -- & exact \\
\quad CMA-ES & in loop & 11 & Lean & Lean & Lean & -- & exact \\
\quad PD grid search & in loop & 11 & Lean & Lean & Lean & -- & exact \\
\quad Astra & in loop & 11 & Lean & Lean & Lean & -- & exact \\
\quad Sol & in loop & 11 & Lean & Lean & Lean & -- & exact \\
\quad Opus & in loop & 11 & Lean & Lean & Lean & -- & exact \\
\quad gpt-oss & in loop & 11 & Lean & Lean & Lean & -- & exact \\
\quad certificate optimizer & in loop & 11 & Lean & Lean & Lean & -- & exact \\
\multicolumn{8}{@{}l}{\emph{attitude: rigid body on SO(3), Cayley step at $h = 0.1$~s; region $\{V \le c\}$}} \\
\quad Gaussian mutation & after search & 44 & one step & -- & -- & -- & -- \\
\quad CMA-ES & after search & 41 & one step & -- & -- & -- & -- \\
\quad Astra & after search & 43 & one step & -- & -- & -- & -- \\
\quad Sol & after search & 39 & one step & -- & -- & -- & -- \\
\quad Opus & after search & 44 & one step & -- & -- & -- & -- \\
\quad gpt-oss & after search & 44 & one step & -- & -- & -- & -- \\
\quad hand-tuned law & exact & 34 & one step & -- & -- & -- & -- \\
\bottomrule
\end{tabular}
\par\smallskip\noindent\parbox{\linewidth}{\footnotesize $^{a}$No thrust-limit certificate: no $\theta$ in the grid both covers the mission states and admits a sector multiplier. The cited designs of the other arms (4 on docking, 4 on pursuit, 2 on SO(3)) each carry the marks of a campaign row of their problem.}

\caption{Guarantees established for the cited designs, grouped by problem.}
\label{tab:coverage}
\end{table}

\end{document}